\documentclass{article}

     \PassOptionsToPackage{numbers, compress}{natbib}
\usepackage[preprint]{neurips_2023}

\usepackage[utf8]{inputenc} 
\usepackage[T1]{fontenc}    
\usepackage{hyperref}       
\usepackage{url}            
\usepackage{booktabs, multirow, dcolumn}       
\usepackage{amsfonts}       
\usepackage{nicefrac}       
\usepackage{microtype}      
\usepackage{xcolor}         
\usepackage{float}
\usepackage{graphicx}
\usepackage{wrapfig}
\usepackage{amsmath, amssymb, amsthm}
\DeclareMathOperator{\Hom}{Hom}
\DeclareMathOperator{\Ind}{Ind}
\DeclareMathOperator{\Res}{Res}

\DeclareMathOperator{\SO}{SO}

\usepackage{wrapfig}
\usepackage{caption}
\graphicspath{ {./figures/} }
\usepackage{tikz-cd}
\usepackage{authblk}

\usepackage{amsthm}

\newtheorem{theorem}{Theorem}

\title{ Equivariant Single View Pose Prediction \\ Via Induced and Restricted Representations  }

\author[1]{Owen Howell \thanks{howell.o@northeastern.edu} }
\author[2]{David Klee}
\author[2]{Ondrej Biza}
\author[2]{Linfeng Zhao}
\author[2]{Robin Walters}
\affil[1]{ Department of Electrical and Computer Engineering, Northeastern University, Boston MA, 02115 }
\affil[2]{ Khoury College of Computer Sciences,
Northeastern University, Boston MA, 02115 }

\begin{document}

\maketitle

\begin{abstract}
Learning about the three-dimensional world from two-dimensional images is a fundamental problem in computer vision. An ideal neural network architecture for such tasks would leverage the fact that objects can be rotated and translated in three dimensions to make predictions about novel images. However, imposing $SO(3)$-equivariance on two-dimensional inputs is difficult because the group of three-dimensional rotations does not have a natural action on the two-dimensional plane. Specifically, it is possible that an element of $SO(3)$ will rotate an image out of plane. We show that an algorithm that learns a three-dimensional representation of the world from two dimensional images must satisfy certain consistency properties which we formulate as $SO(2)$-equivariance constraints. We use the induced and restricted representations of $SO(2)$ on $SO(3)$ to construct and classify architectures which satisfy these consistency constraints. We prove that any architecture which respects said consistency constraints can be realized as an instance of our construction. We show that three previously proposed neural architectures for 3D pose prediction are special cases of our construction. We propose a new algorithm that is a learnable generalization of previously considered methods. We test our architecture on three pose predictions task and achieve SOTA results on both the PASCAL3D+ and SYMSOL pose estimation tasks. 
\end{abstract}


\section{Introduction} \label{sec:introduction}

One of the fundamental problems in computer vision is learning representations of 3D objects from 2D images \cite{marr2010vision, Hartley_2004, Ozyesil_2017}. By understanding how image features correspond to a physical object, a model can generalize better to novel views of the object, for instance, when estimating the pose of an object.
In general, neural networks that respect the symmetries of a problem are more noise robust and data efficient, while also less prone to over-fitting \cite{Bronstein_2021}.
Three-dimensional space has a natural symmetry group of three-dimensional rotations and three dimensional translations, $SE(3)$. While we would like to leverage this symmetry to design improved neural architectures, serious challenges exist to incorporating 3D symmetry when applied to image data. Specifically, a projection of a three-dimensional scene into a two-dimensional plane does not transform equivariantly under all elements of $SE(3)$. This is because there is no a-priori model for how two-dimensional images transform under out-of-plane object rotations. The $SO(3)$ symmetry is reduced to the $SO(2)$ subgroup of $SO(3)$ which corresponds to all rotations that map the projection plane into the projection plane. \citet{Cohen_2016_II} showed how to design neural networks that are explicitly $SO(2)$-equivariant and accept images as inputs. However, this captures only the fact that the group truth lives in a space that is acted on by $SO(2) \subset SO(3)$ and disregards the fact that the $SO(3)$ also acts on the space of allowable ground truths.

The allowed architectures of $G$-equivariant neural networks are much more constrained then general multi-layer perceptrons. The requirement of $G$-equivariance places strict restrictions on the allowed linear maps and the allowed non-linear functions in each network layer \cite{Cohen_2016_II, Kondor_2018}. Because of this, the structure of allowable $G$-equivariant neural networks can be completely classified based on the representation theory of the group $G$ \cite{Bronstein_2021,Cohen_2018,Lang_2020}. Specifically, for compact groups, it is possible to completely characterize the structure of all possible kernels of $G$-equivariant networks \cite{Lang_2020}. 

\begin{wrapfigure}[17]{l}{0.40\textwidth}
\vspace*{-0.5cm}
\hspace*{-0.5cm}
\includegraphics[width=0.45\textwidth]{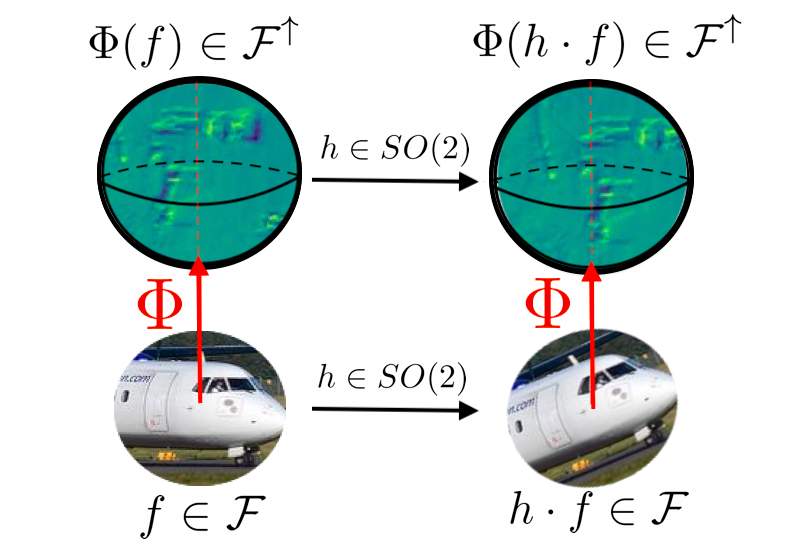}
\caption{ \small A map $\Phi : \mathcal{F} \rightarrow \mathcal{F}^{\uparrow}$ from signals on $\mathbb{R}^{2}$ to signals on $S^{2}$. Let $SO(2)$ be the subgroup that consists of all in-plane rotations ( i.e. about the axis defined by the red arrow). The map $\Phi$ must be equivariant with respect to this $SO(2) \subseteq SO(3)$ subgroup.   }\label{Figure:In plane condition}
\end{wrapfigure}

We argue that any equivarient machine learning algorithm that builds a three dimensional model of the world from two-dimensional images must satisfy a natural geometric consistency property. Using the restricted representation, this consistency property is equivalent to a set of $SO(2)$-equivarience constraints. We give a complete characterization of maps that satisfy this property. Using Frobinious Reciprocity theorem, we show that this geometric constraint can also be derived using induced representations. The classification theorems derived in \cite{Cohen_2016_II,Lang_2020,Bronstein_2021} are derived assuming that both the input and output layers are $G$-equivariant. For the construction presented in \ref{Section:Main:Method}, we instead map $H$-equivariant functions to $G$-equivariant functions. Our restricted/induced representation arguments give a natural generalization of equivarient maps between different groups. We derive the induced and restricted representation analogies of the theorems presented in \cite{Cohen_2016,Lang_2020}.

\subsection{Importance and Contribution}	 

In this work, we will show how the induced and restricted representations can be used to construct neural architectures that accept image data and leverage $SO(3)$-equivarient methods to avoid learning nuisance transformations in three-dimensional space.

We show that our proposed construction satisfies both a \emph{completeness} property and a \emph{universal} property. Specifically, let $H$ be the subgroup of $G$ that maps in-plane images to in-plane images. The induced representation construction is \emph{complete} in that all group valued functions on $G$ can be induced from a set of group valued functions on $H$. The construction is \emph{universal} in that all multi-linear maps which map $H$-equivariant functions to $G$-equivariant functions are specific cases of the induced representation, modulo isomorphism. Furthermore, we show that the architectures proposed in \cite{Klee_2022, Esteves_2019} are special cases of our construction for the icosahedral group $G=A_{5}$ and the construction proposed in \cite{Klee_2023} is a special case of our construction for the three-dimensional rotation group $G=SO(3)$. Our method achieves state of the art performance for orientation prediction on PASCAL3D+ \cite{Xiang_2014} and SYMSOL \cite{Murphy_2022} datasets.

\paragraph{Contributions:}
\vspace{-0.1cm}
\begin{itemize} 
\item We propose a unified theory for learning three dimensional representations from two dimensional images. We show that algorithms which learn three-dimensional representations from two-dimensional images must satisfy certain consistency properties, which are equivalent to $SO(2)$-steerability constraints. 

\item We introduce a fully differentiable layer called an \emph{induction/restriction layer} that maps signals on the plane into signals on the sphere. We show that the induction/restriction layer satisfies a natural consistency constraint and prove both a completeness and universal property for our construction.

\item Our method achieves SOTA performance for orientation prediction on PASCAL3D+ and SYMSOL datasets.
\end{itemize}

\section{Related Work} \label{sec:related-work}
\paragraph{Equivariant Learning} Incorporating problem symmetry into the design of neural networks has been effective in domains such as computer vision \cite{lecun1995convolutional, shaw2018self}, point cloud processing \cite{qi2017pointnet, Thomas_2018}, and robotics \cite{Wang_2022}. \citet{Cohen_2016} introduced the group convolution operation, a trainable layer that can be used to build networks that are equivariant to 2D \cite{Cohen_2016_II, Weiler_2021} and 3D transformations \cite{Weiler_2018, Cohen2018_spherical}. The majority of past works have studied end-to-end equivariant models, where the input can be transformed by the action of the group.

There has been growing interest in leveraging 3D symmetry from 2D inputs. \cite{Falorsi_2018, park2022learning} learned a 3D transformable latent space from images of a single object. \cite{esteves2019cross} trained a convolutional network to predict pre-trained $SO(3)$ equivariant embeddings, while \cite{Esteves_2019, Klee_2022, Klee_2023} mapped image features onto elements of the discrete group of $SO(3)$, using structured view points or a hand-coded projections, respectively. In contrast to prior work, we provide a theoretical foundation for learned equivariant mappings from 2D to 3D, which additionally guides us to introduce a more effective learnable mapping operation.

\paragraph{Object Pose Estimation}
Predicting the 3D rotation of objects is an important problem in fields like autonomous driving \cite{geiger2013vision}, robotics \cite{xiang2017posecnn} and cryogenic electron microscopy \cite{Zhong_2020}. Many works \cite{Tulsiani_2015, Mahendran_2018} have used a regression approach, and others \cite{Zhou_2020_Cont, Bregier_2021, Liao_2019} have identified ways to mitigate the discontinuities along the $SO(3)$ manifold. More recent works have explored ways to model pose as a distribution over 3D rotations, which handles object symmetries and captures uncertainty. \cite{Deng_2020}, 
\cite{Prokudin_2018} and \cite{Yin_2023} predict parameters for Bingham, von Mises and Laplace distributions, respectively.  These families of distributions can have limited expressivity, so other work explored using implicit networks \cite{Murphy_2022} or the Fourier basis \cite{Klee_2023} to model more complex pose distributions.

\section{Background} \label{sec:background}

We introduce the induced and restricted representations. For a more extensive review of representation theory, see \ref{Appendix:Section:Notation and Preliminaries}.

\label{Main:Section:Induced and Restriction Representations}
Let $V$ be a vector space over $\mathbb{C}$. A \emph{representation} $(\rho , V)$ of $G$ is a map $\rho : G \rightarrow \Hom[V,V]$ such that 
\begin{align*}
\forall g , g' \in G, \enspace \forall v\in V, \quad   \rho( g \cdot g' )v =  \rho( g ) \cdot \rho(  g' )v
\end{align*}
Concisely, a group representation is a embedding of a group into a set of matrices. The matrix embedding must obey the multiplication rule of the group. We introduce the \emph{Restricted Representation} and \emph{Induced Representation}.

\paragraph{Restricted Representation}
Let $H \subseteq G$. Let $(\rho , V)$ be a representation of $G$. The restricted representation of $(\rho , V)$ from $G$ to $H$ is denoted as $\Res_{H}^{G}[ (\rho , V) ]$. Intuitively, $\Res_{H}^{G}[ (\rho , V) ]$ can be viewed as $(\rho , V)$ evaluated on the subgroup $H$ of $G$. Specifically, 
\begin{align*}
\forall h\in H, \enspace \forall v\in V, \quad \Res_{H}^{G}[ \rho ](h) v = \rho(h) v
\end{align*}
For a more in depth discussion of the restricted representation, please see \ref{Appendix:Section:Notation and Preliminaries}.

\paragraph{Induced Representation}

The induced representation is a way to construct representations of a larger group $G$ out of representations of a subgroup $H \subseteq G$. Let $(\rho , V)$ be a representation of $H$. The induced representation of $(\rho , V)$ from $H$ to $G$ is denoted as $\Ind_{H}^{G}[ (\rho , V) ]$. Define the space of functions
\begin{align*}
\mathcal{F} = \{ \enspace f \enspace | \enspace f : G \rightarrow V, \enspace \forall h\in H, \enspace f(gh) = \rho(h^{-1}) f(g) \enspace \}
\end{align*}
Then the induced representation is defined as $ ( \pi , \mathcal{F}  ) = \Ind_{H}^{G}[ (\rho , V) ]$
where the induced action $\pi$ acts on the function space $\mathcal{F}$ via
\begin{align*}
    \forall g,g' \in G, \enspace \forall f \in \mathcal{F}, \quad (\pi(g)\cdot f)(g') = f(g^{-1} g')
\end{align*}
Please see \ref{Appendix:Section:Notation and Preliminaries} for an in depth discussion of the induced representation. The induced and restricted representations are adjoint functors \cite{Ceccherini_2008}.


\section{Method} \label{Section:Main:Method}

Convolutional networks or vision transformers are typically used to extract spatial feature maps from 2D images. For convenience we ignore discritization and treat the feature maps as having continuous inputs $f: \mathbb{R}^2 \rightarrow \mathbb{R}^{d}$. To leverage spatial symmetries in 3D, we would like to map our features $f$ from a plane onto a sphere: $g: S^2 \rightarrow \mathbb{R}^{D}$. \citet{Klee_2023} proposed one such mapping, where the planar feature map is stretched over a hemisphere, but other possible mappings exist.

We formalize the equivariance property every projection should have through the theory of induced and restricted representations. The constraints that we impose have a intuitive geometric interpretation. We give a complete characterization of \textit{all possible} linear and equivariant projections, $\Phi$, from planar features to a spherical representation. Our general formulation includes \cite{Klee_2023} as a special case, and we show that a learnable equivariant projection leads to better predictive models.

\subsection{Equivariant 2D to 3D Projection by Induced and Restricted Representations}

We first derive the $SO(2)$-equivariance constraint for the most general linear mapping from images to spherical signals.

\paragraph{Image inputs} We first describe $\mathcal{F}$ the space of image input signals.
Let $V$ and $V^{\uparrow}$ be vector spaces. Let $\mathcal{F}$ be the vector space of all $V$-valued signals defined on the plane
\begin{align*}
    \mathcal{F} = \{ \enspace f \enspace | \enspace f: \mathbb{R}^{2} \rightarrow V \enspace \}.
\end{align*}

Elements of $\mathcal{F}$ are sometimes called $SE(2)$-steerable feature fields \citep{Weiler_2021}. The group $SE(2)= \mathbb{R}^{2} \rtimes SO(2)$ of 2D translations and rotations acts on $\mathcal{F}$ via representation $\pi$. 
Each $h \in SE(2)$ has a unique factorization $h = \bar{h} h_{c}$ where $\bar{h} \in \mathbb{R}^{2}$ is a translation and $h_{c} \in SO(2)$ is a rotation.
Then the action $\pi$ is defined
\begin{align*}
\forall f \in \mathcal{F},  \enspace  r \in \mathbb{R}^{2}, \enspace   h  \in SE(2), \enspace \quad     \pi(h) \cdot f(r) = \rho(h_{c}) f(h^{-1}r)
\end{align*}
where $( \rho , V)$ is an $SO(2)$-representation describing the transformation of the fibers of $f$ and $( \pi , \mathcal{F} ) = \Ind_{ SO(2) }^{ SE(2) } [ ( \rho , V) ]$ so that $( \pi , \mathcal{F} )$ gives a representation of the group $SE(2)$ \cite{Cohen_2016}. 

\paragraph{Spherical outputs} We would like to map signals in $\mathcal{F}$ into functions from $S^{2}$ into the vector space $V^{\uparrow}$. Let $\mathcal{F}^{\uparrow}$ be the vector space of all such outputs defined as
\begin{align*}
    \mathcal{F}^{\uparrow} = \{ \enspace f \enspace | \enspace f: S^{2} \rightarrow V^{\uparrow} \enspace \}
\end{align*}
The group $SO(3)$ acts on the vector space $\mathcal{F}^{\uparrow}$ via
\begin{align*}
 \forall f^{\uparrow} \in \mathcal{F}^{\uparrow}, \enspace  \hat{n} \in S^{2} , \enspace  g  \in SO(3),  \quad    \pi^{\uparrow}(g) \cdot f^{\uparrow}( \hat{n} ) = \rho^{\uparrow}(g) f^{\uparrow}(  g^{-1} \hat{n} ) 
\end{align*}
where $\rho^{\uparrow}(g) $ describes the $SO(3)$ fiber representation.

\paragraph{$SO(2)$-equivariant image to sphere} Let $H = SO(2)$ be the $SO(2)$ subgroup of $SO(3)$ that corresponds to in-plane rotations of the image. Our goal is to classify $H$-equivariant linear maps $\Phi : \mathcal{F} \rightarrow \mathcal{F}^{\uparrow}$.
This is equivalent to the constraint that
\begin{align}\label{Main:Equation:In Plane Equivarence_I}
\forall  h \in H = SO(2), \enspace f \in \mathcal{F}, \quad \Phi( \pi(h) \cdot f ) = \pi^{\uparrow} (h) \cdot \Phi(f)  
\end{align}
The constraint enforces equivarient with respect to $SO(2)$ transformations. By definition, the evaluation of $\pi^{\uparrow}(h)$ at $ h \in SO(2)$ subgroup is the restricted representation $\pi^{\uparrow}(h) = \Res_{SO(2)}^{SO(3)}[ \pi^{\uparrow} ](h)$.

\subsection{Solving the Kernel Constraint}\label{Section:Main:Deriving the Kernel Constraint}

We use tools from \cite{Weiler_2018_II,Lang_2020} to solve for the space of all possible maps satisfying the constraint \ref{Main:Equation:In Plane Equivarence_I}, giving the trainable space for the image to sphere layer. 

\begin{figure}[H]
\centering
\begin{tabular}{cc}
 \includegraphics[width=0.45\textwidth]{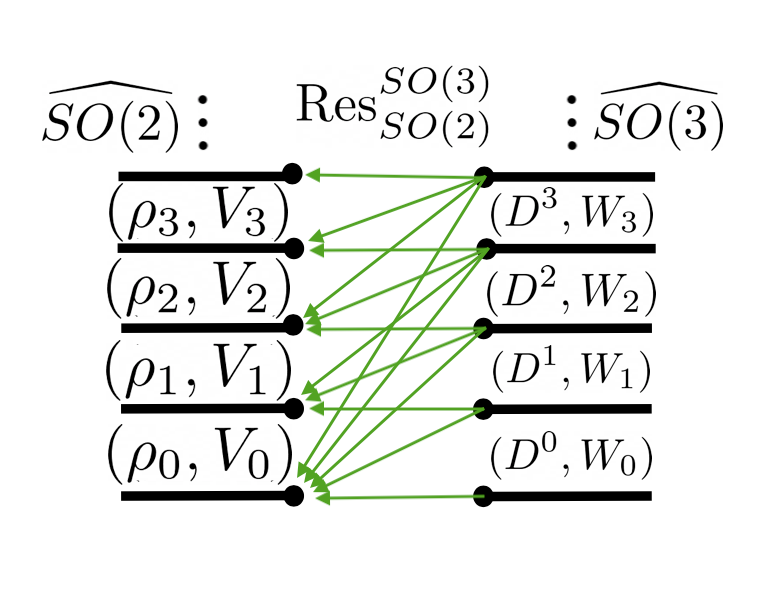} 
 \includegraphics[width=0.45\textwidth]{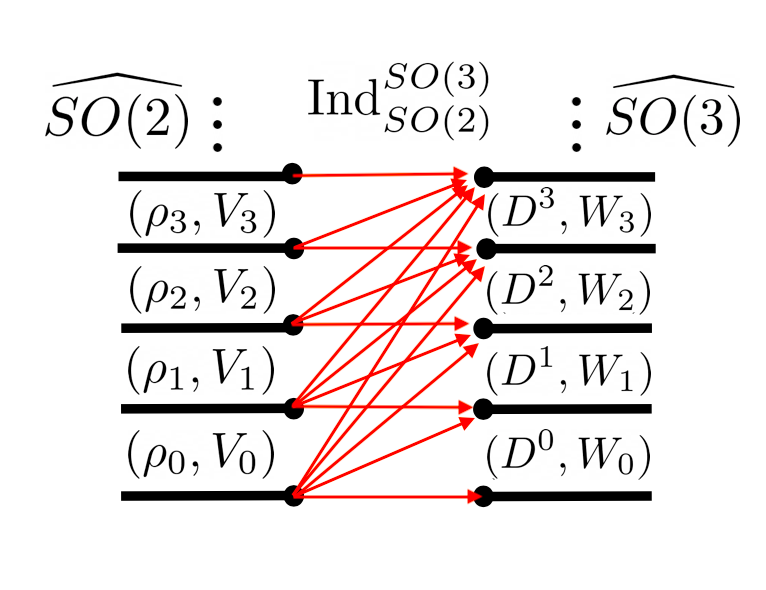}
\end{tabular}
\vspace{-0.5cm}
\caption{ Left: Decomposition of the restricted representation $\Res_{SO(2)}^{SO(3)}$ of $SO(3)$-irreducibles $(D^{\ell} , W_{\ell}) \in \widehat{SO(3)}$ into $SO(2)$-irreducibles  $(\rho_{k} , V_{k}) \in \widehat{SO(2)} $. Not every $SO(2)$-representation can be realized as the restriction of a $SO(3)$-representation. Right: Decomposition of the induced representation $\Ind_{SO(2)}^{SO(3)}$ for $SO(2)$-irreducibles $(\rho_{k} , V_{k}) \in \widehat{SO(2)} $ into $SO(3)$-irreducibles $(D^{\ell} , W_{\ell}) \in \widehat{SO(3)}$. Not every $SO(3)$-representation can be realized as the induction of a $SO(2)$-representation. }\label{Figure:Induced_SO_2_Main}
\end{figure}

Our conclusion is that instead of mapping arbitrary $SO(2)$-input representation to arbitrary $SO(2)$-output representation, the allowed input and output representations $(\rho , V)$ and $( \rho^{\uparrow} , V^\uparrow) $ must satisfy additional constraints. Specifically, not every representation can be realized as the restriction of an $SO(3)$ to $SO(2)$ representation \ref{Figure:Induced_SO_2_Main}. Although in this paper we focus on orientation estimation, the equivariant framework in Section \ref{Section:Deriving the Kernel Constraint} is more general. In the Appendix \ref{Appendix:Section:Plane to Space for Object Reconstruction}, we formulate and solve analogous equivariance constraints for both 6DoF-pose estimation and monocular volume reconstruction.

\begin{theorem}
The constraint in Equation \ref{Main:Equation:In Plane Equivarence_I} can be solved exactly using the results of \cite{Weiler_2018_II,Lang_2020}. The most linear general map $\Phi : \mathcal{F} \rightarrow \mathcal{F}^{\uparrow}$ can be expanded as
\begin{align*}
[\Phi(f)](\hat{n}) = \int_{r \in \mathbb{R}^{2} }dr\text{ } \kappa( \hat{n} , r )f(r)
\end{align*}
where $\kappa: \mathbb{R}^{2} \times S^{2} \rightarrow \Hom[V,V^{\uparrow}]$. Then, the exact form of $\kappa$ can be written as 
\begin{align}
\kappa( \hat{n} , r ) = \sum_{ \ell=0 }^{\infty} F_{\ell}( r )^{T} Y_{\ell}(\hat{n} ) 
\end{align}
where $Y_{\ell}(\hat{n} )$ is the vectorization of the $\ell$-type spherical harmonics and each $F_{\ell}( r )$ is an standard $SO(2)$-steerable kernel \cite{Cohen_2016,Weiler_2018_II} that has input $SO(2)$-representation $(\rho , V)$ and output $SO(2)$-representation $(\rho^{\ell} , V^{\ell} ) =(\rho , V) \otimes \Res_{SO(2)}^{SO(3)}[ (D^{\ell} , V^{\ell} )  ] $.
\end{theorem}
The proof of this statement is given in Appendix \ref{Appendix:Section:Solving the Kernel Constraint}. Note that similar to \cite{Thomas_2018,Kondor_2018} the tensor product structure of the $SO(2)$ and $SO(3)$ irreducible representations determine the allowed input and output representations of the matrix valued harmonic coefficients $F_{\ell}( r )$.

\subsection{Including Non-Linearities }

In section \ref{Section:Main:Deriving the Kernel Constraint}, we considered the most general linear maps that satisfied the generalized equivariance constraint. Adding non-linearities should allow for more expressiveness. Understanding non-linearities between equivariant layers is still an active area of research \cite{Franzen_2021,Dehaan_2021,Poulenard_2021,Xu_2022}. 

One way to include non-linearity is to apply standard $SO(3)$ non-linearities after the linear induction layer. After applying the linear mapping described in \ref{Section:Image to ...}, we apply an additional spherical non-linearity \cite{Geiger_2022} to the signal on $S^{2}$. This is the method we employ for the results presented in \ref{Main:Section:Implementation and Training Details}. As shown in \ref{Appendix:Section:Including Non-linearities} it is also possible to include tensor-product based non-linearity analogous to the results of \cite{Thomas_2018,Kondor_2018}.


\section{Theory} \label{sec:theory}

\subsection{Universal Property}
In section \ref{Section:Main:Method} we showed how the restriction representation arises naturally when trying to construct $SO(3)$-equivariant architectures for image data. However, there is no apriori choice of the hidden $SO(3)$ representation. We show that with this choice, our construction satisfies a universal property, and is unique up to isomorphism \cite{Leinster_2016}.

We have the following universal property of induced representations, as stated in \cite{Ceccherini_2008}: 
\begin{theorem}
Let $H \subseteq G$. Let $(\rho , V)$ be any $H$-representation. Let $\text{Ind}^{G}_{H}( \rho , V )$ be the induced representation of $(\rho , V)$ from $H$ to $G$. Then, there exists a \underline{unique} $H$-equivariant linear map $\Phi_{\rho} : V \rightarrow \text{Ind}^{G}_{H}V$ such that for any $G$-representation $(\sigma , W)$ and any $H$-equivariant linear map $\Psi : V \rightarrow W$, there is a unique $G$-equivariant map $\Psi^{\uparrow} : \text{Ind}_{H}^{G}V \rightarrow W$ such that the diagram \ref{Diagram:Universality Property_I} is commutative.
\end{theorem}

\begin{wrapfigure}{L}{0.3\textwidth}
\begin{tikzcd}
(\rho , V) \arrow{r}{\Phi_{\rho}} \arrow[swap]{dr}{ \Psi } & \text{Ind}^{G}_{H}(\rho ,  V )\arrow{d}{\Psi^{\uparrow}} \\
& (\sigma , W )
\end{tikzcd}
\caption{ Commutative Diagram for Uniqueness Property of Induced Representations.  }\label{Diagram:Universality Property_I}
\end{wrapfigure}

Let $(\rho , V)$ be a $H$-representation and let $(\sigma , W)$ be a $G$-representation. Let $\Psi : V \rightarrow W$ where $\Psi$ is an intertwiner of a the $H$-representation and the restriction of the $G$-representation to an $H$-representation so that
\begin{align*}
    \forall h \in H, \quad \Psi \rho(h) = \Res_{H}^{G}[\sigma ](h) \Psi
\end{align*}
so that $\Psi \in \Hom_{H}[ (\rho , V) , \Res_{H}^{G}(\sigma , W) ]$. The universal property of the induced representation allows us to write any such $\Psi$ in a canonical form. Specifically, as illustrated in \ref{Diagram:Universality Property_Push}, we can always uniquely decompose $\Psi =\Psi^{\uparrow} \circ \Phi_{\rho}$ where $\Psi^{\uparrow} \in\Hom_{G}[ \text{Ind}_{H}^{G}(\rho , V) , (\sigma , W) ]$ and $\Psi_{\rho} : V \rightarrow \text{Ind}_{H}^{G}V$ is $( \sigma , W) $ independent.
	
\begin{center}\label{Diagram:Universality Property_Push}
\begin{tikzcd}\centering
    & (\rho , V) \arrow{r}{ \Psi } \arrow[swap]{d}{ \rho(h) } & (\sigma , W ) \arrow{d}{ \sigma(g) }[swap]{ \sigma(h) } \\
    & ( \rho , V ) \arrow{r}{ \Psi }& (\sigma , W )
\end{tikzcd} \enspace $\cong$
\begin{tikzcd}\centering
    (\rho , V) \arrow{dr}{\Phi_{\rho}} \arrow{rr}{\Psi}  \arrow{ddd}[swap]{\rho(h)}
    && (\sigma , W) \arrow{ddd}{\sigma(g)}[swap]{\sigma(h)} \\
    & \text{Ind}_{H}^{G}(\rho , V) \arrow{ur}{\Psi^{\uparrow}} \arrow{d}[swap]{ [\Ind_{H}^{G}\rho](h)}{ [\Ind_{H}^{G}\sigma](g)}  \\
    & \text{Ind}_{H}^{G}(\rho , V)  \arrow{dr}{\Psi^{\uparrow}}  \\
    (\rho , V) \arrow{ur}{\Phi_{\rho}} \arrow{rr}{\Psi} && (\sigma  , W) \\
\end{tikzcd}
\captionof{figure}{Factorization Identity for Universal Property of Induced Representations}
\end{center}

Convolutional neural networks are compositions of linear functions, interleaved with non-linearities. At each layer of the network, we have a set of functions from a homogeneous space of a group into some vector space \cite{Kondor_2018}. Let $X^{H}_{i}$ be a set of homogeneous spaces of the group $H$ and let $X^{G}_{j}$ be a set homogeneous spaces of the group $G$. Let $V^{H}_{i}$ and $W^{G}_{j}$ be a set of vector spaces. Then, consider the function spaces
\begin{align*}
    \mathcal{F}^{H}_{i} = \{  \enspace f \enspace | \enspace f : X^{H}_{i} \rightarrow V^{H}_{i} \enspace \}, \quad \quad	\mathcal{F}^{G}_{j} = \{ \enspace f' \enspace | \enspace f' : X^{G}_{j} \rightarrow W^{G}_{j} \enspace \}
\end{align*}
The group $H$ acts on the homogeneous spaces $X^{H}_{i}$ and the group $G$ acts on the homogeneous spaces $X^{G}_{j}$ so that the function spaces $\mathcal{F}^{H}_{i}$ and $\mathcal{F}^{G}_{j}$ form representations of $H$ and $G$, respectively
	
Suppose we wish to design a downstream $G$-equivariant neural network that accepts as signals functions that live in the vector space $\mathcal{F}^{H}_{0}$ and transform in the $\rho_{0}$ representation of $H$. Thus, $( \rho_{0} , \mathcal{F}^{H}_{0})$ is a $H$-representation, but not necessarily a $G$-representation. At some point, in the architecture, a layer $\mathcal{F}^{H}_{i}$ must be $H$ equivariant on the left and both $H$ and $G$-equivariant on the right. Let us call the layer that is both $H$ and $G$-equivariant $\mathcal{F}^{G}_{1}$.

\scriptsize
\begin{center}\label{Diagram:Switching}
    \hspace{-0.5cm}
    \begin{tikzcd}[column sep=4ex,row sep=5ex]
        & ...\arrow{r}{\Phi_{i-1}}  & ( \rho_{i}, \mathcal{F}_{i}^{H} ) \arrow{d}{{ \rho_{i}(h) } } \arrow{r}{ \Psi } & ( \sigma_{1} , \mathcal{F}_{1}^{G} ) \arrow{d}{ \sigma_{1}(g) } \arrow{r}{\Psi_{1}} & ...  \\
        & ...\arrow{r}[swap]{\Phi_{i-1}}  &( \rho_{i} , \mathcal{F}_{i}^{H} )\arrow{r}[swap]{\Psi} & ( \sigma_{1} , \mathcal{F}_{1}^{G} ) \arrow{r}[swap]{\Psi_{1}} &  ...
    \end{tikzcd} \quad $\cong$  \begin{tikzcd}[column sep=4ex,row sep=5ex]
        & ...\arrow{r}{\Phi_{i-1}}  & ( \rho_{i} , \mathcal{F}_{i}^{H} ) \arrow{d}{{ \rho_{i}(h) } } \arrow{r}{ \Phi_{ \rho_{i} } } & \text{Ind}_{H}^{G}[ ( \rho_{i} ,   \mathcal{F}^{H}_{i} ) ]\arrow{d}{ \text{Ind}_{H}^{G}[\rho_{i}]} \arrow{r}{\Psi^{\uparrow}}& ( \sigma_{1} , \mathcal{F}_{1}^{G}  ) \arrow{d}{ \sigma_{1}(g) } \arrow{r}{\Psi_{1}} & ...  \\
        & ...\arrow{r}[swap]{\Phi_{i-1}}  &( \rho_{i} , \mathcal{F}_{i}^{H}) \arrow{r}[swap]{\Phi_{\rho_{i}}} &  \text{Ind}_{H}^{G}[  (\rho_{i} ,   \mathcal{F}^{H}_{i} )  ] \arrow{r}[swap]{\Psi^{\uparrow}} & ( \sigma_{1} , \mathcal{F}_{1}^{G} ) \arrow{r}[swap]{\Psi_{1}} & ...
    \end{tikzcd}
    \captionof{figure}{ Factorization of Generic Architecture Using Universal Property of Induced Representation. Any network that has input layer $(\rho_{i} , \mathcal{F}^{H}_{i} )$ that is $H$-equivariant and output layer $(\sigma^{G}_{1} , \mathcal{F}^{G}_{1} )$ that is $G$-equivariant can be factorized in terms of the induced representation. The map $\Psi = \Psi^{\uparrow} \circ \Phi_{\sigma_{i}}$  where $\Psi^{\uparrow}$ is $G$-equivariant and $\Phi_{\sigma_{i}}$ is $H$-equivariant.}
\end{center}
\normalsize

Suppose that $\Psi$ is an intertwiner between $( \rho_{i} , \mathcal{F}^{H}_{i})$ and $(\sigma_{1} , \mathcal{F}^{G}_{1})$. Using the factorization property of induced representations \ref{Diagram:Universality Property_Push}, there is a canonical basis of the space $\Hom_{H}[ (\rho_{i}, \mathcal{F}^{H}_{i}) ,  \Res_{H}^{G}[(\sigma_{1} , \mathcal{F}_{1}^{G} )] ] \cong \Hom_{G}[ \Ind_{H}^{G}[ (\rho_{i} , \mathcal{F}^{H}_{i})] , (\sigma_{1} , \mathcal{F}^{G}_{1}) ] $ and we may write $\Psi$ uniquely as $\Psi = \Psi^{\uparrow } \circ \Phi_{\rho}$ where $\Phi_{\rho}$ is an $H$-equivariant map and $\Psi^{\uparrow }$ is a $G$-equivariant map. Thus, any boundary between $H$ and $G$ layers can be written as an $H$-equivariant layer between $(\rho_{i}, \mathcal{F}^{H}_{i})$ and $\Ind_{H}^{G}[(\rho_{i}, \mathcal{F}^{H}_{i})]$ followed by a $G$-equivariant layer between $\Ind_{H}^{G}[(\rho_{i}, \mathcal{F}^{H}_{i})]$ and $(\sigma_{1}, \mathcal{F}^{H}_{1})$. In this way, induction is all you need and all possible latent $G$-equivariant architectures can be written in terms of the induction representation.

\section{Experiments} \label{Main:Section:Experiments}

\subsection{Datasets \& Evaluation Metrics}

We evaluate the performance of our method on three single-object pose estimation datasets.  These datasets require making predictions in $SO(3)$ from single 2D images.  \textbf{SYMSOL} \cite{Murphy_2022} consists of a set of images of marked and unmarked platonic solids, taken from different vantage points. Training data is annotated with viewing direction. Some objects have symmetries so that there are multiple equivalent viewing directions. which requires learning distributions over poses.
\textbf{PASCAL3D+} \cite{Xiang_2014} is a popular benchmark for object pose estimation composed of real images of objects from twelve categories.  This dataset is challenging to do the large variation in object appearances and the presence of novel object instances in the test set.  To be consistent with the baselines, we augment the training data with synthetic renderings\cite{Su_2015} and evaluate performance on the PASCALVOC\_val set. For more details on the benchmark datasets and additional numerical experiments, see \ref{Appendix:Section:Datasets}.

\begin{wrapfigure}{l}{0.40\textwidth}
\includegraphics[width=0.40\textwidth]{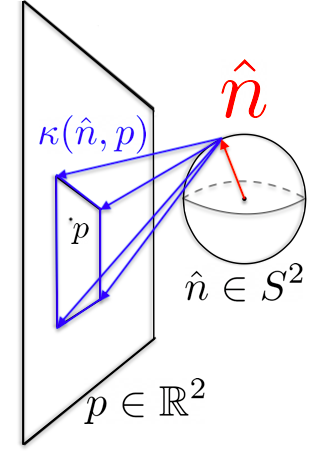}
\caption{ \small Diagram of an Equivariant Image to Sphere Convolution. At each unit vector $\hat{n}\in S^{2}$ the kernel $\kappa( \hat{n}  : p ) $ is dependent on the image point $p = (x,y) \in \mathbb{R}^{2}$. Equivariance constraints put restrictions on the allowed form of $\kappa( \hat{n} : p)$. Similar to a standard convolution, the kernel $\kappa$ has a user defined receptive field. }\label{Figure:kernel_diagram_II.pdf}
\vspace{-1.5cm}
\end{wrapfigure}


When a single ground truth rotation label is provided, we evaluate the method using the geodesic distance between the predicted and ground truth rotation matrices, reported as either median rotation error or accuracy at a given rotation error threshold. For SYMSOL, which provides the full set of equivalent rotations associated with an image, we measure the accuracy of the learned pose distribution using average log likelihood. This is also the accuracy metric used in \cite{Klee_2023}.

\subsection{Implementation \& Training Details}\label{Main:Section:Implementation and Training Details}

For the results presented in \ref{Main:Section:Experiments}, we use a ResNet encoder with weights pre-trained on ImageNet. With 224x224 images as input, this generates a 7x7 feature map with 2048 channels. 

The filters in the induction layer were instantiated using the e2nn \cite{Weiler_2018_II} package. The maximum frequency was chosen to be $\ell = 6$. The output of the induction layer was chosen to be a $64$-channeled $S^{2}$ signal with fibers transforming in the trivial representation of $SO(3)$. After the induction layer, a spherical convolution operation is performed using a filter that is parameterized in the Fourier domain, which generates an 8-channel signal over SO(3). A spherical non-linearity is applied by mapping the signal to the spatial domain, applying a ReLU, then mapping back to Fourier domain. One final spherical convolution with a locally supported filter is performed to generate a one-dimensional signal on SO(3). The output signal is queried using an SO(3) HEALPix grid (recursion level 3 during training, 5 during evaluation) and then normalized using a softmax following \cite{Murphy_2022}. $S^{2}$ and $SO(3)$ convolutions were performed using the e3nn \cite{Geiger_2022} package. The network was initialized and trained using PyTorch \cite{Paszke_2019}.

In order to create a fair comparison to existing baselines, batch size(=64), number of epochs(=40), optimizer(=SGD) and learning rate schedule(=StepLR) were chosen to be the same as that of \cite{Klee_2023}. Numerical experiments were implemented on NVIDA P-100 GPUs.



\subsection{Comparison to Baselines}

We compare our method's performance to competitive pose estimation baselines. We include regression methods, \cite{Tulsiani_2015, Mahendran_2018, Liao_2019}, that perform well on datasets where objects have a single valid pose (e.g. are non-symmetric or symmetry is disambiguated in labels).  We also baseline against methods that model pose with parametric families of distributions, \cite{Prokudin_2018, Mohlin_2021, Deng_2020, Yin_2023}, an implicit model \cite{Murphy_2022}, and the Fourier basis of $\SO(3)$ \cite{Klee_2023}.  To make the comparison fair, all methods use the same-sized ResNet backbone for each experiment, and we report results as stated in the original papers where possible.

\textbf{SYMSOL Results}
The performance on the SYMSOL dataset is reported in Table \ref{Main:Table:SYMSOL}.  Our method achieves highest average log likelihood on SYMSOL I.  Importantly, we observe a significant improvement over \citet{Klee_2023} on all objects, which indicates that our induction layer is more effective than its hand-designed orthographic projection.  On SYMSOL II, our method slightly underperforms \citet{Murphy_2022}, which has much higher expressivity on the output since it is an implicit model.  However, we demonstrate that our approach, which preserves the symmetry present in the images, is better with less data, as shown in Table \ref{Main:Table:subsampled_SYMSOL}.

\begin{table}[H]
\centering
\caption{Average log likelihood (the higher the better $\uparrow$) on SYMSOL I \& II. Per \cite{Murphy_2022}, a single model is trained on all classes in SYMSOL I and a separate model is trained on each class in SYMSOL II.}
\small
\hspace*{-0.2cm}
\begin{tabular}{@{}ll@{\hskip 0.5cm}rrrrr@{\hskip 0.9cm}rrrr@{}}
\toprule
& \multicolumn{6}{c}{SYMSOL I ($\uparrow$)}       & \multicolumn{4}{c}{ SYMSOL II ($\uparrow$)} \\ 
& \textit{avg} & \textit{cone} & \textit{cyl} & \textit{tet} & \textit{cube} & \textit{ico} & \textit{avg}   & \textit{sphX}  & \textit{cylO}  & \textit{tetX}  \\ \midrule
\citet{Deng_2020} &   -1.48  &   0.16   &    -0.95 &    0.27 &    -4.44  &    -2.45 &       2.57 &    1.12   &   2.99    &   3.61    \\
\citet{Prokudin_2018} &   -1.87   &  -3.34    &   -1.28  &  -1.86   &    -0.50  &   -2.39  &     0.48  &   -4.19    &    4.16   &    1.48   \\
\citet{Gilitschenski_2020} & -0.43     & 3.84     &  0.88   &   -2.29  &   -2.29   &    -2.29 &    3.70   &   3.32    &      4.88 &  2.90     \\
\citet{Murphy_2022} & 4.10    &  4.45    &  \textbf{4.26}   &   5.70  &  4.81    &  1.28   &  \textbf{7.57}     &   \textbf{7.30}    &    \textbf{6.91}   &   \textbf{8.49}    \\
\citet{Klee_2023} &  3.41   &   3.75   &  3.10   &  4.78   &   3.27   &    2.15  &        4.84 &   3.74    &   5.18   &  5.61     \\
\textbf{Ours} & \textbf{5.11}   &   \textbf{4.91}   &   4.22  &   \textbf{6.10}  &   \textbf{5.73}  &   \textbf{4.69}  &  6.20    &   7.10    &   6.01 &     5.62 \\  \bottomrule
\end{tabular}
\label{Main:Table:SYMSOL}
\end{table}

\begin{table}[H]
\centering
\caption{Average log likelihood on SYMSOL I \& II with 10\% of training data.}
\small
\begin{tabular}{@{}ll@{\hskip 0.4cm}rrrrr@{\hskip 0.7cm}rrrr@{}}
\toprule
& \multicolumn{6}{c}{10\% SYMSOL I ($\uparrow$)}        & \multicolumn{4}{c}{10\% SYMSOL II ($\uparrow$)} \\ 
& \textit{avg} & \textit{cone} & \textit{cyl} & \textit{tet} & \textit{cube} & \textit{ico} & \textit{avg}   & \textit{sphX}  & \textit{cylO}  & \textit{tetX}  \\ \midrule
\citet{Murphy_2022} & -7.94    &  -1.51  &  -2.92  &  -6.90  &  -10.04    &  -18.34   &  -0.73     &   -2.51    &    2.02   &   -1.70    \\
\citet{Klee_2023} &  2.98   &   3.51   &  2.88   & \textbf{3.62}   &  2.94   &    \textbf{1.94}  &    \textbf{3.61} &   \textbf{3.12}   &   \textbf{3.87}   &  3.84     \\
\textbf{Ours} &  \textbf{3.01}  &   \textbf{3.63}   &   \textbf{3.01}  &   3.53  &  \textbf{3.02}  &   1.91  & 3.54   &   2.88    & 3.71 &  \textbf{4.04} \\  \bottomrule
\end{tabular}
\label{Main:Table:subsampled_SYMSOL}
\end{table}

\textbf{PASCAL3D+ Results}
Our method achieves state-of-the-art performance on PASCAL3D+ with an average median rotation error of 9.2 degrees, as reported in Table \ref{Main:Table:pascal}.  Even though object symmetries are consistently disambiguated in the labels, modeling pose as a distribution is beneficial for noisy images where there is insufficient information to resolve the pose exactly.  Because our induction layer produces representations on the Fourier basis of $SO(3)$, it naturally allows for capturing this uncertainty as a distribution over $\SO(3)$.  While both our method and \cite{Klee_2023} leverage $\SO(3)$ equivariant layers to improve generalization, we find our method achieves higher performance.  We believe our induction layer is more robust to variations in how the images are rendered/captured, which is important for PASCAL3D+, since the data is aggregated from many sources.  Moreover, our method does not restrict features to the hemisphere, which could be beneficial for objects, like bikes and chairs, that do not fully self-occlude their backsides.

\begin{table}[H]
\centering
\caption{Rotation prediction on PASCAL3D+. First column is the average over all categories. }
\hspace*{-0.7cm}   
\scriptsize
\begin{tabular}{ p{2.3cm}p{0.5cm}p{0.5cm}p{0.5cm}p{0.5cm}p{0.5cm}p{0.5cm}p{0.5cm}p{0.5cm}p{0.5cm}p{0.5cm}p{0.5cm}p{0.5cm}p{0.5cm}p{0.5cm} }
\toprule
& \multicolumn{13}{c}{Median rotation error in degrees ($\downarrow$)} \\
& \textit{avg} & \textit{plane} & \textit{bike} & \textit{boat} & \textit{bottle} & \textit{bus} & \textit{car} & \textit{chair} & \textit{table} & \textit{mbike} &\textit{sofa} & \textit{train} & \textit{tv}   \\
\midrule
\citet{Mohlin_2021}  & 11.5 & 10.1 & 15.6 & 24.3 & 7.8 & 3.3 & 5.3 & 13.5 & 12.5 & 12.9 & 13.8 & 7.4 & 11.7 \\
\citet{Prokudin_2018} & 12.2 & 9.7 & 15.5 & 45.6 & \textbf{5.4} & 2.9 & \textbf{4.5} & 13.1 & 12.6 & 11.8 & 9.1 & \textbf{4.3} & 12.0  \\ 
\citet{Tulsiani_2015} & 13.6 & 13.8 & 17.7 & 21.3& 12.9 & 5.8 & 9.1 & 14.8 & 15.2 & 14.7 &  13.7 &  8.7 & 15.4  \\ 
\citet{Mahendran_2018} & 10.1 & \textbf{8.5} & 14.8 & 20.5 & 7.0 &  3.1&  5.1 & 9.5 & 11.3 & 14.2 & 10.2 & 5.6  & 11.7  \\ 
\citet{Liao_2019} & 13.0 & 13.0 & 16.4 & 29.1 & 10.3 & 4.8 & 6.8 &  11.6 & 12.0 & 17.1 & 12.3 & 8.6 & 14.3   \\ 
\citet{Murphy_2022} & 10.3 & 10.8 & 12.9 & 23.4 & 8.8 & 3.4 & 5.3 & 10.0 & 7.3 & 13.6 & 9.5 & 6.4 & 12.3  \\ 
\citet{Klee_2023} & 9.8 & 9.2 & 12.7 & 21.7 & 7.4 & 3.3 & 4.9 & 9.5 & 9.3 & \textbf{11.5} & 10.5 & 7.2 & 10.6  \\ 
\citet{Yin_2023}  & 9.4 & 8.6 & \textbf{11.7} & 21.8 & 6.9 & \textbf{2.8} & 4.8 & \textbf{7.9} & 9.1  & 12.2 & \textbf{8.1}  & 6.9 & 11.6  \\ 
\textbf{Ours (ResNet-50)} & 10.2 & 9.2 & 13.1 & 30.6 & 6.7 & 3.1 & 4.8 & 8.7 & 5.4 & 11.6 & 11.0 & 5.8 & 10.6  \\
\textbf{Ours} & \textbf{9.2} & 9.3 & 12.6 & \textbf{17.0} & 8.0 & 3.0 & \textbf{4.5} & 9.4 & \textbf{6.7} & 11.9 & 12.1 & 6.9 & \textbf{9.9} \\
\bottomrule
\end{tabular}
\label{Main:Table:pascal}
\end{table}

\section{Conclusion}\label{sec:Conclusion}

In conclusion, we have argued that any network that learns a three-dimensional model of the world from two dimensional images must satisfy certain consistency properties. We have shown how these consistency properties translate into an $SO(2)$-equivariance constraint. Using the induced representation we have derived an explicit form for any neural networks that satisfies said consistency constraint. We have proposed an \emph{induction/restriction layer}, which is learnable network layer that satisfies the derived consistency equation. We have shown that the induction layer satisfies both a completeness property and universal property and, up to isomorphism, is unique. Furthermore, we have shown that the methods of \cite{Klee_2023,Klee_2022,Esteves_2019} can be realized as specific instances of the induction layer.

The framework that we have developed is general and can be applied to other computer vision problems with different symmetries. For example, as was noted in \cite{Cesa_2022}, the cryogenic electronic microscopy orientation estimation problem has a latent $SO(3)$ symmetry but a manifest $SO(2) \times \mathbb{Z}_{2} \cong O(2) $ (as opposed to an $SO(2)$) symmetry. With a slight modification \ref{Appendix:Section:Generalization to Arbitrary Homogeneous Spaces}, the results presented in the main text allow for the construction of an induction layer that leverages this observation.

\paragraph{Future Work}
In many structure from motion tasks, one has access to multiple images of the same object, taken at either known or unknown vantage points. Our work considers only single view pose-estimation. A natural generalization of our work is to include stereo measurements into the induced/restricted representation framework. \cite{Biza_2023,Sajjadi2022object} use transformer architectures to learn models of three dimensional objects from two-dimensional images. Another natural extension of our work would be to include transformers into the framework presented here, which only applies to convolutional networks.


In deep learning, we often wish to construct a neural network that respects a latent symmetry $G$ that does not have action on the input data space. We have show how the induced representation can be used to construct latent $G$-equivariant neural networks. Our work provides a systematic way to construct neural architectures that accept any format of inputs and respect the latent symmetries of the problem.

\begin{ack}
Owen Howell thanks Dr. Thomas Sayre-Maccord for logistics help. Owen Howell further thanks Liam Pavlovic and Dr. David Rosen for useful discussions. Owen Howell acknowledges the National Science Foundation Graduate Research Fellowship Program (NSF-GRFP) for financial support.
\end{ack}


{
\small
\bibliography{refs}

\begin{thebibliography}{64}
\providecommand{\natexlab}[1]{#1}
\providecommand{\url}[1]{\texttt{#1}}
\expandafter\ifx\csname urlstyle\endcsname\relax
  \providecommand{\doi}[1]{doi: #1}\else
  \providecommand{\doi}{doi: \begingroup \urlstyle{rm}\Url}\fi

\bibitem[Marr(2010)]{marr2010vision}
David Marr.
\newblock \emph{Vision: A computational investigation into the human
  representation and processing of visual information}.
\newblock MIT press, 2010.

\bibitem[Hartley and Zisserman(2004)]{Hartley_2004}
Richard Hartley and Andrew Zisserman.
\newblock \emph{Multiple View Geometry in Computer Vision}.
\newblock Cambridge University Press, 2 edition, 2004.
\newblock \doi{10.1017/CBO9780511811685}.

\bibitem[Ozyesil et~al.(2017)Ozyesil, Voroninski, Basri, and
  Singer]{Ozyesil_2017}
Onur Ozyesil, Vladislav Voroninski, Ronen Basri, and Amit Singer.
\newblock A survey of structure from motion, 2017.
\newblock URL \url{https://arxiv.org/abs/1701.08493}.

\bibitem[Bronstein et~al.(2021)Bronstein, Bruna, Cohen, and
  Veličković]{Bronstein_2021}
Michael~M. Bronstein, Joan Bruna, Taco Cohen, and Petar Veličković.
\newblock Geometric deep learning: Grids, groups, graphs, geodesics, and
  gauges, 2021.
\newblock URL \url{https://arxiv.org/abs/2104.13478}.

\bibitem[Cohen and Welling(2016{\natexlab{a}})]{Cohen_2016_II}
Taco~S. Cohen and Max Welling.
\newblock Steerable cnns.
\newblock \emph{axriv}, 2016{\natexlab{a}}.
\newblock \doi{10.48550/ARXIV.1612.08498}.
\newblock URL \url{https://arxiv.org/abs/1612.08498}.

\bibitem[Kondor and Trivedi(2018)]{Kondor_2018}
Risi Kondor and Shubhendu Trivedi.
\newblock On the generalization of equivariance and convolution in neural
  networks to the action of compact groups, 2018.
\newblock URL \url{https://arxiv.org/abs/1802.03690}.

\bibitem[Cohen et~al.(2018{\natexlab{a}})Cohen, Geiger, and Weiler]{Cohen_2018}
S.~Cohen, Mario Geiger, and Maurice Weiler.
\newblock Intertwiners between induced representations (with applications to
  the theory of equivariant neural networks), 2018{\natexlab{a}}.
\newblock URL \url{https://arxiv.org/abs/1803.10743}.

\bibitem[Lang and Weiler(2020)]{Lang_2020}
Leon Lang and Maurice Weiler.
\newblock A wigner-eckart theorem for group equivariant convolution kernels,
  2020.
\newblock URL \url{https://arxiv.org/abs/2010.10952}.

\bibitem[Cohen and Welling(2016{\natexlab{b}})]{Cohen_2016}
Taco~S. Cohen and Max Welling.
\newblock Group equivariant convolutional networks.
\newblock \emph{axriv}, 2016{\natexlab{b}}.
\newblock \doi{10.48550/ARXIV.1602.07576}.
\newblock URL \url{https://arxiv.org/abs/1602.07576}.

\bibitem[Klee et~al.(2022)Klee, Biza, Platt, and Walters]{Klee_2022}
David Klee, Ondrej Biza, Robert Platt, and Robin Walters.
\newblock Image to icosahedral projection for $\mathrm{SO}(3)$ object reasoning
  from single-view images, 2022.
\newblock URL \url{https://arxiv.org/abs/2207.08925}.

\bibitem[Esteves et~al.(2019{\natexlab{a}})Esteves, Xu, Allen-Blanchette, and
  Daniilidis]{Esteves_2019}
Carlos Esteves, Yinshuang Xu, Christine Allen-Blanchette, and Kostas
  Daniilidis.
\newblock Equivariant multi-view networks, 2019{\natexlab{a}}.
\newblock URL \url{https://arxiv.org/abs/1904.00993}.

\bibitem[Klee et~al.(2023)Klee, Biza, Platt, and Walters]{Klee_2023}
David Klee, Ondrej Biza, Robert Platt, and Robin Walters.
\newblock Image to sphere: Learning equivariant features for efficient pose
  prediction.
\newblock In \emph{International Conference on Learning Representations}, 2023.
\newblock URL \url{https://openreview.net/forum?id=_2bDpAtr7PI}.

\bibitem[Xiang et~al.(2014)Xiang, Mottaghi, and Savarese]{Xiang_2014}
Yu~Xiang, Roozbeh Mottaghi, and Silvio Savarese.
\newblock Beyond pascal: A benchmark for 3d object detection in the wild.
\newblock In \emph{IEEE Winter Conference on Applications of Computer Vision},
  pages 75--82, 2014.
\newblock \doi{10.1109/WACV.2014.6836101}.

\bibitem[Murphy et~al.(2022)Murphy, Esteves, Jampani, Ramalingam, and
  Makadia]{Murphy_2022}
Kieran Murphy, Carlos Esteves, Varun Jampani, Srikumar Ramalingam, and Ameesh
  Makadia.
\newblock Implicit-pdf: Non-parametric representation of probability
  distributions on the rotation manifold, 2022.

\bibitem[LeCun et~al.(1995)LeCun, Bengio, et~al.]{lecun1995convolutional}
Yann LeCun, Yoshua Bengio, et~al.
\newblock Convolutional networks for images, speech, and time series.
\newblock \emph{The handbook of brain theory and neural networks},
  3361\penalty0 (10):\penalty0 1995, 1995.

\bibitem[Shaw et~al.(2018)Shaw, Uszkoreit, and Vaswani]{shaw2018self}
Peter Shaw, Jakob Uszkoreit, and Ashish Vaswani.
\newblock Self-attention with relative position representations.
\newblock \emph{arXiv preprint arXiv:1803.02155}, 2018.

\bibitem[Qi et~al.(2017)Qi, Su, Mo, and Guibas]{qi2017pointnet}
Charles~R Qi, Hao Su, Kaichun Mo, and Leonidas~J Guibas.
\newblock Pointnet: Deep learning on point sets for 3d classification and
  segmentation.
\newblock In \emph{Proceedings of the IEEE conference on computer vision and
  pattern recognition}, pages 652--660, 2017.

\bibitem[Thomas et~al.(2018)Thomas, Smidt, Kearnes, Yang, Li, Kohlhoff, and
  Riley]{Thomas_2018}
Nathaniel Thomas, Tess Smidt, Steven Kearnes, Lusann Yang, Li~Li, Kai Kohlhoff,
  and Patrick Riley.
\newblock Tensor field networks: Rotation- and translation-equivariant neural
  networks for 3d point clouds, 2018.

\bibitem[Wang et~al.(2022)Wang, Park, Sortur, Wong, Walters, and
  Platt]{Wang_2022}
Dian Wang, Jung~Yeon Park, Neel Sortur, Lawson L.~S. Wong, Robin Walters, and
  Robert Platt.
\newblock The surprising effectiveness of equivariant models in domains with
  latent symmetry, 2022.
\newblock URL \url{https://arxiv.org/abs/2211.09231}.

\bibitem[Weiler and Cesa(2021)]{Weiler_2021}
Maurice Weiler and Gabriele Cesa.
\newblock General $e(2)$-equivariant steerable cnns, 2021.

\bibitem[Weiler et~al.(2018{\natexlab{a}})Weiler, Geiger, Welling, Boomsma, and
  Cohen]{Weiler_2018}
Maurice Weiler, Mario Geiger, Max Welling, Wouter Boomsma, and Taco Cohen.
\newblock 3d steerable cnns: Learning rotationally equivariant features in
  volumetric data, 2018{\natexlab{a}}.

\bibitem[Cohen et~al.(2018{\natexlab{b}})Cohen, Geiger, Koehler, and
  Welling]{Cohen2018_spherical}
Taco~S. Cohen, Mario Geiger, Jonas Koehler, and Max Welling.
\newblock Spherical cnns, 2018{\natexlab{b}}.

\bibitem[Falorsi et~al.(2018)Falorsi, de~Haan, Davidson, Cao, Weiler, Forré,
  and Cohen]{Falorsi_2018}
Luca Falorsi, Pim de~Haan, Tim~R. Davidson, Nicola~De Cao, Maurice Weiler,
  Patrick Forré, and Taco~S. Cohen.
\newblock Explorations in homeomorphic variational auto-encoding, 2018.

\bibitem[Park et~al.(2022)Park, Biza, Zhao, van~de Meent, and
  Walters]{park2022learning}
Jung~Yeon Park, Ondrej Biza, Linfeng Zhao, Jan~Willem van~de Meent, and Robin
  Walters.
\newblock Learning symmetric embeddings for equivariant world models.
\newblock \emph{arXiv preprint arXiv:2204.11371}, 2022.

\bibitem[Esteves et~al.(2019{\natexlab{b}})Esteves, Sud, Luo, Daniilidis, and
  Makadia]{esteves2019cross}
Carlos Esteves, Avneesh Sud, Zhengyi Luo, Kostas Daniilidis, and Ameesh
  Makadia.
\newblock Cross-domain 3d equivariant image embeddings.
\newblock In \emph{International Conference on Machine Learning}, pages
  1812--1822. PMLR, 2019{\natexlab{b}}.

\bibitem[Geiger et~al.(2013)Geiger, Lenz, Stiller, and
  Urtasun]{geiger2013vision}
Andreas Geiger, Philip Lenz, Christoph Stiller, and Raquel Urtasun.
\newblock Vision meets robotics: The kitti dataset.
\newblock \emph{The International Journal of Robotics Research}, 32\penalty0
  (11):\penalty0 1231--1237, 2013.

\bibitem[Xiang et~al.(2017)Xiang, Schmidt, Narayanan, and
  Fox]{xiang2017posecnn}
Yu~Xiang, Tanner Schmidt, Venkatraman Narayanan, and Dieter Fox.
\newblock Posecnn: A convolutional neural network for 6d object pose estimation
  in cluttered scenes.
\newblock \emph{arXiv preprint arXiv:1711.00199}, 2017.

\bibitem[Zhong et~al.(2020)Zhong, Bepler, Davis, and Berger]{Zhong_2020}
Ellen~D. Zhong, Tristan Bepler, Joseph~H. Davis, and Bonnie Berger.
\newblock Reconstructing continuous distributions of 3d protein structure from
  cryo-em images, 2020.

\bibitem[Tulsiani and Malik(2015)]{Tulsiani_2015}
Shubham Tulsiani and Jitendra Malik.
\newblock Viewpoints and keypoints, 2015.

\bibitem[Mahendran et~al.(2018)Mahendran, Ali, and Vidal]{Mahendran_2018}
Siddharth Mahendran, Haider Ali, and Rene Vidal.
\newblock A mixed classification-regression framework for 3d pose estimation
  from 2d images, 2018.

\bibitem[Zhou et~al.(2020)Zhou, Barnes, Lu, Yang, and Li]{Zhou_2020_Cont}
Yi~Zhou, Connelly Barnes, Jingwan Lu, Jimei Yang, and Hao Li.
\newblock On the continuity of rotation representations in neural networks,
  2020.

\bibitem[Brégier(2021)]{Bregier_2021}
Romain Brégier.
\newblock Deep regression on manifolds: A 3d rotation case study, 2021.

\bibitem[Liao et~al.(2019)Liao, Gavves, and Snoek]{Liao_2019}
Shuai Liao, Efstratios Gavves, and Cees G.~M. Snoek.
\newblock Spherical regression: Learning viewpoints, surface normals and 3d
  rotations on n-spheres, 2019.

\bibitem[Deng et~al.(2020)Deng, Bui, Navab, Guibas, Ilic, and
  Birdal]{Deng_2020}
Haowen Deng, Mai Bui, Nassir Navab, Leonidas Guibas, Slobodan Ilic, and Tolga
  Birdal.
\newblock Deep bingham networks: Dealing with uncertainty and ambiguity in pose
  estimation, 2020.

\bibitem[Prokudin et~al.(2018)Prokudin, Gehler, and Nowozin]{Prokudin_2018}
Sergey Prokudin, Peter Gehler, and Sebastian Nowozin.
\newblock Deep directional statistics: Pose estimation with uncertainty
  quantification, 2018.

\bibitem[Yin et~al.(2023)Yin, Wang, Wang, and Chen]{Yin_2023}
Yingda Yin, Yang Wang, He~Wang, and Baoquan Chen.
\newblock A laplace-inspired distribution on {SO}(3) for probabilistic rotation
  estimation.
\newblock In \emph{The Eleventh International Conference on Learning
  Representations}, 2023.
\newblock URL \url{https://openreview.net/forum?id=Mvetq8DO05O}.

\bibitem[Ceccherini-Silberstein et~al.(2008)Ceccherini-Silberstein, Scarabotti,
  and Tolli]{Ceccherini_2008}
Tullio Ceccherini-Silberstein, Fabio Scarabotti, and Filippo Tolli.
\newblock \emph{Harmonic Analysis on Finite Groups: Representation Theory,
  Gelfand Pairs and Markov Chains}.
\newblock Cambridge Studies in Advanced Mathematics. Cambridge University
  Press, 2008.
\newblock \doi{10.1017/CBO9780511619823}.

\bibitem[Weiler et~al.(2018{\natexlab{b}})Weiler, Hamprecht, and
  Storath]{Weiler_2018_II}
Maurice Weiler, Fred~A. Hamprecht, and Martin Storath.
\newblock Learning steerable filters for rotation equivariant cnns,
  2018{\natexlab{b}}.

\bibitem[Franzen and Wand(2021)]{Franzen_2021}
Daniel Franzen and Michael Wand.
\newblock Nonlinearities in steerable so(2)-equivariant cnns, 2021.

\bibitem[de~Haan et~al.(2021)de~Haan, Weiler, Cohen, and Welling]{Dehaan_2021}
Pim de~Haan, Maurice Weiler, Taco Cohen, and Max Welling.
\newblock Gauge equivariant mesh cnns: Anisotropic convolutions on geometric
  graphs, 2021.

\bibitem[Poulenard and Guibas(2021)]{Poulenard_2021}
Adrien Poulenard and Leonidas~J. Guibas.
\newblock A functional approach to rotation equivariant non-linearities for
  tensor field networks.
\newblock In \emph{2021 IEEE/CVF Conference on Computer Vision and Pattern
  Recognition (CVPR)}, pages 13169--13178, 2021.
\newblock \doi{10.1109/CVPR46437.2021.01297}.

\bibitem[Xu et~al.(2022)Xu, Lei, Dobriban, and Daniilidis]{Xu_2022}
Yinshuang Xu, Jiahui Lei, Edgar Dobriban, and Kostas Daniilidis.
\newblock Unified fourier-based kernel and nonlinearity design for equivariant
  networks on homogeneous spaces, 2022.

\bibitem[Geiger and Smidt(2022)]{Geiger_2022}
Mario Geiger and Tess Smidt.
\newblock e3nn: Euclidean neural networks, 2022.

\bibitem[Leinster(2016)]{Leinster_2016}
Tom Leinster.
\newblock Basic category theory, 2016.

\bibitem[Su et~al.(2015)Su, Qi, Li, and Guibas]{Su_2015}
Hao Su, Charles~R. Qi, Yangyan Li, and Leonidas Guibas.
\newblock Render for cnn: Viewpoint estimation in images using cnns trained
  with rendered 3d model views, 2015.

\bibitem[Paszke et~al.(2019)Paszke, Gross, Massa, Lerer, Bradbury, Chanan,
  Killeen, Lin, Gimelshein, Antiga, Desmaison, Köpf, Yang, DeVito, Raison,
  Tejani, Chilamkurthy, Steiner, Fang, Bai, and Chintala]{Paszke_2019}
Adam Paszke, Sam Gross, Francisco Massa, Adam Lerer, James Bradbury, Gregory
  Chanan, Trevor Killeen, Zeming Lin, Natalia Gimelshein, Luca Antiga, Alban
  Desmaison, Andreas Köpf, Edward Yang, Zach DeVito, Martin Raison, Alykhan
  Tejani, Sasank Chilamkurthy, Benoit Steiner, Lu~Fang, Junjie Bai, and Soumith
  Chintala.
\newblock Pytorch: An imperative style, high-performance deep learning library,
  2019.

\bibitem[Mohlin et~al.(2021)Mohlin, Bianchi, and Sullivan]{Mohlin_2021}
David Mohlin, Gerald Bianchi, and Josephine Sullivan.
\newblock Probabilistic regression with huber distributions, 2021.

\bibitem[Gilitschenski et~al.(2020)Gilitschenski, Sahoo, Schwarting, Amini,
  Karaman, and Rus]{Gilitschenski_2020}
Igor Gilitschenski, Roshni Sahoo, Wilko Schwarting, Alexander Amini, Sertac
  Karaman, and Daniela Rus.
\newblock Deep orientation uncertainty learning based on a bingham loss.
\newblock In \emph{International Conference on Learning Representations}, 2020.
\newblock URL \url{https://openreview.net/forum?id=ryloogSKDS}.

\bibitem[Cesa et~al.(2022)Cesa, Behboodi, Cohen, and Welling]{Cesa_2022}
Gabriele Cesa, Arash Behboodi, Taco Cohen, and Max Welling.
\newblock On the symmetries of the synchronization problem in cryo-{EM}:
  Multi-frequency vector diffusion maps on the projective plane.
\newblock In Alice~H. Oh, Alekh Agarwal, Danielle Belgrave, and Kyunghyun Cho,
  editors, \emph{Advances in Neural Information Processing Systems}, 2022.
\newblock URL \url{https://openreview.net/forum?id=owDcdLGgEm}.

\bibitem[Biza et~al.(2023)Biza, van Steenkiste, Sajjadi, Elsayed, Mahendran,
  and Kipf]{Biza_2023}
Ondrej Biza, Sjoerd van Steenkiste, Mehdi S.~M. Sajjadi, Gamaleldin~F. Elsayed,
  Aravindh Mahendran, and Thomas Kipf.
\newblock Invariant slot attention: Object discovery with slot-centric
  reference frames, 2023.

\bibitem[Sajjadi et~al.(2022)Sajjadi, Duckworth, Mahendran, van Steenkiste,
  Pavetić, Lučić, Guibas, Greff, and Kipf]{Sajjadi2022object}
Mehdi S.~M. Sajjadi, Daniel Duckworth, Aravindh Mahendran, Sjoerd van
  Steenkiste, Filip Pavetić, Mario Lučić, Leonidas~J. Guibas, Klaus Greff,
  and Thomas Kipf.
\newblock Object scene representation transformer, 2022.

\bibitem[Zee(2016)]{Zee_2016}
A.~Zee.
\newblock \emph{Group Theory in a Nutshell for Physicists}.
\newblock In a Nutshell. Princeton University Press, 2016.
\newblock ISBN 9780691162690.
\newblock URL \url{https://books.google.com/books?id=FWkujgEACAAJ}.

\bibitem[Serre(2005)]{Serre_2005}
J.~P. Serre.
\newblock Groupes finis, 2005.
\newblock URL \url{https://arxiv.org/abs/math/0503154}.

\bibitem[Ceccherini-Silberstein et~al.(2018)Ceccherini-Silberstein, Scarabotti,
  and Tolli]{Ceccherini_2018}
Tullio Ceccherini-Silberstein, Fabio Scarabotti, and Filippo Tolli.
\newblock \emph{Induced representations and Mackey theory}, page 399–425.
\newblock Cambridge Studies in Advanced Mathematics. Cambridge University
  Press, 2018.
\newblock \doi{10.1017/9781316856383.012}.

\bibitem[Wu et~al.(2015)Wu, Song, Khosla, Yu, Zhang, Tang, and Xiao]{Wu_2015}
Zhirong Wu, Shuran Song, Aditya Khosla, Fisher Yu, Linguang Zhang, Xiaoou Tang,
  and Jianxiong Xiao.
\newblock 3d shapenets: A deep representation for volumetric shapes, 2015.

\bibitem[Lin et~al.(2021)Lin, Li, Chen, Lu, and Jia]{Lin_2021}
Jiehong Lin, Hongyang Li, Ke~Chen, Jiangbo Lu, and Kui Jia.
\newblock Sparse steerable convolutions: An efficient learning of
  {SE}(3)-equivariant features for estimation and tracking of object poses in
  3d space.
\newblock In A.~Beygelzimer, Y.~Dauphin, P.~Liang, and J.~Wortman Vaughan,
  editors, \emph{Advances in Neural Information Processing Systems}, 2021.
\newblock URL \url{https://openreview.net/forum?id=Fa-w-10s7YQ}.

\bibitem[Jenner and Weiler(2022)]{Jenner_2022}
Erik Jenner and Maurice Weiler.
\newblock Steerable partial differential operators for equivariant neural
  networks, 2022.

\bibitem[Spencer et~al.(2022)Spencer, Russell, Hadfield, and
  Bowden]{Spencer_2022_Deconstructing}
Jaime Spencer, Chris Russell, Simon Hadfield, and Richard Bowden.
\newblock Deconstructing self-supervised monocular reconstruction: The design
  decisions that matter, 2022.

\bibitem[Saxena et~al.(2023)Saxena, Kar, Norouzi, and
  Fleet]{Saxena_2023_Monocular}
Saurabh Saxena, Abhishek Kar, Mohammad Norouzi, and David~J. Fleet.
\newblock Monocular depth estimation using diffusion models, 2023.

\bibitem[Liu et~al.(2019)Liu, Sinha, Ishii, Hager, Reiter, Taylor, and
  Unberath]{Liu_2019_Dense}
Xingtong Liu, Ayushi Sinha, Masaru Ishii, Gregory~D. Hager, Austin Reiter,
  Russell~H. Taylor, and Mathias Unberath.
\newblock Dense depth estimation in monocular endoscopy with self-supervised
  learning methods, 2019.

\bibitem[Batlle et~al.(2022)Batlle, Montiel, and Tardos]{Batlle_2022}
Victor~M. Batlle, J.~M.~M. Montiel, and Juan~D. Tardos.
\newblock Photometric single-view dense 3d reconstruction in endoscopy, 2022.

\bibitem[Fonder et~al.(2022)Fonder, Ernst, and
  Droogenbroeck]{Fonder_2022_M4Depth}
Michaël Fonder, Damien Ernst, and Marc~Van Droogenbroeck.
\newblock M4depth: Monocular depth estimation for autonomous vehicles in unseen
  environments, 2022.

\bibitem[Passaro and Zitnick(2023)]{Passaro_2023_Reducing}
Saro Passaro and C.~Lawrence Zitnick.
\newblock Reducing so(3) convolutions to so(2) for efficient equivariant gnns,
  2023.

\bibitem[Hornik et~al.(1989)Hornik, Stinchcombe, and White]{Hornik_1989}
Kurt Hornik, Maxwell Stinchcombe, and Halbert White.
\newblock Multilayer feedforward networks are universal approximators.
\newblock \emph{Neural Networks}, 2\penalty0 (5):\penalty0 359--366, 1989.
\newblock ISSN 0893-6080.
\newblock \doi{https://doi.org/10.1016/0893-6080(89)90020-8}.
\newblock URL
  \url{https://www.sciencedirect.com/science/article/pii/0893608089900208}.

\end{thebibliography}
}

\newpage
\appendix
\section{Notation and Preliminaries}\label{Appendix:Section:Notation and Preliminaries}

We establish some notation and review some elements of representation theory. For a comprehensive review of representation theory, please see \cite{Zee_2016,Serre_2005}. The identity element of any group $G$ will be denoted as $e$. A subgroup $H$ of $G$ will be denoted as $H \subseteq G$. We will always work over the field $\mathbb{R}$ unless otherwise specified.

\subsubsection{Group Actions}\label{Section:Group Actions}

Let $\Omega$ be a set. A group action $\Phi$ of $G$ on $\Omega$ is a map $\Phi : G \times \Omega \rightarrow \Omega$ which satisfies 
\begin{align}\label{Equation:Group_Action}
    &\text{Identity: } \forall \omega \in \Omega, \quad \Phi(e , \omega )  =  \omega \\
    &\nonumber \text{Compositionality: }\forall g_{1},g_{2} \in G,\enspace \forall \omega \in \Omega, \quad  \Phi( g_{1}g_{2} , \omega  ) = \Phi(g_{1} , \Phi(g_{2}, \omega ))
\end{align}
We will often suppress the $\Phi$ function and write $\Phi( g , \omega ) = g \cdot \omega$.

\begin{center}\label{Diagram:G-Equivarient_Map}
    \begin{tikzcd}\centering
        &\Omega \arrow{d}{\Phi(g,\cdot)} \arrow{r}{ \Psi } & \Omega' \arrow{d}{ \Phi'(g, \cdot ) }  \\
        & \Omega \arrow{r}{\Psi }  & \Omega'
    \end{tikzcd}
    \captionof{figure}{Commutative Diagram For $G$-equivariant function: Let $\Phi(g, \cdot ): G \times \Omega \rightarrow \Omega$ denote the action of $G$ on $\Omega$. Let $\Phi'(g, \cdot ): G \times \Omega' \rightarrow \Omega'$ denote the action of $G$ on $\Omega'$. The map $\Psi: \Omega \rightarrow \Omega'$ is $G$-equivariant if and only if the following diagram is commutative for all $g\in G$.    }
\end{center}

Let $G$ have group action $\Phi$ on $\Omega$ and group action $\Phi'$ on $\Omega'$. A mapping $\Psi : \Omega \rightarrow \Omega'$ is said to be $G$-equivariant if and only if
\begin{align}\label{Equation:G_Equivariece_Def}
    \forall g \in G, \forall \omega \in \Omega, \quad	 \Psi(  \Phi( g , \omega )  ) = \Phi'( g , \Psi(\omega) )
\end{align}
Diagrammatically, $\Psi$ is $G$-equivariant if and only if the diagram \ref{Diagram:G-Equivarient_Map} is commutative. 

\subsubsection{Induced and Restricted Representations}
Let $V$ be a vector space over $\mathbb{C}$. A \emph{representation} $(\rho , V)$ of $G$ is a map $\rho : G \rightarrow \Hom[V,V]$ such that 
\begin{align*}
\forall g , g' \in G, \enspace \forall v\in V \quad   \rho( g \cdot g' )v =  \rho( g ) \cdot \rho(  g' )v
\end{align*}

\paragraph{Restricted Representation}
Let $H \subseteq G$. Let $(\rho , V)$ be a representation of $G$. The restricted representation of $(\rho , V)$ from $G$ to $H$ is denoted as $\Res_{H}^{G}[ (\rho , V) ]$. Intuitively, $\Res_{H}^{G}[ (\rho , V) ]$ can be viewed as $(\rho , V)$ evaluated on the subgroup $H$. Specifically, 
\begin{align}
\forall v\in V, \quad \Res_{H}^{G}[ \rho ](h) v = \rho(h) v
\end{align}
Note that the restricted representation and the original representation both live on the same vector space $V$.

\paragraph{Induced Representation}

The induction representation is a way to construct representations of a larger group $G$ out of representations of a subgroup $H \subseteq G$. Let $(\rho , V)$ be a representation of $H$. The induced representation of $(\rho , V)$ from $H$ to $G$ is denoted as $\Ind_{H}^{G}[ (\rho , V) ]$. Define the space of functions
\begin{align*}
\mathcal{F} = \{ \enspace f \enspace | \enspace f : G \rightarrow V, \enspace \forall h\in H, \enspace f(gh) = \rho(h^{-1}) f(g) \enspace \}
\end{align*}
Then the induced representation is defined as $ ( \pi , \mathcal{F}  ) = \Ind_{H}^{G}[ (\rho , V) ]$
where the induced action $\pi$ acts on the function space $\mathcal{F}$ via
\begin{align*}
    \forall g,g' \in G, \enspace \forall f \in \mathcal{F} \quad (\pi(g)\cdot f)(g') = f(g^{-1} g')
\end{align*}

\paragraph{Induced Representation for Finite Groups}\label{Appendix:Section:Induced Representation for Finite Groups}
There is also an equivalent definition of the induced representation for finite groups that is slightly more intuitive \cite{Ceccherini_2018}. Let $G$ be a group and let $H \subseteq G$. The set of left cosets of $G/H$ form a partition of $G$ so that
\begin{align*}
    G = \bigcup_{i=1}^{|G/H|} g_{i} H
\end{align*}
where $\{g_{i}\}_{i=1}^{|G/H|}$ are a set of representatives of each unique left coset. Note that the choice of left coset representatives is not unique. Now, left multiplication by the element $g\in G$ is an automorphism of $G$. Left multiplication by $g \in G$ must thus permute left cosets of $G / H$ so that
\begin{align*}
    \forall g \in G, \quad   g \cdot g_{i} = g_{j_{g}(i)} h_{i}(g)
\end{align*}
where $j_{g} : \{1,2,...,m\} \rightarrow \{1,2,...,m\} \in S_{m}$ is a permutation of left coset representatives. The $h_{i}(g) \in H$ is an element of subgroup $H$. The map $j_{g}(i)$ and group element $h_{i}(g ) \in H$ satisfy a compositionality property. Specifically, we have that
\begin{align*}
    \forall g,g' \in G, \quad	j_{g'} \circ j_{g} = j_{g'g} , \quad  h_{i}(g'g) = h_{ j_{g}(i) }(g') \cdot  h_{i}(g) 
\end{align*}
which can be seen by acting on the left cosets with $g$ followed by $g'$ versus acting on the left cosets with $g'g$. Note that
\begin{align*}
    e \cdot g_{i} = g_{i} \cdot e= g_{j_{e}(i)} h_{i}(e) 
\end{align*}
holds so $j_{e} = e$ and $h_{i}(e) = e$ holds. Now, let $(\rho , V)$ be a representation of the group $H$. Let us define the vector space $W$ as
\begin{align*}
W = \bigoplus_{i=1}^{|G/H|} g_{i} V_{(i)}
\end{align*}
where the (standard albeit somewhat confusing) notation $ g_{i} V_{(i)}$ denotes an independent copy of the vector space $V$. This notation is simply a labeling and all copies of $ g_{i} V^{H}_{(i)}$ are isomorphic to $V^{H}$,
\begin{align*}
V \cong	g_{1} V_{1} \cong	g_{2} V_{2} \cong ... \cong g_{|G/H|} V_{|G/H|}
\end{align*}
so that the space $W \cong \bigoplus_{ i=1 }^{|G/H|} V$ is just $|G/H|$ independent copies of $V$. The induced representation lives on this vector space, $ (\pi , W) = \Ind_{H}^{G}[ (\rho , V) ] $. The induced action $\pi = \Ind_{H}^{G} \rho$ acts on the vector space $W$ via
\begin{align*}
\forall g \in G, \enspace \forall w = \sum_{i=1}^{|G/H|} g_{i} v_{i} \in W, \quad  \pi(g) \cdot w =  \sum_{i=1}^{|G/H|} \sigma( h_{i}(g)  ) v_{j_{g}(i) } \in W
\end{align*}
where $v_{i} \in V_{(i)}$ is in the $i$-th independent copy of the vector space $V$. Using the compositionality property of $j_{g}$ and $h_{i}(g)$, it is easy to see that this is a valid group action so that $(\pi , W) = \Ind_{H}^{G}[ (\rho , V) ]$ is a valid representation. Note that the induced action $\pi$ acts on the vector space $W$ by permuting and left action by the $H$-representation $\rho(h)$. There is a natural geometric interpretation of the induced representation which we discuss in a later section \ref{Appendix:Section:Tetra_Explict}.

\subsubsection{$G$-Intertwiners}

Let $(\rho , V)$ and $(\sigma , W)$ be two $G$-representations. The set of all $G$-equivariant linear maps between $(\rho , V)$ and $(\sigma , W)$ will be denoted as
\begin{align*}
    \Hom_{G}[ (\rho , V) , (\sigma , W) ] = \{ \enspace \Phi \enspace| \enspace \Phi: V \rightarrow W,\text{ s.t. }\enspace \forall g\in G, \enspace \Phi ( \rho(g) v ) = \sigma(g) \Phi(v) \enspace \}
\end{align*}
$\Hom_{G}$ is a vector space over $\mathbb{C}$. A linear map $\Phi \in \Hom_{G}[ (\rho , V) , (\sigma , W) ] $ is said to \emph{intertwine} the representations $(\rho , V)$ and $(\sigma , W)$. Pictorially, an intertwiner $\Phi$ is a map that makes the \ref{Diagram:G-Intertwiner} diagram commutative.

\begin{center}\label{Diagram:G-Intertwiner}
\captionof{figure}{Commutative Diagram For $G$-intertwiner. The map $\Psi \in \Hom_{G}[ (\rho , V) , (\sigma , W) ] $ if and only if the following diagram is commutative for all $g\in G$.    }
\end{center}

Computing a basis for the vector space $\Hom_{G}[ (\rho , V) , (\sigma , W) ]$ is one of the triumphs of classical group theory \cite{Serre_2005,Zee_2016}. The weights of Steerable CNNs are intertwiners between representations \cite{Cohen_2016}.

\subsubsection{$(H \subseteq G)$-Intertwiners}
We will also consider another definition of intertwiners between different groups. Let $H \subseteq G$. Let $(\rho , V)$ be a $H$-representation. Let $(\sigma , W)$ be a $G$-representation. We define the vector space of intertwiners of $(\rho , V)$ and $(\sigma ,W)$ as
\begin{align*}
    \Hom_{H}[  (\rho , V) , \Res_{H}^{G}[(\sigma, W)]  ] = \{ \enspace \Phi \enspace | \enspace  \Phi: V \rightarrow W, \text{ s.t. } \enspace \forall h\in H, \enspace \Phi ( \rho(h) v ) = \sigma(h) \Phi(v) \enspace \}
\end{align*}
We say that a linear map $\Phi:V \rightarrow W$ is an $(H\subseteq G)$-intertwiner of the $H$-representation $(\rho , V)$ and the $G$-representation $(\sigma , W)$ if $\Phi \in \Hom_{H}[  (\rho , V) , \Res_{H}^{G}[(\sigma, W)]  ] $. The induction and restriction operations are adjoint functors \cite{Ceccherini_2008}. By the Frobinous reciprocity theorem \cite{Ceccherini_2008},
\begin{align*}
    \Hom_{H}[  (\rho , V) , \Res_{H}^{G}[(\sigma, W)]  ]  \cong 	\Hom_{G}[  \Ind_{H}^{G}[ (\rho , V)] , (\sigma, W)  ] 
\end{align*}
and so for every $\Phi: V \rightarrow W$ which intertwines $(\rho , V)$ and $\Res_{H}^{G}[(\sigma, W)]$ over $H$ there is a unique $\Phi^{\uparrow} : \Ind_{H}^{G}[V] \rightarrow W$ that intertwines $\Ind_{H}^{G}[(\rho , V)]$ and $(\sigma , W)$ over $G$. Not every $H$-representation can be realized as the restriction of a $G$-representation. Thus, the universe of $(H\subseteq G)$-intertwiners is a proper subset of the universe of $H$-intertwiners. As explained in the main text, $(SO(2)\subseteq SO(3))$-intertwiners arise naturally when trying to design $SO(3)$-equivarient neural networks for image data.

\begin{center}\label{Diagram:HsG-Intertwiner}
    \begin{tikzcd}\centering
        & (\rho , V) \arrow{d}[swap]{ \rho(h) } \arrow{r}{ \Phi } & ( \sigma , W) \arrow{d}[swap]{ \sigma(h)  }{\sigma(g)}  \\
        & (\rho , V) \arrow{r}{ \Phi }  & ( \sigma , W)
    \end{tikzcd}
    \captionof{figure}{Commutative Diagram For $(H \subseteq G )$-intertwiner. $\Phi: V \rightarrow W$. The map $\Phi \in \Hom_{H}[ (\rho , V) , \Res_{H}^{G}[(\sigma , W)] ] \cong \Hom_{G}[  \Ind_{H}^{G}[ (\rho , V)] , (\sigma, W)  ] $ if and only if the following diagram is commutative for all $h \in H$. Note that the group $G$ also has $\sigma(g)$ action on the vector space $W$.   }
\end{center}

A map $\Phi : V \rightarrow W$ is a $(H \subseteq G)$-intertwiner if and only if the diagram in \ref{Diagram:HsG-Intertwiner} is commutative.

\section{Additional Experiments}\label{Appendix:Section:Datasets}

\paragraph{ModelNet10-SO(3) Results}\label{Appendix:Section:ModelNet10-SO(3) Results}

The first dataset, ModelNet10-SO(3) \cite{Liao_2019}, is composed of rendered images of synthetic, untextured objects from ModelNet10 \cite{Wu_2015}. The dataset includes 4,899 object instances over 10 categories, with novel camera viewpoints in the test set. Each image is labelled with a single 3D rotation matrix, even though some categories, such as desks and bathtubs, can have an ambiguous pose due to symmetry. For this reason, the dataset presents a challenge to methods that cannot reason about uncertainty over orientation.

\textbf{ModelNet10-SO(3) Results}

\begin{table}[H]
\caption{Rotation prediction on ModelNetSO(3). First column is the average over all categories.}\label{Figure:ModelNet_Table}
\centering
\scriptsize
\hspace*{-0.5cm}   
\begin{tabular}{ p{2.1cm}p{0.6cm}p{0.6cm}p{0.6cm}p{0.6cm}p{0.6cm}p{0.6cm}p{0.6cm}p{0.6cm}p{0.6cm}p{0.6cm}p{0.6cm}p{0.6cm}}
\toprule
& \multicolumn{11}{c}{Median rotation error in degrees ($\downarrow$)} \\
& \textit{avg} & \textit{bathtub} & \textit{bed} & \textit{chair} & \textit{desk} & \textit{dresser} & \textit{monitor} & \textit{stand} & \textit{sofa} & \textit{table} & \textit{toilet}   \\
\midrule
\citet{Mohlin_2021} & 17.1 & 89.1 & 4.4 & 5.2 & 13.0 &  6.3 &  5.8 & 13.5 & 4.0 & 25.8 & 4.0  \\
\citet{Prokudin_2018} & 49.3 & 122.8 & 3.6 & 9.6 & 117.2 & 29.9 & 6.7 & 73.0 & 10.4 & 115.5 & 4.1  \\
\citet{Deng_2020} &32.6 & 147.8 & 9.2 &8.3 & 25.0 & 11.9& 9.8& 36.9 & 10.0 & 58.6 &  8.5  \\
\citet{Liao_2019} & 36.5 & 113.3 & 13.3 & 13.7 & 39.2 & 26.9 & 16.4& 44.2 & 12.0 & 74.8 & 10.9 \\
\citet{Bregier_2021} & 39.9 & 98.9 & 17.4 & 18.0  & 50.0 & 31.5& 18.7& 46.5 & 17.4 & 86.7 & 14.2  \\
\citet{Zhou_2020_Cont} & 41.1 & \textbf{103.3} & 18.1 & 18.3 & 51.5 &  32.2 & 19.7 & 48.4 & 17.0 & 88.2 & 13.8  \\
\citet{Murphy_2022} & 21.5 & 161.0 & 4.4 & 5.5 & 7.1 & 5.5 & 5.7 & 7.5 & 4.1 & 9.0 & 4.8  \\
\citet{Klee_2023} & \textbf{16.3} & 124.7 & \textbf{3.1} & \textbf{4.4} & \textbf{4.7} & \textbf{3.4} & \textbf{4.4} & \textbf{4.1} & \textbf{3.0} & \textbf{7.7} & \textbf{3.6} \\
\textbf{Ours} & 17.8 & 123.7 & 4.6 & 5.5 & 6.9 & 5.2 & 6.1 & 6.5 & 4.5 & 12.1 & 4.9  \\
\bottomrule
\end{tabular}
\end{table}

The performance on the ModelNet dataset is reported in Table \ref{Figure:ModelNet_Table}. Our induction layer outputs signals on $S^{2}$, and naturally allows for capturing uncertainty as a distribution over $\SO(3)$. Both our method and \cite{Klee_2023} use equivariant layers to improve generalization but our method slightly under-performs \cite{Klee_2023} on the ModelNet dataset. ModelNet-10 is a synthetic dataset consisting of totally opaque objects and it seems that the image formation model used in \cite{Klee_2023} is a good approximation to the true image formation model. 

\section{Image to $\mathbb{R}^{3} \times S^{2} $ for 6DOF-Pose Estimation}\label{Section:Image to ...}

The goal in 6DOF-pose estimation is to estimate the location of an object in three-dimensional space and the orientation of said object. Orientation estimation is a sub-problem of pose estimation where the goal is to estimate just the orientation of an object and disregard the objects position in three-dimensional space.

Let us see how induced and restriction representations arise naturally in the design of neural architectures for 6DOF-pose estimation. Let $V$ and $V^{\uparrow}$ be vector spaces.
\paragraph{Image inputs} We first describe $\mathcal{F}$ the space of image input signals. Let $\mathcal{F}$ be the vector space of all $V$-valued signals defined on the plane
\begin{align*}
    \mathcal{F} = \{ \enspace f \enspace | \enspace f: \mathbb{R}^{2} \rightarrow V \enspace \}.
\end{align*}
Elements of $\mathcal{F}$ are referred to as $SE(2)$-steerable feature fields \citep{Weiler_2021}.

The group $SE(2)= \mathbb{R}^{2} \rtimes SO(2)$ of 2D translations and rotations acts on $\mathcal{F}$ via representation $\pi$. Each $h \in SE(2)$ has a unique factorization $h = \bar{h} h_{c}$ where $\bar{h} \in \mathbb{R}^{2}$ is a translation and $h_{c} \in SO(2)$ is a rotation.
Then $\pi$ is defined
\begin{align*}
r \in \mathbb{R}^{2}, \enspace \forall f \in \mathcal{F}, \enspace  h  \in SE(2), \quad     \pi(h) \cdot f(r) = \rho(h_{c}) f(h^{-1}r)
\end{align*}
where $( \rho , V)$ is an $SO(2)$-representation describing the transformation of the fibers of $f$ and $( \pi , \mathcal{F} ) = \Ind_{ SO(2) }^{ SE(2) } [ ( \rho , V) ]$ so that $( \pi , \mathcal{F} )$ gives a representation of the group $SE(2)$ \cite{Cohen_2016}. 

\paragraph{ 6DoF Pose outputs} 
In pose estimation tasks, the output of our neural network will be functions from $\mathbb{R}^{3} \times S^{2}$ into the vector space $V^{\uparrow}$. Let $\mathcal{F}^{\uparrow}$ be the vector space of all such outputs defined as
\begin{align*}
    \mathcal{F}^{\uparrow} = \{ \enspace f \enspace | \enspace f: \mathbb{R}^{3} \times S^{2} \rightarrow V^{\uparrow} \enspace \}
\end{align*}
The group $SE(3) = \mathbb{R}^{3} \rtimes SO(3)$ acts on the vector space $\mathcal{F}^{\uparrow}$ via
\begin{align*}
    \forall f^{\uparrow} \in \mathcal{F}^{\uparrow}, \enspace  \forall (p , \hat{n}) \in \mathbb{R}^{3} \times S^{2} , \enspace	\enspace \forall g = \bar{g} g_{c} \in SE(3),  \quad \pi^{\uparrow}(g) \cdot f^{\uparrow}(p, \hat{n} ) = \rho^{\uparrow}(g_{c}) f^{\uparrow}( g^{-1}p , g^{-1}_{c} \hat{n} )
\end{align*}
where $\rho^{\uparrow}(g_{c}) $ is a representation of $SO(3)$. Elements of $\mathcal{F}^{\uparrow}$ are referred to as $SE(3)$-steerable feature fields \citep{Weiler_2021}.

Analogous to the argument presented in the main text. We would like to characterize all maps from $\mathcal{F}$ to $\mathcal{F}^{\uparrow}$ that preserve $SE(2)$-equivarience. Consider the space of linear maps $\Phi : \mathcal{F} \rightarrow \mathcal{F}^{\uparrow} $ that intertwine $ ( \pi , \mathcal{F} )$ and $(\pi^{\uparrow} , \mathcal{F}^{\uparrow})$. The map $\Phi : \mathcal{F} \rightarrow \mathcal{F}^{\uparrow}$ must satisfy the relation
\begin{align*}
    \forall h\in SE(2), \enspace \forall f \in \mathcal{F}, \quad \Phi( \pi(h) \cdot f ) = \Res_{SE(2)}^{SE(3)}[ \pi^{\uparrow} ](h) \cdot \Phi(f)
\end{align*}
where $ \Res_{SE(2)}^{SE(3)}[ \pi^{\uparrow} ]$ is the restriction of the $SE(3)$-representation $(\pi^{\uparrow} , \mathcal{F}^{\uparrow})$ to a $SE(2)$ subgroup.

\subsubsection{Kernel Constraint for Image to 6DoF Pose}\label{Section:Deriving the Kernel Constraint}

The most general linear map $\Phi : \mathcal{F} \rightarrow \mathcal{F}^{\uparrow}$ between $( \pi , \mathcal{F})$ and $( \pi ^{\uparrow} , \mathcal{F}^{\uparrow})$ can be written as
\begin{align*}
    \forall (p , \hat{n} ) \in \mathbb{R}^{3} \times S^{2} , \quad	[ \Phi(f) ](p , \hat{n} ) = \int_{r \in \mathbb{R}^{2}} dr \text{ } \kappa(p , \hat{n} : r) f(r)
\end{align*}
where $\kappa : ( \mathbb{R}^{3} \times S^{2} ) \times \mathbb{R}^{2} \rightarrow \Hom[V,V^{\uparrow}]$. Let us enforce the $(H \subseteq G)$-equivarience condition
\begin{align*}
    \forall h\in SE(2) , \quad	\pi^{\uparrow}(h) \cdot  \Phi(f)   =  \Phi(  \pi(h) \cdot f) 
\end{align*}
This constraint places a restriction on the allowed space of kernels. We have that
\begin{align*}
    \forall h \in SE(2) , \quad	 \Phi [  \pi(h) \cdot f ] = \int_{ r \in \mathbb{R}^{2}} dr \text{ } \kappa(p , r) [ \pi(h) \cdot f(r) ] = \int_{r \in \mathbb{R}^{2}} dr \text{ } \kappa(p , \hat{n} : r) \rho(h_{c})  f(h^{-1}r) 
\end{align*}
Now, making the change of variables $r \rightarrow hr$ gives
\begin{align*}
    \forall h \in SE(2) , \quad	 \Phi [  \pi(h) \cdot f ] = \int_{r \in \mathbb{R}^{2}} dr \text{ }  \kappa(p , \hat{n} : h \cdot r ) \rho(h_{c}) f(r) 
\end{align*}
Now, by assumption $\Phi(f) \in (\pi^{\uparrow} , \mathcal{F}^{\uparrow})$ so
\begin{align*}
    \forall h \in SE(2) , \quad \pi^{\uparrow}(h) \cdot \Phi(f) = \int_{r \in \mathbb{R}^{2}} dr \text{ } \rho^{\uparrow}(h_{c}) \kappa( h^{-1}p, h^{-1} \hat{n} : r ) f(r) 
\end{align*}
Thus, the kernel $\kappa$ satisfies the constraint
\begin{align*}
    \forall h  \in SE(2) , \quad	\rho^{\uparrow}(h_{c}) \kappa( h^{-1} \cdot p, h^{-1}\hat{n} : r ) = \kappa(p , \hat{n} : h\cdot r) \rho(h_{c}) 
\end{align*}
We can write this in the more compact form as
\begin{align*}
    \forall h \in SO(2) , \quad  \kappa(h \cdot p, h \cdot \hat{n} :  h \cdot r) = \rho^{\uparrow}(h_{c}) \kappa( p  , \hat{n} : r ) \rho(h_{c}^{-1}  ) 
\end{align*}

This constraint is linear and solutions $\kappa$ form a vector space over $\mathbb{R}$. We reduce this constraint to the steerable kernel constraint considered in \cite{Cohen_2018,Weiler_2018,Cohen_2016, Lang_2020}.

First, note that the $SE(2)$ action does not mix the $z$-component of $ [ \Phi(f) ](\hat{n} , x,y,z)$. Thus, the most general linear map can be written as
\begin{align*}
    [ \Phi (f ) ] (\hat{n} , x,y,z ) = \int_{ (r_{x} ,r_{y}) \in \mathbb{R}^{2}} dr_{x}dr_{y} \text{ } \kappa(\hat{n} , x -  r_{x} , y - r_{y} , z) f(r_{x} , r_{y})
\end{align*}
where for each fixed $z$, the kernel $\kappa$ is an intertwiner of
$\Res_{SO(2)}^{SO(3)}[( \rho^{\uparrow} , V^{\uparrow} )]$ and $(\rho , V)$ and satisfies
\begin{align*}
    \forall h\in SO(2) , \quad \kappa(  h\cdot \hat{n} ,  h\cdot r : z) = \rho^{\uparrow}(h)\kappa(  \hat{n} ,  r : z)\rho(h^{-1})
\end{align*}

\begin{figure}[h!]\label{Figure:Kernel_Schematic}
\centering
\hspace{-1.0cm}
\includegraphics[width=1.05\textwidth]{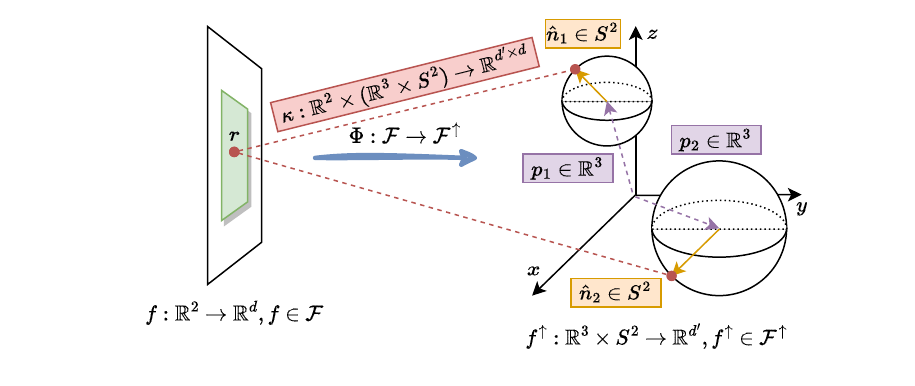 }
\caption{  Right: Diagram of an Equivariant Image to Sphere Convolution. At each point $p = (x,y,z) \in \mathbb{R}^{3}$ and each unit vector $\hat{n}\in S^{2}$ the kernel $\kappa( \hat{n} , p : p ') $ is dependent on the image point $p' = (x',y') \in \mathbb{R}^{2}$. Equivarience constraints put restrictions on the allowed form of $\kappa( \hat{n} , p : p ')$ \ref{Section:Deriving the Kernel Constraint}. Similar to a standard convolution, the kernel $\kappa$ has a user defined receptive field.  }
\end{figure}

Let simplify this constraint further. The set of spherical harmonics form an orthonormal basis for functions on $S^{2}$. We can expand the kernel $\kappa$ as
\begin{align*}
    \kappa(  \hat{n} ,  r :  z ) =  \sum_{\ell = 0}^{\infty} \sum_{k=-\ell}^{\ell} F^{k}_{\ell}(r , z) Y^{k}_{\ell}(\hat{n})
\end{align*}
where $F^{k}_{\ell}(r,z) : \mathbb{R}^{2} \times \mathbb{R} \rightarrow \Hom[ V , V^{\uparrow}]$. The kernel constraint places additional restrictions on the set of allowed $F^{k}_{\ell}(r,z)$. We have that,
\begin{align*}
    \forall h\in SO(2) , \quad	\kappa( h\cdot \hat{n} , h\cdot r : z) = \sum_{\ell = 0}^{\infty} \sum_{k=-\ell}^{\ell} F^{k}_{\ell}( h \cdot r , z) Y^{k}_{\ell}( h \cdot \hat{n}) = \sum_{\ell = 0}^{\infty} \sum_{k=-\ell}^{\ell} F^{k}_{\ell}( h \cdot r , z) D^{\ell}_{kk'}(h) Y^{k'}_{\ell}( \hat{n})
\end{align*}
and,
\begin{align*}
    \forall h\in SO(2) , \quad \rho^{\uparrow}(h)\kappa( \hat{n} , z : r )\rho(h^{-1}) =  \sum_{\ell = 0}^{\infty} \sum_{k=-\ell}^{\ell} \rho^{\uparrow}(h)  F^{k}_{\ell}( r, z)  \rho(h^{-1}) Y^{k}_{\ell}( \hat{n}) 
\end{align*}
Thus, the functions $F^{k}_{\ell}( r, z) : \mathbb{R}^{2} \times \mathbb{R} \rightarrow \Hom[V , V^{\uparrow}]$ must satisfy,
\begin{align*}
    \forall h \in SO(2) , \quad	  \rho^{\uparrow}(h)  F^{k}_{\ell}( r , z )  \rho(h^{-1}) = \sum_{k'=-\ell}^{\ell} F^{k'}_{\ell}( h \cdot r , z) D^{\ell}_{k'k}(h) 
\end{align*}
Now, the Wigner $D$-matrices are unitary and the above constraint is equivalent to
\begin{align*}
    \forall h \in SO(2) , \quad	F^{k}_{\ell}( h \cdot r , z) = \rho^{\uparrow}(h)  \sum_{k'=-\ell}^{+\ell} F^{k'}_{\ell}( r , z)  \rho(h^{-1}) D^{\ell}_{k'k}(h^{-1}) = \rho^{\uparrow}(h)  \sum_{k'=-\ell}^{+\ell} F^{k'}_{\ell}( r, z) [ D^{\ell}_{k'k}(h ) \rho(h ) ]^{-1}
\end{align*}
Now, let us vectorize the matrix valued functions $F^{k}_{\ell}(r,z)$ as
\begin{align*}
    F_{\ell}(r,z) = \begin{bmatrix}
        F^{\ell}_{\ell}(r,z) , & F^{\ell - 1}_{\ell}(r,z) , & ... & F^{-\ell + 1}_{\ell}(r,z)  , & F^{-\ell }_{\ell}(r,z) 
    \end{bmatrix} \in \Hom[ V \otimes W^{\ell} , V^{\uparrow} ]
\end{align*}
Let us define the tensor product representation of $(\rho , V)$ and $ \Res_{SO(2)}^{SO(3)}[ ( D^{\ell}, W^{\ell} )]$ as
\begin{align*}
    ( \rho^{\ell} , V^{\ell} ) =  ( \rho , V ) \otimes \Res_{SO(2)}^{SO(3)}[ ( D^{\ell}, W^{\ell} )]
\end{align*} 
which is a $SO(2)$-representation. Then the functions $F_{\ell}(r): \mathbb{R}^{2} \rightarrow \Hom[ V \otimes W^{\ell} ,V^{\uparrow} ]$ satisfy the constraint
\begin{align*}
    \forall h \in SO(2) ,\quad	F_{\ell}(h \cdot r , z) = \rho^{\uparrow}(h) F_{\ell}(r , z) \rho^{\ell}(h^{-1})
\end{align*}
For fixed $z$, this is exactly the constraint on an $SO(2)$-steerable kernel with input representation $( \rho^{\ell} , V^{\ell} ) = (\rho , V) \otimes \Res_{SO(2)}^{SO(3)}[(D^{\ell} , W^{\ell}) ]$ and output representation $ \Res_{SO(2)}^{SO(3)}[ \rho^{\uparrow} , V^{\uparrow} ) ]$. \cite{Weiler_2021,Lang_2020} give a complete classification of kernel spaces that satisfy this constraint. Note that by demanding that $SE(3)$ has action on the space $(\pi^{\uparrow} , \mathcal{F}^{\uparrow})$ we have added additional constraints to the set of allowed kernels. Specifically, instead of mapping arbitrary $SO(2)$-input representation to arbitrary $SO(2)$-output representation, the allowed input and output representations must satisfy additional constraints. Specifically, not every representation can be realized as the restriction of an $SE(3)$ to $SE(2)$ representation. The induction and restriction operations of $SO(2) \subset SO(3)$ on irreducible representations are shown in \ref{Figure:Induced_SO_2_Main}.

In practice, once the multiplicities of the input $SO(2)$-representation and the output $SO(3)$-representation are specified, the $SO(2)$-steerable kernels can be explicitly constructed using numerical programs defined in \cite{Weiler_2021}. To summarize, all equivariant linear maps between a function $f: \mathbb{R}^{2} \rightarrow V$ and a function $f^{\uparrow} : \mathbb{R}^{3} \times S^{2} \rightarrow V^{\uparrow}$ can be written as
\begin{align*}
f^{\uparrow}( \hat{n} , x , y , z ) = \sum_{\ell = 0 }^{\infty} ( F_{\ell , z} \star f )(x,y) \cdot Y_{\ell}(\hat{n}) = \sum_{\ell = 0 }^{\infty}  \int_{ (x',y') \in \mathbb{R}^{2}} dx'dy' \text{ } f(x',y') F_{\ell , z}(x-x',y-y') \cdot Y_{\ell}(\hat{n}) 
\end{align*}
where for each fixed $z$, $F_{\ell , z}(x,y)$ is a $SO(2)$-steerable kernel that takes input representation $( \rho^{\ell}  , V^{\ell} )= ( \rho , V) \otimes \Res_{SO(2)}^{SO(3)}[( D^{\ell} , W^{\ell} ) ]$ to output representation $ \Res_{SO(2)}^{SO(3)}[ ( \rho^{\uparrow} , V^{\uparrow}  ) ]$. Once the coefficients of the spherical harmonics
\begin{align*}
C_{\ell}(x,y,z) = (F_{\ell,z} \star f)(x,y) = \int_{ (x',y') \in \mathbb{R}^{2}} dx'dy' \text{ } f(x',y') F_{\ell , z}(x-x',y-y')
\end{align*}
are computed, the resultant function $f^{\uparrow}(\hat{n} , x, y , z) = \sum_{\ell=0}^{\infty} C^{T}_{\ell}(x,y,z) Y^{\ell}(\hat{n})$ is defined on a homogeneous space of $SE(3)$ and we can utilize $SE(3)$-steerable CNNs to make predictions about 6DoF poses \cite{Weiler_2018,Lin_2021,Jenner_2022}.

\section{Plane to Space for Object Reconstruction}\label{Appendix:Section:Plane to Space for Object Reconstruction}

Another problem of interest in single view geometric construction is monocular density reconstruction (also sometimes called monocular depth estimation). The goal in monocular density reconstruction problems is to build a three-dimensional model of the world given a single two-dimensional images \cite{Spencer_2022_Deconstructing,Saxena_2023_Monocular}. Monocular depth reconstruction tasks are of specific interest in endoscopy \cite{Liu_2019_Dense} and autonomous driving \cite{Batlle_2022,Fonder_2022_M4Depth}.

\paragraph{Volume Outputs}
In monocular reconstruction tasks, the output of our neural network will be a density map which is a function from $\mathbb{R}^{3}$ into a vector space $V^{\uparrow}$. Let $\mathcal{F}^{\uparrow}$ be the vector space of all such outputs,
\begin{align*}
\mathcal{F}^{\uparrow} = \{ \enspace f \enspace | \enspace f: \mathbb{R}^{3} \rightarrow V^{\uparrow} \enspace \}
\end{align*}
The group $\mathbb{R}^{3} \rtimes SO(3)$ acts on the vector space $\mathcal{F}^{\uparrow}$ via
\begin{align*}
    \forall f^{\uparrow} \in \mathcal{F}^{\uparrow} , \enspace \forall g\in SE(3),  \quad \pi^{\uparrow}(g) \cdot f^{\uparrow}(r) = \rho^{\uparrow}(g_{c}) f^{\uparrow}(g^{-1}r)
\end{align*}
where $\rho^{\uparrow}(g_{c}) $ is a representation of $SO(3)$. $\mathcal{F}^{\uparrow}$ are often refered to as $SE(3)$-steerable features. Now, consider the space of linear maps $\Phi : \mathcal{F} \rightarrow \mathcal{F}^{\uparrow} $ that intertwine $ ( \pi , \mathcal{F} )$ and $(\pi^{\uparrow} , \mathcal{F}^{\uparrow})$. The map $\Phi : \mathcal{F} \rightarrow \mathcal{F}^{\uparrow}$ must satisfy the relation
\begin{align*}
    \forall h\in SE(2), \enspace \forall f \in \mathcal{F}, \quad \Phi( \pi(h) f ) = \pi^{\uparrow}(h)\Phi(f)
\end{align*}
by definition of the restricted representation this is equivalent to 
\begin{align*}
    \forall h\in SE(2), \enspace \forall f \in \mathcal{F}, \quad \Phi( \pi(h) f ) = \Res_{H}^{G}[\pi^{\uparrow}](h)\Phi(f)
\end{align*}
where $ \Res_{SO(2)}^{SO(3)}[ ( \pi^{\uparrow} , \mathcal{F}^{\uparrow} ) ]$ is the restriction of the $SE(3)$-representation $(\pi^{\uparrow} , \mathcal{F}^{\uparrow})$ to a $SE(2)$ subgroup. 
\subsection{Kernel Constraint for Object Reconstruction}
Similar to \ref{Section:Image to ...}, the most general linear map between $( \pi , \mathcal{F})$ and $( \pi ^{\uparrow} , \mathcal{F}^{\uparrow})$ can be written as
\begin{align*}
    \forall p \in \mathbb{R}^{3} , \quad	( k\cdot f )(p) = \int_{r \in \mathbb{R}^{2}} dr \text{ } \kappa(p , r) f(r)
\end{align*}
where $\kappa : \mathbb{R}^{3} \times \mathbb{R}^{2} \rightarrow \Hom[V,V^{\uparrow}]$ satisfies the constraint
\begin{align*}
    \forall h  \in SE(2) , \quad	\rho^{\uparrow}(h_{c}) \kappa( h^{-1} \cdot p , r ) = \kappa(p , h \cdot r) \rho(h_{c}) 
\end{align*}
We can write this in the more compact form
\begin{align*}
    \forall h \in SO(2) , \quad  \kappa(h \cdot p , h \cdot r) = \rho^{\uparrow}(h_{c}) \kappa( p , r ) \rho(h_{c}) 
\end{align*}
Note that the $SO(2)$ action does not mix the $z$-component of $ [ \Phi(f)](x,y,z)$. Thus, the most general linear map can be written as
\begin{align*}
    [ \Phi(f) ](x,y,z) = \int_{r \in \mathbb{R}^{2}} dr_{x}dr_{y} \text{ } \kappa(x -  r_{x} , y - r_{y} , z) f(r_{x} , r_{y}) = ( \kappa_{z} \star f )(x,y)
\end{align*}
where for each fixed $z$, the kernel $\kappa$ is an intertwiner of
$\Res_{SO(2)}^{SO(3)}[(\rho^{\uparrow} , V^{\uparrow})]$ and $(\rho , V)$ and satisfies
\begin{align*}
    \forall h\in SO(2) , \quad \kappa(g\cdot r , z) = \rho^{\uparrow}(h)\kappa( r , z)\rho(h^{-1})
\end{align*}

To summarize, a function $f: \mathbb{R}^{2} \rightarrow V$ can be mapped into a function 
\begin{align*}
    f^{\uparrow}(x,y,z) =  \Phi(f) (x,y,z) = \int_{r \in \mathbb{R}^{2}} dr \text{ } k(x-x',y-y',z)f(x',y') = [ \kappa_{z} \star f ](x,y)
\end{align*}
where for fixed $z$, $\kappa_{z}$ is an $SO(2)$-steerable kernel with input representation $(\rho , V)$ and output representation $ \Res_{SO(2)}^{SO(3)}[(\rho^{\uparrow} , V^{\uparrow}) ]$.

\section{Image to $SO(3)$ for Rotation Estimation}\label{Appendix:Section:Plane to SO(3) for Rotation Estimation}

Instead of inducing from signals on the plane to signals on the $S^{2}$ as in \ref{Section:Main:Method}, we can induce directly from image to $SO(3)$. 
\paragraph{Rotation Outputs}
Let $\mathcal{F}^{\uparrow}$ be the vector space of all $SO(3)$ valued functions
\begin{align*}
    \mathcal{F}^{\uparrow} = \{ \enspace f \enspace | \enspace f: SO(3) \rightarrow V^{\uparrow} \enspace \}
\end{align*}
The group $SO(3)$ acts on the vector space $\mathcal{F}^{\uparrow}$ via
\begin{align*}
    \forall f^{\uparrow} \in \mathcal{F}^{\uparrow} , \enspace \forall g,g'\in SO(3),  \quad \pi^{\uparrow}(g) \cdot f^{\uparrow}(g') = \rho^{\uparrow}(g) f^{\uparrow}(g^{-1}g')
\end{align*}
where $\rho^{\uparrow}(g) $ is a representation of $SO(3)$. Now, consider the space of linear maps $\Phi : \mathcal{F} \rightarrow \mathcal{F}^{\uparrow} $ that intertwine $ ( \pi , \mathcal{F} )$ and $(\pi^{\uparrow} , \mathcal{F}^{\uparrow})$. The map $\Phi : \mathcal{F} \rightarrow \mathcal{F}^{\uparrow}$ must satisfy the relation
\begin{align*}
    \forall h\in SO(2), \enspace \forall f \in \mathcal{F}, \quad \Phi( \pi(h) f ) = \Res_{SO(2)}^{SO(3)}[\pi^{\uparrow}](h) \Phi(f) = \pi^{\uparrow}(h) \Phi(f)
\end{align*}
where $ \Res_{SO(2)}^{SO(3)}[\pi^{\uparrow}]$ is the restriction of the $SO(3)$-representation $(\pi^{\uparrow} , \mathcal{F}^{\uparrow})$ to a $SO(2)$ subgroup.
\subsection{Kernel Constraint for Image to $SO(3)$}
Using an argument similar to \ref{Section:Image to ...}, the most general linear equivariant map from functions on $\mathbb{R}^{2}$ to functions on the $SO(3)$ is 
\begin{align*}
    \forall g\in SO(3) , \quad	[ \Phi(f) ](g) = \int_{ (x,y) \in \mathbb{R}^{2}}dA \text{ } \kappa(  g , x , y  )  f(x,y)
\end{align*}
where the map $\kappa: SO(3) \times  \mathbb{R}^{2} \rightarrow \Hom[V,V^{\uparrow}]$. The kernel $\kappa$ satisfies 
\begin{align*}
    \forall h \in SO(2) , \quad \kappa( h^{-1} g , h^{-1} r ) = \rho^{\uparrow}(h) \kappa(g,r) \rho(h^{-1})
\end{align*}
The set of Wigner $D$-matrices form an orthonormal basis for functions on $SO(3)$ and we can uniquely expand $\kappa$ as
\begin{align*}
    \kappa(  g , x,y ) =  \sum_{\ell = 0}^{\infty} \sum_{k,k'=-\ell}^{\ell} F^{kk'}_{\ell}(x,y) D^{\ell}_{kk'}(g) 
\end{align*}
where $F^{kk'}_{\ell}(x,y) : \mathbb{R}^{2} \rightarrow \Hom[V,V^{\uparrow}]$ are matrix valued coefficients. The kernel constraint places restrictions on the allowed form of $F^{kk'}_{\ell}(x,y)$. Let us define the $SO(2)$-representations
\begin{align*}
    (\rho_{\ell} , V_{\ell}) =(\rho , V) \otimes  \Res_{SO(2)}^{SO(3)}[(D^{\ell} , W^{\ell}) ] , \quad 		(\rho^{\uparrow}_{\ell} , V^{\uparrow}_{\ell}) = \Res_{SO(2)}^{SO(3)}[  (\rho^{\uparrow} , V^{\uparrow}) \otimes (D^{\ell} , W^{\ell}) ]
\end{align*}
Then, the kernel constraint holds only if
\begin{align*}
    \forall h\in SO(2) , \enspace \forall r \in \mathbb{R}^{2} , \quad 	F^{\ell}_{kk'}(h \cdot r)  = \rho^{\uparrow}(h) [  \sum_{nn'=-\ell}^{\ell} D^{\ell}_{kn}(h) F^{\ell}_{nn'}(r) D^{\ell}_{n'k'}(h^{-1})  ] \rho(h^{-1}) 
\end{align*}
We can reduce this constraint to a standard $SO(2)$-kernel constraint by considering the $F_{\ell}(r)_{kk'} = F^{\ell}_{kk'}$ as a larger matrix. Then, the matrixed $F_{\ell}(x,y) : \mathbb{R}^{2} \rightarrow \Hom[ V \otimes W^{\ell}  ,V^{\uparrow} \otimes W^{\ell} ]$ are constrained to satisfy
\begin{align*}
    \forall h\in SO(2) , \quad F_{\ell}( h \cdot r ) = 	\rho^{\uparrow}_{\ell}(h) F_{\ell}(  r ) \rho_{\ell}(h^{-1})
\end{align*}
so that each $F_{\ell}(x,y)$ is an $SO(2)$-steerable kernel with input representation $	(\rho_{\ell} , V_{\ell}) = (\rho , V) \otimes \Res_{SO(2)}^{SO(3)}[ (D^{\ell} , W^{\ell}) ]$ and output representation $(\rho^{\uparrow}_{\ell} , V^{\uparrow}_{\ell}) = \Res_{SO(2)}^{SO(3)}[  (\rho^{\uparrow} , V^{\uparrow}) \otimes (D^{\ell} , W^{\ell}) ]$. The type of $F_{\ell}$ is determined by the Clebsch-Gordon coefficients and the branching/induction rules of $SO(2)$ and $SO(3)$.

\subsection{Ablation Study: Image to $S^{2}$ vs Image to $SO(3)$}

We rerun the experiments in the main text using an induction layer that maps images directly to $SO(3)$. The direct induction to $SO(3)$ slightly outperforms the induction to $S^{2}$ on the ModelNet dataset.

\begin{table}[H]
\caption{Rotation prediction on ModelNetSO(3). First column is the average over all categories.}\label{Appendix:Figure:ModelNet_Table}
\centering
\scriptsize
\hspace*{-0.5cm}   
\begin{tabular}{ p{2.1cm}p{0.6cm}p{0.6cm}p{0.6cm}p{0.6cm}p{0.6cm}p{0.6cm}p{0.6cm}p{0.6cm}p{0.6cm}p{0.6cm}p{0.6cm}p{0.6cm}}
\toprule
& \multicolumn{11}{c}{Median rotation error in degrees ($\downarrow$)} \\
& \textit{avg} & \textit{bathtub} & \textit{bed} & \textit{chair} & \textit{desk} & \textit{dresser} & \textit{monitor} & \textit{stand} & \textit{sofa} & \textit{table} & \textit{toilet}   \\
\midrule
$S^{2}$-Method & 17.8 & 123.7 & 4.6 & 5.5 & 6.9 & 5.2 & 6.1 & 6.5 & 4.5 & 12.1 & 4.9  \\
$SO(3)$-Method & 17.3 & 117.3 & 4.3  & 5.6 & 6.8  & 5.2 &  5.8  &  5.8  &  6.3 & 11.8 & 4.3  \\
\bottomrule
\end{tabular}
\end{table}

On both the SYMSOL and PASCAL3D+ datasets, the induction to $S^{2}$ followed by a standard spherical convolution outperform the direction induction to $SO(3)$ by a slight margin.
\begin{table}[H]
\centering
\caption{Average log likelihood (the higher the better $\uparrow$) on SYMSOL I \& II. Per \cite{Murphy_2022}, a single model is trained on all classes in SYMSOL I and a separate model is trained on each class in SYMSOL II.}
\small
\hspace*{-0.2cm}
\begin{tabular}{@{}ll@{\hskip 0.5cm}rrrrr@{\hskip 0.9cm}rrrr@{}}
\toprule
& \multicolumn{6}{c}{SYMSOL I ($\uparrow$)}       & \multicolumn{4}{c}{ SYMSOL II ($\uparrow$)} \\ 
& \textit{avg} & \textit{cone} & \textit{cyl} & \textit{tet} & \textit{cube} & \textit{ico} & \textit{avg}   & \textit{sphX}  & \textit{cylO}  & \textit{tetX}  \\ 
\midrule
$S^{2}$-Method & 5.11   &   4.91   &   4.22  &   6.10  &   5.73  &  4.69  &  6.20    &   7.10    &   6.01 &     5.62 \\ 
$SO(3)$-Method & 5.09   &  5.01  &   4.25  &  6.20  &   5.67  &  4.35 &   6.19   &   7.03   &  6.10 &   5.49 \\  \bottomrule
\end{tabular}
\label{Appendix:Table:SYMSOL}
\end{table}

\begin{table}[H]\centering\caption{Rotation prediction on PASCAL3D+. First column is the average over all categories. The feature encoder is either ResNet-50 or ResNet-101 head.  }
\hspace*{-0.7cm}   
\scriptsize
\begin{tabular}{ p{2.3cm}p{0.5cm}p{0.5cm}p{0.5cm}p{0.5cm}p{0.5cm}p{0.5cm}p{0.5cm}p{0.5cm}p{0.5cm}p{0.5cm}p{0.5cm}p{0.5cm}p{0.5cm}p{0.5cm} }
\toprule
& \multicolumn{13}{c}{Median rotation error in degrees ($\downarrow$)} \\
& \textit{avg} & \textit{plane} & \textit{bike} & \textit{boat} & \textit{bottle} & \textit{bus} & \textit{car} & \textit{chair} & \textit{table} & \textit{mbike} &\textit{sofa} & \textit{train} & \textit{tv}   \\
\midrule
$S^{2}$ (ResNet-50) & 10.2 & 9.2 & 13.1 & 30.6 & 6.7 & 3.1 & 4.8 & 8.7 & 5.4 & 11.6 & 11.0 & 5.8 & 10.6  \\
$SO(3)$ (ResNet-50) & 10.5 & 9.4 & 13.3 & 30.8 &  6.5 & 3.4 & 4.7 & 9.0 &  5.5 & 11.7 & 11.1  & 6.0 & 10.4 \\
\hline
$S^{2}$ (ResNet-101)& 9.2 & 9.3 & 12.6 & 17.0 & 8.0 & 3.0 & 4.5 & 9.4 & 6.7 & 11.9 & 12.1 & 6.9 & 9.9 \\
$SO(3)$ (ResNet-101)& 9.7 & 8.9 & 14.8 & 21.3 & 9.9 & 3.0 & 4.7 & 9.2 &  5.9 & 12.8 & 8.7 & 6.3 & 10.3 \\
\bottomrule
\end{tabular}
\label{Appendix:Table:pascal}
\end{table}

\section{Solving the Kernel Constraint For Image to Sphere}\label{Appendix:Section:Solving the Kernel Constraint}

Let us solve the kernel constraint presented in the main text \ref{Main:Equation:In Plane Equivarence_I}. The most general linear map $\Phi : \mathcal{F} \rightarrow \mathcal{F}^{\uparrow}$ between $( \pi , \mathcal{F})$ and $( \pi ^{\uparrow} , \mathcal{F}^{\uparrow})$ can be written as
\begin{align*}
    \forall \hat{n}  \in S^{2}, \quad	[ \Phi(f) ]( \hat{n} ) = \int_{r \in \mathbb{R}^{2}} dr \text{ } \kappa( \hat{n} , r) f(r)
\end{align*}
where $\kappa : S^{2} \times \mathbb{R}^{2} \rightarrow \Hom[V,V^{\uparrow}]$. Let us enforce the $SO(2)$-equivarience condition  derived in \ref{Main:Equation:In Plane Equivarence_I}. We have that,
\begin{align*}
    \forall h  \in SE(2), \quad	\pi^{\uparrow}(h_{c}) \cdot  \Phi(f)   =  \Phi(  \pi(h) \cdot f)
\end{align*}
This constraint places a restriction on the allowed space of kernels. We have that, $ \forall h= \bar{h}h_{c} \in SE(2)$,
\begin{align*}
\Phi [  \pi(h) \cdot f ] = \int_{ r \in \mathbb{R}^{2}} dr \text{ } \kappa(p , r) [ \pi(h) \cdot f(r) ] = \int_{r \in \mathbb{R}^{2}} dr \text{ } \kappa(p , \hat{n} : r) \rho(h_{c})  f(h^{-1}r) 
\end{align*}
Now, making the change of variables $r \rightarrow hr$ gives
\begin{align*}
    \forall h \in SE(2) , \quad	 \Phi [  \pi(h) \cdot f ] = \int_{r \in \mathbb{R}^{2}} dr \text{ }  \kappa(p , \hat{n} : h \cdot r ) \rho(h_{c}) f(r) 
\end{align*}
Now, by assumption $\Phi(f) \in (\pi^{\uparrow} , \mathcal{F}^{\uparrow})$ so
\begin{align*}
\forall h_{c} \in SO(2), \quad \pi^{\uparrow}(h_{c}) \cdot \Phi(f) = \int_{r \in \mathbb{R}^{2}} dr \text{ } \rho^{\uparrow}(h_{c}) \kappa(  h_{c}^{-1} \hat{n} : r ) f(r) 
\end{align*}
Thus, the kernel $\kappa$ satisfies the linear constraint
\begin{align*}
    \forall h \in SE(2) , \quad	\rho^{\uparrow}(h_{c}) \kappa(  h_{c}^{-1}\hat{n} : r ) = \kappa(p , \hat{n} : h\cdot r) \rho(h_{c}) 
\end{align*}
Fiber representations are unitary and left multiplying, we can the kernel constraint in the more compact form
\begin{align*}
    \forall h \in SO(2) , \quad  \kappa( h_{c} \cdot \hat{n} :  h \cdot r) = \rho^{\uparrow}(h_{c}) \kappa(  \hat{n} : r ) \rho(h_{c}^{-1}  ) 
\end{align*}
We can further reduce this to a standard steerable kernel constraint studied in \cite{Cohen_2018, Weiler_2018, Cohen_2016}. The set of spherical harmonics $Y^{k}_{\ell}$ form an orthonormal basis for functions on $S^{2}$. We can expand the kernel $\kappa$ as
\begin{align*}
    \kappa(  \hat{n} ,  r  ) =  \sum_{\ell = 0}^{\infty} \sum_{k=-\ell}^{\ell} F^{k}_{\ell}(r ) Y^{k}_{\ell}(\hat{n})
\end{align*}
where $F^{k}_{\ell}(r) : \mathbb{R}^{2} \rightarrow \Hom[ V , V^{\uparrow}]$. The kernel constraint places additional restrictions on the set of allowed $F^{k}_{\ell}(r)$. We have that,
\begin{align*}
    \forall h = \bar{h} h_{c} \in SO(2) , \quad	\kappa( h_{c} \cdot \hat{n} , h\cdot r ) = \sum_{\ell = 0}^{\infty} \sum_{k=-\ell}^{\ell} F^{k}_{\ell}( h \cdot r ) Y^{k}_{\ell}( h_{c} \cdot \hat{n}) = \sum_{\ell = 0}^{\infty} \sum_{k=-\ell}^{\ell} F^{k}_{\ell}( h \cdot r ) D^{\ell}_{kk'}(h_{c}) Y^{k'}_{\ell}( \hat{n})
\end{align*}
and,
\begin{align*}
    \forall h = \bar{h} h_{c} \in SO(2) , \quad \rho^{\uparrow}(h)\kappa( \hat{n}  : r )\rho(h^{-1}) =  \sum_{\ell = 0}^{\infty} \sum_{k=-\ell}^{\ell} \rho^{\uparrow}(h)  F^{k}_{\ell}( r, z)  \rho(h^{-1}) Y^{k}_{\ell}( \hat{n}) 
\end{align*}
Thus, the functions $F^{k}_{\ell}( r ) : \mathbb{R}^{2} \rightarrow \Hom[V , V^{\uparrow}]$ must satisfy,
\begin{align*}
    \forall h \in SO(2) , \quad	  \rho^{\uparrow}(h)  F^{k}_{\ell}( r  )  \rho(h^{-1}) = \sum_{k'=-\ell}^{\ell} F^{k'}_{\ell}( h \cdot r ) D^{\ell}_{k'k}(h) 
\end{align*}

Now, the Wigner $D$-matrices are unitary and the above constraint is equivalent to
\begin{align*}
    \forall h \in SO(2) , \quad	F^{k}_{\ell}( h \cdot r ) = \rho^{\uparrow}(h)  \sum_{k'=-\ell}^{+\ell} F^{k'}_{\ell}( r )  \rho(h^{-1}) D^{\ell}_{k'k}(h^{-1}) = \rho^{\uparrow}(h)  \sum_{k'=-\ell}^{+\ell} F^{k'}_{\ell}( r) [ D^{\ell}_{k'k}(h ) \rho(h ) ]^{-1}
\end{align*}
Now, let us vectorize the matrix valued functions $F^{k}_{\ell}(r)$ as
\begin{align*}
    F_{\ell}(r) = \begin{bmatrix}
        F^{\ell}_{\ell}(r) , & F^{\ell - 1}_{\ell}(r) , & ... & F^{-\ell + 1}_{\ell}(r)  , & F^{-\ell }_{\ell}(r) 
    \end{bmatrix} \in \Hom[ V \otimes W^{\ell} , V^{\uparrow} ]
\end{align*}
We define the tensor product representation of $(\rho , V)$ and $ \Res_{SO(2)}^{SO(3)}[( D^{\ell}, W^{\ell} )]$ as
\begin{align*}
    ( \rho^{\ell} , V^{\ell} ) =  ( \rho , V ) \otimes  \Res_{SO(2)}^{SO(3)}[( D^{\ell}, W^{\ell} )]
\end{align*} 
which is a $SO(2)$-representation. Then the functions $F_{\ell}(r): \mathbb{R}^{2} \rightarrow \Hom[ V \otimes W^{\ell} ,V^{\uparrow} ]$ satisfy the constraint
\begin{align*}
    \forall h \in SO(2) ,\quad	F_{\ell}(h \cdot r ) = \rho^{\uparrow}(h) F_{\ell}(r ) \rho^{\ell}(h^{-1})
\end{align*}
This is exactly the constraint on an $SO(2)$-steerable kernel with input representation $( \rho^{\ell} , V^{\ell} ) = (\rho , V) \otimes  \Res_{SO(2)}^{SO(3)}[( D^{\ell}, W^{\ell} )]$ and output representation $\Res_{SO(2)}^{SO(3)}[( \rho^{\uparrow} , V^{\uparrow} )]$. \cite{Weiler_2021,Lang_2020} give a complete classification of kernel spaces that satisfy this constraint. Note that by enforcing that the output transforms in an $SO(3)$-representation, we have added additional constraints to the set of allowed kernels.

\section{Including Non-linearities}\label{Appendix:Section:Including Non-linearities}


In section \ref{Section:Main:Deriving the Kernel Constraint}, we considered the most general linear maps that satisfied the generalized equivariance constraint.
After applying the linear layer described in \ref{Section:Image to ...}, we apply an additional RELU activation to the signal on $S^{2}$. It is also possible to use tensor-product based non-linearities analogous to the results of \cite{Thomas_2018,Kondor_2018}. In this section, we will consider how to include non-linearities for the general $H\subseteq G$ case where $G$ is a compact group. Let $(\rho, V)$ and $(\sigma ,W)$ be two irreducible $H$-representations. The tensor product representation of $(\rho, V)$ and $(\sigma ,W)$ will in general not be irreducible and will break down into irreducibles as
\begin{align*}
    (\rho, V) \otimes (\sigma ,W) = \bigoplus_{ \tau \in \hat{H} } c^{\tau}_{\rho \sigma} ( \tau , V_{\tau} )  
\end{align*}
where $c^{\tau}_{\rho \sigma}$ counts the number of copies of the $H$-irreducible $( \rho , V_{\tau} )$ in the tensor product representation. Analogous to the Clebsch-Gordon coefficients \cite{Lang_2020}, we can define $C^{\tau}_{\rho_{1}\rho_{2}}$ to be the coefficients of the representation $(\tau , V_{\tau})$ in the tensor product basis. Specifically, let
\begin{align*}
    | \tau i_{\tau} \rangle = \sum_{j_{1}=1}^{d_{1}} \sum_{j_{2}=1}^{d_{2}} \underbrace{ \langle \rho_{1} j_{1} ,  \rho_{2} j_{2} | \tau i_{\tau} \rangle }_{ ( C^{\tau}_{\rho_{1}\rho_{2}} )_{i_{\tau} , j_{1}j_{2} } } | \rho_{1} j_{1},  \rho_{2} j_{2}  \rangle 
\end{align*}
with $C^{\tau}_{\rho_{1}\rho_{2} }$ we can use the results of \cite{Thomas_2018} to project the tensor product unto a desired output representation. By choosing the output representation $(\tau , V_{\tau} )$ to be the restriction of an $G$ representation, we can use tensor products as non-linearites in the induction layer. One difficulty with this procedure is that it is too computationally expensive for practical use. It may be possible to simplify the complexity of implementation using the results of \cite{Passaro_2023_Reducing}. 
Tensor product based non-linearities for the construction in \ref{Main:Equation:In Plane Equivarence_I} is a promising future direction that we leave for future work.

\section{Generalization to Arbitrary Homogeneous Spaces }\label{Appendix:Section:Generalization to Arbitrary Homogeneous Spaces}

The results of \ref{Section:Deriving the Kernel Constraint} can be generalized to any $H \subseteq G$. Let $G$ be a compact group and let $H \subseteq G $. Let $H_{c} \subseteq H$ and let $X_{H} = H/H_{c}$ be a homogeneous space of $H$. Let $\mathcal{F}(X_{H})$ be the set of functions on $X_{H}$ that transform in representation $( \rho_{H} , V_{H}) $ of $H$,
\begin{align*}
    \mathcal{F}(X_{H}) = \{ \enspace f \enspace | \enspace f: X_{H} \rightarrow V_{H}, \quad [ h \cdot f](x) =  f(h^{-1}\cdot x) = \rho_{H}(h) f(x) \enspace \}
\end{align*}
Similarly, let $G_{c} \subseteq G$ and let $X_{G}= G/G_{c}$ be a homogeneous space of $G$. Let $\mathcal{F}(X_{G})$ be the set of functions on $X_{G}$ that transform in the representation $( \rho_{G} , V_{G}) $ of $G$,
\begin{align*}
    \mathcal{F}(X_{G}) = \{ \enspace f \enspace |  \enspace f: X_{G} \rightarrow V_{G} , \quad [g \cdot f](x) =  f(g^{-1}\cdot x) = \rho_{G}(g) f(x) \enspace \}
\end{align*}
We are interested in characterizing all equivariant maps $\Phi :\mathcal{F}(X_{H}) \rightarrow \mathcal{F}(X_{G})$ from $\mathcal{F}(X_{H})$ to $\mathcal{F}(X_{G})$. Now, generalizing the consistency condition derived in \ref{Main:Equation:In Plane Equivarence_I} to any $H \subseteq G$, the condition we seek to enforce is that 
\begin{align}\label{Equation:Generalized_Equ condition}
    \forall h\in H, \quad	\Phi(  \rho_{H}(h) \cdot f  ) = \rho_{G}(h) \cdot \Phi(f)
\end{align}
By definition of the restriction representation, \ref{Main:Section:Induced and Restriction Representations}, this is equivalent to the condition,
\begin{align}\label{Equation:Generalized_Equ condition_II}
    \forall h\in H, \quad	\Phi(  \rho_{H}(h) \cdot f  ) = \Res_{H}^{G}[ \rho_{G}(h)] \cdot \Phi(f)
\end{align}
Now, the most general linear map $ \Phi :\mathcal{F}(X_{H}) \rightarrow \mathcal{F}(X_{G})$ between the function spaces $\mathcal{F}(X_{H})$ and $\mathcal{F}(X_{G})$ can be written as
\begin{align*}
    \Phi(f)(x_{g}) = \int_{x_{h} \in X_{H}} dx_{h} \text{ }\kappa( x_{g} , x_{h} )f(x_{h})
\end{align*}
where the kernel $\kappa( x_{g} , x_{h} ) : X_{G} \times X_{H} \rightarrow \Hom[ V_{H} , V_{G} ]$ must satisfy the relation
\begin{align*}
    \forall h \in H, \quad  k(h \cdot x_{g} , h\cdot x_{h}) = \rho_{G}(h) k(x_{g},x_{h}) \rho_{H}(h)
\end{align*}
This is a generalization of the steerable kernel constraint first derived in \cite{Cohen_2016} and solved completely in \cite{Lang_2020}. Let us simplify this constraint to a more tractable form. Using a result stated in \cite{Lang_2020}, the functions on any homogeneous space of a compact group can always be decomposed into a sum of harmonic functions. Let $G$ be a compact group, and $X$ a homogeneous space of $G$, then for every $(\rho , V_{\rho}) \in \hat{G}$, there exist multiplicities $0 \leq m_{\rho} \leq d_{\rho}$ such that there exist a orthonormal basis $\{ Y^{\rho}_{ij} \}$ where the indices range over $\rho\in \hat{G}$ and $ i \in \{1,2,...,d_{\rho}\} , j \in \{1,2,...,m_{\rho}\}$  such that
\begin{align*}
    \forall j \in {1,2,...,m_{\rho}}, \quad	\forall g\in G, \enspace \forall x\in X, \quad	Y^{\rho}_{ij}( g^{-1} x ) = \sum_{i=1}^{d_{j}} \rho_{ii'}(g) Y^{\rho}_{i'j}(x)
\end{align*}

Let us denote the harmonic basis functions on the homogeneous space $X_{G}$ as $Y^{\sigma}_{ij}$. Using the orthogonality of harmonic functions, we can expand the $\kappa$ uniquely in terms of harmonics as
\begin{align*}
    k( x_{g} , x_{h} ) = \sum_{\sigma \in \hat{G} }  \sum_{i = 1}^{ d_{\sigma} } \sum_{j=1}^{m_{\sigma}}   F^{ \sigma }_{ij}(x_{h}) Y^{\sigma}_{ij}(x_{g}) 
\end{align*}
where $F^{ \sigma }_{ij} : X_{H} \rightarrow \Hom[ V_{H} , V_{G} ] $ are the matrix valued expansion coefficients of $\kappa$. We can simplify this expression for $\kappa$ by vectorizing,
\begin{align*}
    k( x_{g} , x_{h} ) = \sum_{\sigma \in \hat{G} } [ Y^{\sigma}(x_{g} )  ]^{T}  F^{  \sigma }(x_{h}) 
\end{align*}
where 
\begin{align*}
    F^{  \sigma }(x_{h})  : X_{H} \rightarrow \Hom[  V_{H} , V_{G} \otimes ( \underbrace{ V_{\sigma} \oplus V_{\sigma}\oplus  ... \oplus V_{\sigma} }_{m_{\sigma}  \text{ copies}  } )  ]
\end{align*}
Let us denote $( m_{\sigma} \sigma , m_{\sigma} V_{\sigma})$ as $m_{\sigma}$ copies of the $G$-irreducible $(\sigma , V_{\sigma})$,
\begin{align*}
    ( m_{\sigma} \sigma , m_{\sigma} V_{\sigma} ) = \underbrace{ (\sigma , V_{\sigma}) \oplus (\sigma , V_{\sigma})  \oplus ... \oplus (\sigma , V_{\sigma})  }_{ m_{\sigma} \text{ copies} }
\end{align*}
The kernel constraint places a restriction on the allowed form of the $F^{  \sigma }(x_{h})$. We have that
\begin{align*}
    \forall h \in H, \quad		k( h\cdot x_{g} , h\cdot x_{h} ) = \sum_{\sigma \in \hat{G} }  [ Y^{\sigma}( h\cdot x_{g} ) ]^{T}   F^{  \sigma }( h\cdot x_{h})  = \sum_{\sigma \in \hat{G} }  [  m_{\sigma}\sigma(h^{-1})  \cdot Y^{\sigma}( x_{g} ) ]^{T}   F^{  \sigma }( h\cdot x_{h})
\end{align*}
Using the identity $\sigma(h^{-1})^{T} = \sigma(h)$, we have that,
\begin{align*}
    \forall h \in H, \quad		k( h\cdot x_{g} , h\cdot x_{h} ) = \sum_{\sigma \in \hat{G} }  [  Y^{\sigma}( x_{g} ) ]^{T} [ m_{\sigma}\sigma(h)  \cdot   F^{  \sigma }( h\cdot x_{h}) ]
\end{align*}
Now, using \ref{Equation:Generalized_Equ condition}, $k( h\cdot x_{g} , h\cdot x_{h} ) $ must be equal to $\rho_{G}(h) k( x_{g} ,  x_{h} ) \rho_{H}(h)$. This is only satisfied if and only if
\begin{align*}
    \forall h \in H, \quad	 F^{  \sigma }( h\cdot x_{h})  =  ( \rho_{G} \otimes m_{\sigma}\sigma )(h)  \cdot F^{  \sigma }( x_{h}) \cdot \rho_{H}(h)
\end{align*}
Thus, $F^{  \sigma }$ is a $H$-steerable kernel with input representation $\rho_{H}$ and output representation  $ \Res_{H}^{G}[( \rho_{G} \otimes m_{\sigma}\sigma ) ]$. Note that the Clebsch-Gordon coefficients, the multiplicities $m_{\sigma}$ and the induction/restriction coefficients completely determine the output representation type of the $H$-steerable kernels $F^{  \sigma }$.

\begin{figure}[H]\label{Figure:Induced_General}
    \centering
    \begin{tabular}{cc}   
        \includegraphics[width=0.45\textwidth]{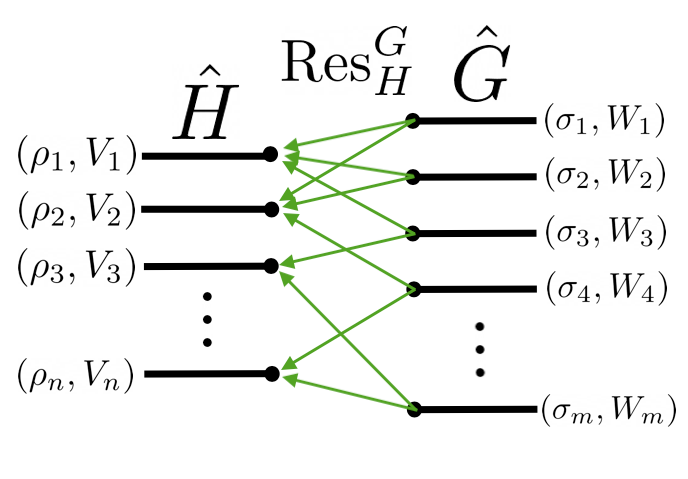}
        \hspace*{+0.3cm}   
        \includegraphics[width=0.45\textwidth]{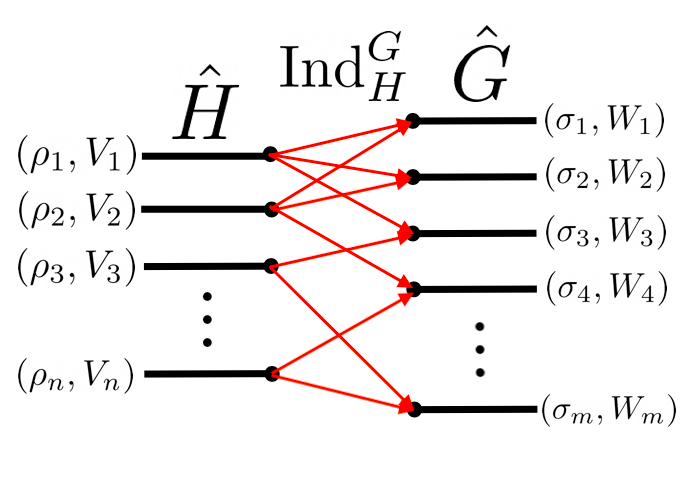}
    \end{tabular}
    \caption{Left: Restricted representation $\Res_{H}^{G}$ from $G$ to $H$ of $G$-irreducibles $(\sigma_{i} , W_{i})$ to $H$-irreducibles $(\rho_{j} , V_{j})$. Not every $H$-representation can be realized as the restriction of a $G$-representation. Right: Induced representation $\Ind_{H}^{G}$ from $H$ to $G$ of $H$-irreducibles $(\rho_{j} , V_{j})$ to $G$-irreducibles $(\sigma_{i} , W_{i})$. Not every $H$-representation can be realized as the induction of a $H$-representation. The restriction and induction operations are adjoint functors. In general, the restriction and induction operations are generically \emph{sparse}. This sparsity places restrictions on what irreducibles can appear in $(H \subseteq G)$-equivariant maps.   }
\end{figure}

\section{A Completeness Property For Induced Representations }
	
Much of the early work on machine learning focused on proving that sufficiently wide and deep neural networks can approximate any function within some accuracy \cite{Hornik_1989}. A network that can approximate any function is said to be expressive. The induced representation satisfies a completeness property.

\subsection{Group Valued Functions and Completeness}

Can every function $f : G \rightarrow \mathbb{R}^{c}$ be realized as the induced mapping of functions in $\mathbb{R}^{H}$? We show that this is the case. We have the following compositional property of induced representations \cite{Ceccherini_2018}: Let $ K \subseteq H \subseteq G$. Let $(\rho ,V)$ be any representation of $K$. Then,
\begin{align}\label{Equation:Induced_rep_composition}
    \text{Ind}_{K}^{G}[ (\rho ,V ) ] = \text{Ind}_{H}^{G}[  \text{Ind}_{H}^{K}[ (\rho , V) ] ]
\end{align}
which states that the induced representation of $(\rho , V)$ from $K$ to $G$ can be constructed by first inducing $(\rho ,V)$ from $K$ to $H$ and then inducing from $H$ to $G$.

Now, choose $K = \{e\}$ to be the identity element of $G$. Let $(\rho ,V)$ be the trivial one dimensional representation of $K = \{e\}$ with
\begin{align*}
    \dim V = 1,	\quad		\rho(e) v = v
\end{align*}
Consider the set of left cosets of $H$ in $K=\{e\}$. We have that
\begin{align*}
    H/K = H/\{e\} = \{ he | h \in G  \} = H
\end{align*}
so the set of coset representatives of $H/K$ is just elements of $H$. Using a from \cite{Ceccherini_2018}, the induced representation of $(\rho , V)$ from $K = \{e\}$ to $H$ is the left regular representation of $H$. By the same argument, the induced representation of $(\rho , V)$ from $K = \{e\}$ to $G$ is the left regular representation of $G$. Thus,
\begin{align*}
    & \text{Ind}_{K}^{H}[ (\rho , V) ] = ( L , \mathbb{C}^{H} ) , \quad \text{Ind}_{K}^{G}[ (\rho , V) ] = (  L , \mathbb{C}^{G} )
\end{align*}
Using the compositionality property of the induced representation \eqref{Equation:Induced_rep_composition}, we thus have that
\begin{align*}
    ( L , \mathbb{C}^{G}) = \text{Ind}_{H}^{G}[  ( L , \mathbb{C}^{H} ) ]
\end{align*}

Thus, the induced representation from $H$ to $G$ of the left regular representation of $H$ is the left regular representation of $G$.

\begin{center}\label{Diagram:Completeness_Property}
    \begin{tikzcd}[row sep=large, column sep = huge]\centering
        & ( L , \mathbb{C}^{H} ) \arrow{r}{ \text{Ind}^{G}_{H}[ ( L ,  \mathbb{C}^{H} ) ] } \arrow{d}[swap]{L(h)} & (L , \mathbb{C}^{G} ) \arrow{d}{L(g)}[swap]{L(h)} \\
        & ( L , \mathbb{C}^{H} ) \arrow{r}{\text{Ind}^{G}_{H}[ ( L , \mathbb{C}^{H} ) ] } & (L, \mathbb{C}^{G} )
    \end{tikzcd}
    \captionof{figure}{Commutative Diagram for Completeness Property of Induced Representations. $L_{h}$ denotes the left regular action of $H$ on $\mathbb{C}^{H}$. $L_{g}$ denotes the left regular action of $G$ on $\mathbb{C}^{G}$. The induced representation of the left regular representation of $H$ is the left regular representation of $G$, $	( L , \mathbb{C}^{G}) = \text{Ind}_{H}^{G}[  ( L , \mathbb{C}^{H} ) ]$. The induced representation makes the diagram commutative. This should be contrasted with the definition of $G$-equivarience defined in \ref{Diagram:G-Equivarient_Map}.   }
\end{center}

Thus, the induction operation maps the space of all group valued functions on $H$ into the space of all group valued functions on $G$.

\section{Irriducibility and Induced and Restricted Representations}

Let $H$ be a subgroup of compact group $G$. We can use the induced representation to map representations of $H$ to representations of $G$ and the restricted representation to map representations of $G$ to representations of $H$. All representations of $H$ break down into direct sums of irreducible representations of $H$. Similarly, all representations of $G$ break down into direct sums of irreducible representations of $G$. Let use denote $\hat{H}$ as a set of representatives of all irreducible representations of $H$ and $\hat{G}$ as a set of representatives of all irreducible representations of $G$,
\begin{align*}
& \hat{H} = \{ \enspace ( \rho , V_{\rho} ) \enspace | \enspace  \text{ Representative irreducibles of } H \enspace \} \\
& \hat{G} = \{ \enspace ( \sigma , W_{\sigma} ) \enspace | \enspace  \text{  Representative irreducibles of }G \enspace \}
\end{align*}

We want to understand how the restriction and induction operations transform $H$-irreducibles to $G$-irreducibles and vice versa. We can completely characterize how irreducibles change under the restriction and induction procedures using \emph{branching rules} and \emph{induction rules}, respectively.

\subsection{Restricted Representation and Branching Rules}
Let $(\sigma , W)$ and $(\sigma' , W')$ be $G$-representations. The restriction operation is linear and 
\begin{align*}
    \Res_{H}^{G}[( \sigma  , W  )\oplus (  \sigma' ,  W' ) ] = 	\Res_{H}^{G}[( \sigma  , W )] \oplus 	\Res_{H}^{G}[( \sigma'  , W' )]
\end{align*}
We can study the restriction operation by looking at restrictions of the set of $G$-irreducibles $\hat{G}$. The restriction of an $G$-irreducible is not necessarily irreducible in $H$ and will decompose as a direct sum of $H$-irreducibles. Let $(\sigma , W_{\sigma})\in \hat{G}$. We can define a set of integers $B_{\sigma , \rho} : \hat{G} \times \hat{H} \rightarrow \mathbb{Z}^{\geq 0}$,
\begin{align*}
    \Res_{H}^{G}[(\sigma , W_{\sigma} ) ] = \bigoplus_{ \rho \in \hat{H} } B_{\sigma, \rho}(\rho , W_{\rho})
\end{align*}
so that $B_{\sigma, \rho}$ counts the multiplicities of the $H$-irreducible $(\rho , W_{\rho})$ in the restricted representation of the $G$-irreducible $(\sigma , W_{\sigma}) $. The $B_{\sigma, \rho}$ are called \emph{branching rules} and they have been well studied in the context of particle physics \cite{Zee_2016}. Let $(\sigma' , W')$ be any $G$-representation. $(\sigma' , W')$  will decompose into $G$-irreducibles as
\begin{align*}
    ( \sigma' , W') = \bigoplus_{ \sigma \in \hat{G} } m_{\sigma} (\sigma , W_{\sigma} )
\end{align*}
where $m_{\sigma}$ counts the number of copies of the $G$-irreducible $(\sigma,W_{\sigma})$ in $( \sigma' , W')$. Then, the restriced representation of $(\sigma', W')$ decomposes into $H$-irreducibles as
\begin{align*}
    \Res_{H}^{G}[ (\sigma' , W') ] = \bigoplus_{ \sigma \in \hat{G} } m_{\sigma} \Res_{H}^{G}[(\sigma , W_{\sigma} )] = \bigoplus_{ \rho \in \hat{G} }  \sum_{\sigma \in \hat{G}} [m_{\sigma}B_{\sigma , \rho } ] (\rho , W_{\rho} )
\end{align*}
So that the multiplicity of the $(\rho , W_{\rho} )$ irreducible in the restriction of $(\sigma' , W')$ is $\sum_{\sigma \in \hat{G}} m_{\sigma}B_{\sigma , \rho } $. Thus, the branching rules $B_{\sigma,\rho}$ completely determine how an arbitrary $G$-representation restricts to an $H$-representation.

\subsection{Induced Representation and Induction Rules}

The induction operation acts linearly on representations composed of direct sums of representations. Specifically, if $(\rho_{1} , V_{1})$ and $(\rho_{2} , V_{2})$ are representations of $H$, then
\begin{align*}
    \text{Ind}_{H}^{G}[(\rho_{1} , V_{1}) \oplus (\rho_{2} , V_{2})  ] = 	\text{Ind}_{H}^{G}[(\rho_{1} , V_{1}) ]\oplus \text{Ind}_{H}^{G}[ (\rho_{2} , V_{2})  ] 
\end{align*}

The induction operation $\text{Ind}^{G}_{H}$ maps every irreducible representation $( \rho , V_{\rho} ) \in \hat{H}$ to a $G$-representation. The induced representation of an irreducible representation of $H$ is not necessarily irreducible in $G$ and will break into irreducibles in $\hat{G}$ as
\begin{align*}
    \Ind_{H}^{G}[( \rho , V_{\rho}) ] = \bigoplus_{ \sigma \in  \hat{G}  } I_{ \rho ,  \sigma } ( \sigma , W_{\sigma} )
\end{align*} 
where the integers $I_{\rho,\sigma}: \hat{H} \times \hat{G} \rightarrow \in \mathbb{Z}^{\geq 0}$ denotes the number of copies of the irreducible $( \sigma, W_{\sigma} ) \in \hat{G}$ in the induced representation $\Ind_{H}^{G}( \rho , V_{\rho}) $ of the irreducible $( \rho , V_{\rho})$. The $I_{ \rho , \sigma}$ are called \emph{Induction Rules} and completely determine the multiplicities of $G$-irreducibles in the induced representation of any $H$-representation. Specifically, let $(\rho' , V')$ be any representation of $H$. Then, $(\rho' , V')$ breaks into $H$-irreducibles as
\begin{align*}
    (\rho' , V') = \bigoplus_{ \rho \in  \hat{H} } n_{\rho} ( \rho , V_{\rho} )
\end{align*}
The induced representation is linear and maps $(\rho' , V')$ into a representation of $G$ which will break into $G$-irreducibles as
\begin{align*}
    \Ind_{H}^{G}[ (\rho' , V') ] = \bigoplus_{ \rho \in \hat{H} } n_{\rho} \Ind_{H}^{G}( \rho , V_{\rho} ) =  \bigoplus_{ \sigma \in \hat{G} } ( \sum_{ \rho \in \hat{H}} n_{\rho} I_{\rho , \sigma} ) ( \sigma , W_{\sigma} )
\end{align*}
so that the multiplicity of $( \sigma , W_{\sigma} ) \in \hat{G}$ in the induced representation of $(\rho, V_{\rho})\in \hat{H}$ is given by $\sum_{ \rho \in \hat{H} } m_{\sigma} I_{ \rho , \sigma} $. Thus, the induction rules $I_{\rho,\sigma}$ completely determine the multiplicities of $G$-representations in the induced representation of any $H$-representation.

\subsection{Irriducibility and Frobinous Reciprocity}

The induction rules $I_{\rho \sigma}:\hat{H}\times\hat{G} \rightarrow \mathbb{Z}^{\geq0}$ and the branching rules $B_{ \sigma \rho}:\hat{G} \times \hat{H} \rightarrow \mathbb{Z}^{\geq0}$ are related by the Frobinous reciprocity theorem \cite{Ceccherini_2008}. Let $(\rho',V')$ be any $H$-representation and let $(\sigma' , W')$ be any $G$-representation. Then,
\begin{align*}
    \Hom_{H}[ (\rho' , V') , \Res_{H}^{G}[ (\sigma', W') ] ] \cong \Hom_{G}[ \Ind_{H}^{G}[(\rho' , V')] , (\sigma', W') ]
\end{align*}
Choosing $(\rho' , V') = (\rho , V_{\rho})  \in \hat{H}$ and $(\sigma' , W') = (\sigma , W_{\sigma})  \in \hat{G} $ gives $	I_{ \rho, \sigma } = B_{\sigma , \rho}$. So that when viewed as matrices, $ B = I^{T}$. All information about how $H$-representations are induced to $G$-representations and $G$-representations are restricted to $H$-representations is encoded in both $B_{\sigma,\rho}$ and $I_{\rho,\sigma}$. It should be noted for many cases of interest, $B_{\sigma,\rho}$ and $I_{\rho,\sigma}$ are sparse, and have non-zero entries for only a small number of $\rho$ and $\sigma$ pairs. In the next section, we discuss how the structure of $B_{\sigma,\rho}$ and $I_{\rho,\sigma}$ constraint the design of equivariant neural architectures.

\subsection{Induced and Restriction Representation Based Architectures}

Heuristically, convolutional neural networks are compositions of linear functions, interleaved with non-linearities. At each layer of the network, we have a set of functions from a homogeneous space of a group into some vector space \cite{Kondor_2018}. Let $X^{H}_{i}$ be a set of homogeneous spaces of the group $H$ and let $X^{G}_{j}$ be a set homogeneous spaces of the group $G$. Let $V^{H}_{i}$ and $W^{G}_{j}$ be a set of vector spaces .Then, consider the function spaces
\begin{align*}
    \mathcal{F}^{H}_{i} = \{ \enspace f \enspace | \enspace f : X^{H}_{i} \rightarrow V^{H}_{i} \enspace \}, \quad \quad 	\mathcal{F}^{G}_{j} = \{ \enspace  f' \enspace | \enspace f' : X^{G}_{j} \rightarrow W^{G}_{j} \enspace \}
\end{align*}
The group $H$ acts on the homogeneous spaces $X^{H}_{i}$ and the group $G$ acts on the homogeneous spaces $X^{G}_{j}$ so that the function spaces $\mathcal{F}^{H}_{i}$ and $\mathcal{F}^{G}_{j}$ form representations of $H$ and $G$, respectively

Suppose we wish to design a downstream $G$-equivariant neural network that accepts as signals functions that live in the vector space $\mathcal{F}^{H}_{0}$ and transform in the $\rho_{0}$ representation of $H$. Thus, $( \rho_{0} , \mathcal{F}^{H}_{0})$ is a $H$-representation, but not necessarily a $G$-representation. At some point, in the architecture, a layer $\mathcal{F}^{H}_{i}$ must be $H$ equivariant on the left and both $H$ and $G$-equivariant on the right. Let us call the layer that is both $H$ and $G$-equivariant $\mathcal{F}^{G}_{1}$.

\tiny
\begin{center}\label{Appendix:Diagram:Switching}
\begin{tikzcd}[column sep=4ex,row sep=5ex]
    & ...\arrow{r}{\Phi_{i-1}}  & ( \rho_{i}, \mathcal{F}_{i}^{H} ) \arrow{d}{{ \rho_{i}(h) } } \arrow{r}{ \Psi } & ( \sigma_{1} , \mathcal{F}_{1}^{G} ) \arrow{d}{ \sigma_{1}(g) } \arrow{r}{\Psi_{1}} & ...  \\
    & ...\arrow{r}[swap]{\Phi_{i-1}}  &( \rho_{i} , \mathcal{F}_{i}^{H} )\arrow{r}[swap]{\Psi} & ( \sigma_{1} , \mathcal{F}_{1}^{G} ) \arrow{r}[swap]{\Psi_{1}} &  ...
\end{tikzcd} \quad $\cong$  \begin{tikzcd}[column sep=4ex,row sep=5ex]
    & ...\arrow{r}{\Phi_{i-1}}  & ( \rho_{i} , \mathcal{F}_{i}^{H} ) \arrow{d}{{ \rho_{i}(h) } } \arrow{r}{ \Phi_{ \rho_{i} } } & \text{Ind}_{H}^{G}[ ( \rho_{i} ,   \mathcal{F}^{H}_{i} ) ]\arrow{d}{ \text{Ind}_{H}^{G}[\rho_{i}]} \arrow{r}{\Psi^{\uparrow}}& ( \sigma_{1} , \mathcal{F}_{1}^{G}  ) \arrow{d}{ \sigma_{1}(g) } \arrow{r}{\Psi_{1}} & ...  \\
    & ...\arrow{r}[swap]{\Phi_{i-1}}  &( \rho_{i} , \mathcal{F}_{i}^{H}) \arrow{r}[swap]{\Phi_{\rho_{i}}} &  \text{Ind}_{H}^{G}[  (\rho_{i} ,   \mathcal{F}^{H}_{i} )  ] \arrow{r}[swap]{\Psi^{\uparrow}} & ( \sigma_{1} , \mathcal{F}_{1}^{G} ) \arrow{r}[swap]{\Psi_{1}} & ...
\end{tikzcd}
\captionof{figure}{ Factorization of Generic Architecture Using Universal Property of Induced Representation \ref{Diagram:Universality Property_Push} $\Psi = \Psi^{\uparrow} \circ \Phi_{\sigma_{i}}$ }
\end{center}
\normalsize

Suppose that $\Psi$ is an intertwiner between $( \rho_{i} , \mathcal{F}^{H}_{i})$ and $(\sigma_{1} , \mathcal{F}^{G}_{1})$. Using \ref{Diagram:Universality Property_Push}, there is a canonical basis of the space $\Hom_{H}[ (\rho_{i}, \mathcal{F}^{H}_{i}) ,  \Res_{H}^{G}[(\sigma_{1} , \mathcal{F}_{1}^{G} )] ] \cong \Hom_{G}[ \Ind_{H}^{G}[ (\rho_{i} , \mathcal{F}^{H}_{i})] , (\sigma_{1} , \mathcal{F}^{G}_{1}) ] $ and we may write $\Psi$ uniquely as $\Psi = \Psi^{\uparrow } \circ \Phi_{\rho}$ where $\Phi_{\rho}$ is an $H$-equivariant map and $\Psi^{\uparrow }$ is a $G$-equivariant map.
 
\tiny
\begin{center}\label{Diagram:General_Neural_Network}
		\begin{tikzcd}\centering
			&( \rho_{0} ,  \mathcal{F}_{0}^{H} ) \arrow{d}{\rho_{0}(h)}\arrow{r}{\Phi_{0}} & ( \rho_{1} ,  \mathcal{F}_{1}^{H} ) \arrow{d}{\rho_{1}(h)} \arrow{r} {\Phi_{1}} & ...\arrow{r}{\Phi_{i-1}} \arrow{d} &( \rho_{i} ,  \mathcal{F}_{i}^{H} ) \arrow{d}{{ \rho_{i}(h) } } \arrow{r}{ \text{Ind}_{H}^{G} } & ( \sigma_{1} , \mathcal{F}_{1}^{G} )  \arrow{d}{ \sigma_{1}(g) } \arrow{r}{\Psi_{1}} & ( \sigma_{2} , \mathcal{F}_{2}^{G} )  \arrow{r}{\Psi_{2}}  \arrow{d}{ \sigma_{2}(g) } & ... \arrow{r}{\Psi_{j-1}}  \arrow{d} & ( \sigma_{j} , \mathcal{F}_{j}^{G} ) \arrow{d}{ \sigma_{j}(g) }\\
			&( \rho_{0} ,  \mathcal{F}_{0}^{H} )\arrow{r}[swap]{\Phi_{0}}  & ( \rho_{1} ,  \mathcal{F}_{1}^{H} )\arrow{r}[swap]{\Phi_{1}} & ...\arrow{r}[swap] {\Phi_{i-1}}& ( \rho_{0} ,  \mathcal{F}_{0}^{H} ) \arrow{r}[swap]{\text{Ind}_{H}^{G}} & ( \sigma_{1} , \mathcal{F}_{1}^{G} ) \arrow{r}[swap]{\Psi_{1}} & ( \sigma_{2} , \mathcal{F}_{2}^{G} ) \arrow{r}[swap]{\Psi_{2}}  & ... \arrow{r}[swap]{\Psi_{j-1}} & (\sigma_{j} , \mathcal{F}_{j}^{G} )
		\end{tikzcd}
		\captionof{figure}{ Most general downstream $G$-equivariant architecture that accepts signals of capsule type $\rho_{0}$ that live in vector space $\mathcal{F}_{0}^{H}$. Using the universal property of the induction layer, all downstream $G$-equivariant architectures can be written in this form.}
	\end{center}
    \normalsize
 
	Using this decomposition, we may write any $G$-equivariant neural architecture that accepts signals in the function space $\mathcal{F}^{H}_{0}$ as \ref{Diagram:General_Neural_Network}. Each layer $\mathcal{F}^{H}_{i}$ transforms in the $\rho_{i}$ representation of the group $H$. Each layer $\mathcal{F}^{G}_{j}$ transforms in the $\sigma_{j}$ representation of the group $G$. Each map $\Phi_{i} \in \Hom_{H}[ ( \rho_{i} , \mathcal{F}^{H}_{i} ) , (\rho_{i+1} , \mathcal{F}^{H}_{i+1} )]$ is an intertwiner of $H$ representations. Each map $\Psi_{i} \in \Hom_{G}[ ( \sigma_{i} , \mathcal{F}^{G}_{i} ) , (\sigma_{i+1} , \mathcal{F}^{G}_{i+1} )]$ is an intertwiner of $G$ representations. All layers preceding the induced mapping are $H$-equivariant. All layers succeeding the induced mapping are $G$-equivariant.
	
	Uniformly $G$-equivariant networks are the topic of a significant amount of research. End to end $G$-equivariant networks can be essentially fully categorized \cite{Lang_2020}. Each layer is labeled by the number of multiplicity of irreducibles that it falls into and the non-linear activation function. Thus, an architectures of the form \ref{Diagram:General_Neural_Network} can be completely specified by decomposition of each layer into irreducibles
	\begin{align*}
		&( \rho_{0} ,  \mathcal{F}^{H}_{0} ) = \bigoplus_{\rho \in \hat{H} } m_{0 \rho } (\rho , V_{\rho}  )  \\
		&(\rho_{1} , \mathcal{F}^{H}_{1} )= \bigoplus_{\rho \in \hat{H} } m_{1 \rho} (\rho , V_{\rho}  ),  \quad (\rho_{2} , \mathcal{F}^{H}_{2} ) = \bigoplus_{\rho \in \hat{H} } m_{2 \rho } (\rho , V_{\rho}   ),  \quad  ... , \quad 	(\rho_{i} , \mathcal{F}^{H}_{i} )= \bigoplus_{\rho \in \hat{H} } m_{i \rho } (\rho , V_{\rho}  )  \\
		& (\sigma_{1} , \mathcal{F}^{G}_{1} )= \bigoplus_{ \sigma  \in \hat{G} } n_{1 \tau } ( \sigma   , W_{\sigma }  ), \quad	(\sigma_{2} , \mathcal{F}^{G}_{2} ) = \bigoplus_{ \sigma   \in \hat{G} } n_{2 \sigma   } (\sigma   , W_{\sigma } ), \quad ... , \quad 	(\sigma_{j} , \mathcal{F}^{G}_{j} ) = \bigoplus_{ \sigma  \in \hat{G} } n_{j\sigma  } (\sigma   , W_{\sigma} ) 
	\end{align*}
	where $m_{i,\rho}$ are the multiplicities of the $H$-irreducible $( \rho , V_{\rho}) $ in the $i$-th $H$-equivariant layer and $n_{j,\sigma}$ are the multiplicities of the $G$-irreducible $( \sigma , W_{\sigma})$ in the $j$-th $G$-equivariant layer. \cite{Kondor_2018} introduced the concept of \emph{fragments}, which label how a layer breaks into irreducibles. For networks that are initially $H$-equivariant but downstream $G$-equivariant, we need to specify the group as well as the fragment type. 
	
	A induced representation based network is characterized by the non-linearities and $(i+1)$ $H$-fragments and $j$ $G$-fragments,
	\begin{align*}
		&\text{$H$-Equivariant Input Space: }( m_{0,1}  , 	m_{0,2} , ... 	m_{0,|\hat{H}|} ) \\
		& \text{$H$-Equivariant Layers: } ( m_{1,1}  , 	m_{1,2} , ... 	m_{1,|\hat{H}|} ) \enspace ( m_{1,1}  ,  m_{1,2} , ... 	m_{1,|\hat{H}|} ) \enspace ...\enspace ( m_{i,1}  , 	m_{i,2} , ... 	m_{i,|\hat{H}|} ) \\
		&\text{$G$-Equivariant Layers: }  ( n_{1,1}  , 	n_{1,2} , ... 	n_{1,|\hat{G}|} ) , \enspace ( n_{1,1}  , 	n_{1,2} , ... 	n_{1,|\hat{G}|} ) \enspace ...\enspace ( n_{i,1}  , 	n_{i,2} , ... 	n_{i,|\hat{G}|} ) 
	\end{align*}
	where each of the $i$ $H$-equivariant layers is specified by a fragment $( m_{x,1}  , 	m_{x,2} , ... 	m_{x,|\hat{H}|} )$ which specifies the decomposition of the $x$-th layer into $H$-irreducibles. Similarly, each of the $j$ $G$-equivariant layers is specified by a fragment $( n_{y,1}  , 	n_{y,2} , ... 	n_{y,|\hat{G}|} )$ which specifies the decomposition of the $y$-th layer into $G$-irreducibles. The fragments $\enspace ( m_{i,1}  , 	m_{i,2} , ... 	m_{i,|\hat{H}|} )$ and $( n_{1,1}  , 	n_{1,2} , ... 	n_{1,|\hat{G}|} )$ can not be arbitrarily chosen and are related by induced and restriction representations. Specifically, the linear maps between boundary layers must satisfy,
	\begin{align*}
	\Psi \in \Hom_{H}[   (\rho_{i} , \mathcal{F}^{H}_{i} ) , \Res_{H}^{G}[ (\sigma_{1} , \mathcal{F}_{1}^{G} )  ] ] \cong \Hom_{G}[    \Ind_{H}^{G}[ (\rho_{i} , \mathcal{F}^{H}_{i} ) ] , (\sigma_{1} , \mathcal{F}_{1}^{G} )    ]
	\end{align*}
	Specifically, if $( \rho_{i} , \mathcal{F}^{H}_{i})$ and $( \sigma_{1} , \mathcal{F}^{G}_{1})$ decompose into irreducibles as
	\begin{align*}
	 ( \rho_{i} ,	\mathcal{F}^{H}_{i} ) = \bigoplus_{ \rho \in \hat{H} } m_{i \rho } (\rho , V_{\rho} ) , \quad \quad (\sigma_{1} , \mathcal{F}^{G}_{1} ) = \bigoplus_{ \sigma  \in \hat{G} } n_{1 \sigma } ( \sigma  , W_{\sigma}  ) 
	\end{align*}
    Then, we can write the induced and restricted representations in terms of the branching and induction rules,
    \begin{align*}
    \Res_{H}^{G}[ ( \sigma_{1} , \mathcal{F}^{G}_{1})  ] = \bigoplus_{ \rho \in \hat{H} }[ (\sum_{  \sigma \in \hat{G} } n_{1 \sigma } B_{ \sigma, \rho} ) ( \rho , V_{\rho} ) ] \quad \Ind_{H}^{G}[ ( \rho_{i} , \mathcal{F}^{H}_{i} )    ] = \bigoplus_{ \sigma \in \hat{G} }[ (\sum_{  \rho \in \hat{H} } m_{ i, \rho } I_{\rho, \sigma}) ( \sigma , W_{\sigma} ) ]
    \end{align*}

\subsubsection{Generalization to Multiple Groups}
We have chosen to consider the case where we induce directly from $H \subset G $ to $G$. It should be noted that this induction procedure can also be performed incrementally for any sequence of nested ascending subgroups $H = G_{1} \subset G_{2} ... \subset G_{N-1} \subset G = G_{N}$. A network architecture is then completely specified by a set of layers that decompose into $G_{i}$-irreducibles,
\begin{align*}
&( \rho^{G_{1}}_{0} , \mathcal{F}^{G_{1}}_{0} ) = \bigoplus_{\sigma \in \hat{G}_{1} } n^{G_{1}}_{0 \sigma } (\sigma , V_{\sigma}), \quad ( \rho^{G_{1}}_{1} , \mathcal{F}^{G_{1}}_{1} ) = \bigoplus_{\sigma \in \hat{G}_{1} } n^{G_{1}}_{1 \sigma } (\sigma , V_{\sigma}) , \quad ... \quad ( \rho^{G_{1}}_{i_{1}} , \mathcal{F}^{G_{1}}_{i_{1}} ) = \bigoplus_{\sigma \in \hat{G}_{1} } n^{G_{1}}_{i_{1} \sigma } (\sigma , V_{\sigma}) \\
&( \rho^{G_{2}}_{1} , \mathcal{F}^{G_{2}}_{1} )  = \bigoplus_{\sigma \in \hat{G}_{2} } n^{G_{2}}_{1 \sigma } (\sigma , V_{\sigma}),  \quad ( \rho^{G_{2}}_{2} , \mathcal{F}^{G_{2}}_{2} )  = \bigoplus_{\sigma \in \hat{G}_{2} } n^{G_{2}}_{2 \sigma } (\sigma , V_{\sigma}),  \quad  ... \quad 	( \rho^{G_{2}}_{i_{2}} , \mathcal{F}^{G_{2}}_{i_{2}} )  = \bigoplus_{\sigma \in \hat{G}_{2} } n^{G_{2}}_{i_{2} \sigma } (\sigma , V_{\sigma}), \\
&...\\
& (\rho^{G_{N}}_{1} , \mathcal{F}^{G_{N}}_{1} ) = \bigoplus_{\sigma \in \hat{G}_{N} } n^{G_{N}}_{1 \sigma } (\sigma , V_{\sigma}), \quad	(\rho^{G_{N}}_{2} , \mathcal{F}^{G_{N}}_{2} ) = \bigoplus_{\sigma \in \hat{G}_{N} } n^{G_{N}}_{2 \sigma } (\sigma , V_{\sigma}), \quad ... \quad 	(\rho^{G_{N}}_{i_{N}} , \mathcal{F}^{G_{N}}_{i_{N}} ) = \bigoplus_{\sigma \in \hat{G}_{N} } n^{G_{N}}_{i_{N} \sigma } (\sigma , V_{\sigma})
\end{align*}	

Let $\Psi^{B}_{i}$ be the intertwiner at the $i$-th boundary layer. The equivarience conditions require that
\begin{align*}
& \Psi^{B}_{1} \in \Hom_{G_{1}}[   (\rho^{G_{1}}_{i_{1}} , \mathcal{F}^{G_{1}}_{i_{1}} ) , \Res_{G_{1}}^{G_{2}}[ (\rho^{G_{2}}_{i_{2}} , \mathcal{F}^{G_{2}}_{i_{2}} )  ] ] \cong \Hom_{G_{2}}[    \Ind_{G_{1}}^{G_{2}}[ (\rho^{G_{1}}_{i_{1}} , \mathcal{F}^{G_{1}}_{i_{1}} ) ] , (\rho^{G_{2}}_{i_{2}} , \mathcal{F}^{G_{2}}_{i_{2}} )    ] \\
& \Psi^{B}_{2} \in \Hom_{G_{2}}[   (\rho^{G_{2}}_{i_{2}} , \mathcal{F}^{G_{2}}_{i_{2}} ) , \Res_{G_{2}}^{G_{3}}[ (\rho^{G_{3}}_{i_{3}} , \mathcal{F}^{G_{3}}_{i_{3}} )  ] ] \cong \Hom_{G_{3}}[    \Ind_{G_{2}}^{G_{3}}[ (\rho^{G_{2}}_{i_{2}} , \mathcal{F}^{G_{2}}_{i_{2}} ) ] , (\rho^{G_{3}}_{i_{3}} , \mathcal{F}^{G_{3}}_{i_{3}} )    ] \\
& ... \\
& \Psi^{B}_{N-1} \in \Hom_{G_{N-1}}[   (\rho^{G_{N-1}}_{i_{N-1}} , \mathcal{F}^{G_{N-1}}_{i_{N-1}} ) , \Res_{G_{N-1}}^{G_{N}}[ (\rho^{G_{N}}_{i_{N}} , \mathcal{F}^{G_{N}}_{i_{N}} )  ] ] \cong \Hom_{G_{N}}[    \Ind_{G_{N-1}}^{G_{N}}[ (\rho^{G_{N-1}}_{i_{N-1}} , \mathcal{F}^{G_{N-1}}_{i_{N-1}} ) ] , (\rho^{G_{N}}_{i_{N}} , \mathcal{F}^{G_{N}}_{i_{N}} )    ] 
\end{align*}

Let $I^{G_{i}G_{i+1}} : \hat{G}_{i} \times \hat{G}_{i+1} \rightarrow \mathbb{Z}^{\geq0}$ and $B^{G_{i}G_{i+1}} : \hat{G}_{i+1} \times \hat{G}_{i} \rightarrow \mathbb{Z}^{\geq0}$ be the induction rules and the branching rules for the groups $G_{i} \subset G_{i+1}$, respectively. Then, we can write the induced and restricted representations at each layer in terms of the branching and induction rules,
\small
\begin{flalign*}
& \Res_{G_{1}}^{G_{2}}[ ( \rho^{G_{2}}_{i_{2}} , \mathcal{F}^{G_{2}}_{i_{2}})  ] = \bigoplus_{ \rho \in \hat{G}_{1} }[ (\sum_{  \sigma \in \hat{G}_{2} } n^{G_{2}}_{1 \sigma } B^{G_{1}G_{2}}_{ \sigma, \rho} ) ( \rho , V_{\rho} ) ], \quad \Ind_{G_{1}}^{G_{2}}[ ( \rho^{G_{1}}_{i_{1}} , \mathcal{F}^{G_{1}}_{i_{1}} )    ] = \bigoplus_{ \rho \in \hat{G}_{2} }[ (\sum_{  \sigma \in \hat{G}_{1} } n^{G_{1}}_{ i_{1}, \sigma } I^{G_{1}G_{2}}_{\sigma, \rho}) ( \rho , V_{\rho} ) ] \\
&  \Res_{G_{2}}^{G_{3}}[ ( \rho^{G_{3}}_{i_{3}} , \mathcal{F}^{G_{3}}_{i_{3}})  ] = \bigoplus_{ \rho \in \hat{G}_{2} }[ (\sum_{  \sigma \in \hat{G}_{3} } n^{G_{3}}_{1 \sigma } B^{G_{2}G_{3}}_{ \sigma, \rho} ) ( \rho , V_{\rho} ) ], \quad \Ind_{G_{2}}^{G_{3}}[ ( \rho^{G_{2}}_{i_{2}} , \mathcal{F}^{G_{2}}_{i_{2}} )    ] = \bigoplus_{ \rho \in \hat{G}_{3} }[ (\sum_{  \sigma \in \hat{G}_{2} } n^{G_{2}}_{ i_{2}, \sigma } I^{G_{2}G_{3}}_{\sigma, \rho}) ( \rho , V_{\rho} ) ] \\
&  ... \\
& \hspace{-2.0cm} \Res_{G_{N-1}}^{G_{N}}[ ( \rho^{G_{N}}_{i_{N}} , \mathcal{F}^{G_{N}}_{i_{N}})  ] = \bigoplus_{ \rho \in \hat{G}_{N-1} }[ (\sum_{  \sigma \in \hat{G}_{N} } n^{G_{N}}_{1 \sigma } B^{G_{N-1}G_{N}}_{ \sigma, \rho} ) ( \rho , V_{\rho} ) ], \quad \Ind_{G_{N-1}}^{G_{N}}[ ( \rho^{G_{N-1}}_{i_{N-1}} , \mathcal{F}^{G_{N-1}}_{i_{N-1}} )    ] = \bigoplus_{ \rho \in \hat{G}_{N} }[ (\sum_{  \sigma \in \hat{G}_{N-1} } n^{G_{N-1}}_{ i_{N-1}, \sigma } I^{G_{N-1}G_{N}}_{\sigma, \rho}) ( \rho , V_{\rho} ) ] 
\end{flalign*}
\normalsize
Thus, the induced representation allows for the design of networks that are equivariant with respect a sequence of ascending nested larger groups. It should be noted that it is also possible to move in the `other direction'. The restriction representation can be used for \emph{coset pooling} \cite{Weiler_2021} to design networks that are equivariant with respect to a descending sequence of nested subgroups $G'_{1} \supset G'_{2} \supset ... \supset G'_{N}$. Thus, the induced representation, combined with coset pooling allow for the design of neural networks that are at different stages equivariant with respect to an arbitrary sequence of groups $G_{1}, G_{2},..., G_{N}$, so long as each group in the sequence either contains or is contained by the previous group.

\begin{wrapfigure}{r}{0.40\textwidth}
\vspace{-1.9cm}
\includegraphics[width=0.40\textwidth]{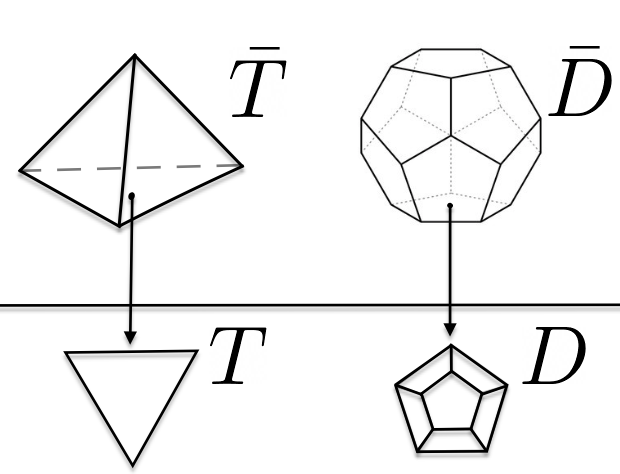}
\caption{ \small Left: Three dimensional tetrahedron $\bar{T}$ with symmetry group $A_{4}$. The projection of $\bar{T}$ into a plane is an equilateral triangle $T$. The symmetry group of $T$ is $\mathbb{Z}_{3}$. Right: Three dimensional dodecehedron $\bar{D}$ with symmetry group $A_{5}$. The projection of $\bar{D}$ into a plane is an pentagon $D$. The symmetry group of $T$ is $\mathbb{Z}_{5}$. }\label{Figure:Projection_Map}
\vspace{-1.0cm}
\end{wrapfigure}

\section{Toy Example: Tetrahedral Signals}\label{Appendix:Section:Tetra_Explict}

We work out one toy example to help build intuition for induced representations.

Let $\bar{T}$ denote a tetrahedron in three dimensional space. $\bar{T}$ is composed of four vertices and four equilateral triangular faces. Let $T$ be the projection of $\bar{T}$ in a direction normal to a face of $\bar{T}$. As show in \ref{Figure:Projection_Map}, the image of a projection in a direction normal to a face is a equilateral triangle which we will call $T$. The induced representation has a natural geometric interpretation that relates the symmetry subgroup of the projected platonic solid $T$ to the full Platonic solid $\bar{T}$. The same argument presented here for the dodecehedron $\bar{D}$ recovers the results of \cite{Esteves_2019}.

The group of orientation preserving symmetries of the equilateral triangle $T$ is $\mathbb{Z}_{3}$ which corresponds to rotations through the origin an angle of $0$, $\frac{2\pi}{3}$ or $\frac{4\pi}{3}$. The group of orientation preserving symmetries of $\bar{T}$ is $A_{4}$.

Let $f: T \rightarrow \mathbb{R}^{c}$ be a signal defined on $T$. Take $\{\Phi_{k}\}_{k=1}^{4}$ to be four independent filters with $\Phi_{k} : T \rightarrow \mathbb{R}^{K \times c}$ each transforming in the \emph{same} representation of $\mathbb{Z}_{3}$. We can then convolve each $\Phi_{k}$ with $f$,
\begin{align*}
    \forall g \in \mathbb{Z}_{3}, \quad	\Psi_{k}(g) = ( \Phi_{k} \star f )(g) = \int_{x \in T} \Phi_{k}(x)f(g^{-1}x)
\end{align*}
so that each $ \Psi_{k} : \mathbb{Z}_{3} \rightarrow \mathbb{R}^{K} \in ( \mathbb{R}^{K} )^{\mathbb{Z}_{3}}$. The group $\mathbb{Z}_{3}$ has action on each $ \Psi_{k}  $. Now, let us vectorize the $\Psi_{k}$ group valued functions into one variable $\Psi$ with $\Psi : \mathbb{Z}^{3} \rightarrow \mathbb{R}^{4K}$,
\begin{align*}
    g \in \mathbb{Z}_{3} , \quad \Psi(g) =  \begin{bmatrix}
        \Psi_{1}(g) \\
        \Psi_{2}(g)  \\
        \Psi_{3}(g)  \\
        \Psi_{4}(g)  \\
    \end{bmatrix}
\end{align*}

We can now compute the induced action. The computations involved with this map are straightforward but somewhat tedious and are described in \ref{Section:Tetra Group Calculations}. We just state the results in this section. Let $\Psi^{\uparrow}$ be the function defined on $A_{4}$, which has $A_{4}$ induced action. First, consider $\Psi^{\uparrow}$ on elements of $\mathbb{Z}_{3} = \{ e  , (1,2,3) , (1,3,2) \}$,
\begin{align*}
    \Psi^{\uparrow}[ e ] = \begin{bmatrix}
        \Psi_{1}[e] \\
        \Psi_{2}[e]  \\
        \Psi_{3}[e]  \\
        \Psi_{4}[e]  \\
    \end{bmatrix}, \quad \Psi^{\uparrow}[ (1,2,3) ] = \begin{bmatrix}
        \Psi_{1}[ (1,2,3)] \\
        \Psi_{4}[(1,2,3)]  \\
        \Psi_{2}[(1,2,3)]  \\
        \Psi_{3}[(1,2,3)]  \\
    \end{bmatrix} \quad \Psi^{\uparrow}[ (1,3,2) ] = \begin{bmatrix}
        \Psi_{1}[ (1,3,2)] \\
        \Psi_{3}[(1,3,2)]  \\
        \Psi_{4}[(1,3,2)]  \\
        \Psi_{2}[(1,3,2)]  \\
    \end{bmatrix} 
\end{align*} 
Note that on $\mathbb{Z}_{3}$ coset $\Psi^{\uparrow}$ acts only via permutations.

Now, consider the $(1,2,4)H$ coset, we have that
\begin{align*}
    \Psi^{\uparrow}[ (1,2,4) ] = \begin{bmatrix}
        \Psi_{2}[e] \\
        \Psi_{4}[(1,3,2)]  \\
        \Psi_{3}[(1,3,2)]  \\
        \Psi_{1}[(1,2,4)]  \\
    \end{bmatrix}, \quad \Psi^{\uparrow}[ (1,3)(2,4) ] = \begin{bmatrix}
        \Psi_{2}[(1,2,3)] \\
        \Psi_{1}[(1,3,2)]  \\
        \Psi_{4}[e]  \\
        \Psi_{3}[e]  \\
    \end{bmatrix} \quad \Psi^{\uparrow}[ (2,4,3) ] = \begin{bmatrix}
        \Psi_{2}[(1,3,2)] \\
        \Psi_{3}[(1,2,3)]  \\
        \Psi_{1}[e]  \\
        \Psi_{4}[(1,2,3)]  \\
    \end{bmatrix} 
\end{align*}
Similarly, for the $(2,3,4)H$ coset, we have that,
\begin{align*}
    \Psi^{\uparrow}[ (2,3,4) ] = \begin{bmatrix}
        \Psi_{3}[e] \\
        \Psi_{1}[(1,2,3)]  \\
        \Psi_{2}[(1,3,2)]  \\
        \Psi_{4}[(1,3,2)]  \\
    \end{bmatrix}, \quad \Psi^{\uparrow}[ (1,2)(3,4) ] = \begin{bmatrix}
        \Psi_{3}[(1,2,3)] \\
        \Psi_{4}[e]  \\
        \Psi_{1}[(1,3,2)]  \\
        \Psi_{2}[e]  \\
    \end{bmatrix} \quad \Psi^{\uparrow}[ (3,4,1) ] = \begin{bmatrix}
        \Psi_{3}[(1,3,2)] \\
        \Psi_{2}[(1,2,3)]  \\
        \Psi_{4}[(1,2,3)]  \\
        \Psi_{1}[e]  \\
    \end{bmatrix} 
\end{align*}
Lastly for the $(3,1,4)H$ coset, we have that	
\begin{align*}
    \Psi^{\uparrow}[ (3,1,4) ] = \begin{bmatrix}
        \Psi_{4}[e] \\
        \Psi_{2}[(1,3,2)]  \\
        \Psi_{1}[(1,2,3)]  \\
        \Psi_{3}[(1,3,2)]  \\
    \end{bmatrix}, \quad \Psi^{\uparrow}[ (2,3)(1,4) ] = \begin{bmatrix}
        \Psi_{4}[(1,2,3)] \\
        \Psi_{3}[e]  \\
        \Psi_{2}[e]  \\
        \Psi_{1}[(1,3,2)]  \\
    \end{bmatrix} \quad \Psi^{\uparrow}[ (1,4,2) ] = \begin{bmatrix}
        \Psi_{4}[(1,3,2)] \\
        \Psi_{1}[e]  \\
        \Psi_{3}[(1,2,3)]  \\
        \Psi_{2}[(1,2,3)]  \\
    \end{bmatrix} 
\end{align*}
Thus, we have constructed a function $\Psi^{\uparrow}: A_{4} \rightarrow \mathbb{R}^{4K}$ from a set of four filters $\Phi_{k}: T \rightarrow \mathbb{R}^{K \times c}$ defined on the triangle $T$. The important observation is that the group $A_{4}$ acts on $\Psi^{\uparrow}$ via permutation and action by an element $\mathbb{Z}_{3} \subset A_{4}$. This is the same as the induced representation which has $G$-action that is a mix of permutation and $H$-action \ref{Appendix:Section:Induced Representation for Finite Groups}. It should be noted that unlike the projection trick used in \cite{Klee_2022}, this construction requires no padding or projections. Furthermore, it is not even required that the signal $f$ be lifted from $T$ into $\bar{T}$.

\subsubsection{Comparison With Orthographic Projection}

In analogy with \cite{Klee_2023,Klee_2022,Esteves_2019}, another way to create a signal on $\bar{T}$ would be to first lift the signal from $T$ to $\bar{T}$ via orthographic projection and then use an $A_{4}$-equivariant neural network to extract features. Note that this approach is a specific instance of our construction in \ref{Appendix:Section:Tetra_Explict} and corresponds to setting 
\begin{align*}
    \Phi_{1} = \Phi(x) \quad \Phi_{2} = \Phi_{3} = \Phi_{4} = 0
\end{align*}
where $\Phi(x):T \rightarrow T$ is a feature map defined on the equilateral triangle. With this choice of $\Phi_{k}$, occluded faces of the tetrahedron have no signal defined on them.

\section{Group Calculations for Induced Representation of $\mathbb{Z}_{3}$ to $A_{4}$}\label{Section:Tetra Group Calculations}

This section details the calculations in computing induced representations of $\mathbb{Z}_{3}$ on $A_{4}$. Computations were done with symbolic computer program, which is available upon request. Let us take $\mathbb{Z}_{3} \subset A_{4}$ to be the group
\begin{align*}
\mathbb{Z}_{3} = \langle (1,2,3) \rangle = \{ e , (1,2,3) , (1,3,2) \}
\end{align*}
Let us calculate the representatives of the four left cosets of $ A_{4} / \mathbb{Z}_{3}$. We have that
\small
\begin{align*}
& e \cdot \mathbb{Z}_{3} = \{ e , (1,2,3) , (1,3,2) \} \\
& ( 1 , 2 ,  4 ) \cdot \mathbb{Z}_{3} = \{ (1,2,4) , (1,3) (2,4) , (2,4,3)  \}  \\
& ( 2, 3 , 4 ) \cdot  \mathbb{Z}_{3} = \{ (2,3,4) , (1,2)(3,4) , (3,4,1) \}   \\
& (3 , 1 , 4) \cdot \mathbb{Z}_{3} = \{ (1,4,3), (2,3)(1,4) , (1,4,2) \} 
\end{align*}
\normalsize
Thus, the elements $ g_{1} = e$, $g_{2} =  ( 1 , 2 ,  4 )$, $g_{3} =  ( 2, 3 , 4 ) $, $g_{4} = (3 , 1 , 4)$ are representatives of $A_{4} / \mathbb{Z}_{3}$. Now, we know that,
\begin{align*}
\forall g \in A_{4}, \quad \forall g_{i} \in \{g_{1},g_{2},g_{3},g_{4}\} , \quad \exists h_{i}(g) \in \mathbb{Z}_{3} \text{ s.t. } g \cdot g_{i} = g_{j_{g}(i)} h_{i}(g)
\end{align*}

where $j_{g}$ is a permutation and $h_{i}(g) \in H$.
We thus need to compute the permutations $j_{g} \in S_{4} : \{1,2,3,4\} \rightarrow \{1,2,3,4\}$ and $h_{i}(g) \in H$. The identity element coset has
\small
\begin{align*}
& j_{e} = \begin{bmatrix}
    1 & 2 & 3 & 4 \\
    1 & 2 & 3 & 4 \\
\end{bmatrix}, \quad j_{(1,2,3)} = \begin{bmatrix}
    1 & 2 & 3 & 4 \\
    1 & 4 & 2 & 3 \\
\end{bmatrix},  \quad j_{(1,3,2)} = \begin{bmatrix}
    1 & 2 & 3 & 4 \\
    1 & 3 & 4 & 2 \\
\end{bmatrix}, \\
&h(e) = \begin{bmatrix}
    1 & 2 & 3 & 4 \\
    e & e & e & e \\
\end{bmatrix}, \\
& h(1,2,3) = \begin{bmatrix}
    1 & 2 & 3 & 4 \\
    (1,2,3) & (1,2,3) & (1,2,3) & (1,2,3) \\
\end{bmatrix}, \\
& h(1,3,2) = \begin{bmatrix}
1 & 2 & 3 & 4 \\
(1,3,2) & (1,3,2) & (1,3,2) & (1,3,2) 
\end{bmatrix}
\end{align*}
\normalsize
Now, for the $g_{2} = ( 1 , 2 ,  4 )$ coset,
\small
\begin{align*}
& j_{( 1 , 2 ,  4 )} = \begin{bmatrix}
1 & 2 & 3 & 4 \\
2 & 4 & 3 & 1 \\
\end{bmatrix}, \quad j_{(1,3)(2,4)} = \begin{bmatrix}
1 & 2 & 3 & 4 \\
2 & 1 & 4 & 3 \\
\end{bmatrix},  \quad j_{(2,4,3)} = \begin{bmatrix}
1 & 2 & 3 & 4 \\
2 & 3 & 1 & 4 \\
\end{bmatrix}, \\
&h(1,2,4) = \begin{bmatrix}
        1 & 2 & 3 & 4 \\
        e & (1,3,2) & (1,3,2) & (1,2,3) \\
    \end{bmatrix}, \\
    & h( (1,3)(2,4) ) = \begin{bmatrix}
        1 & 2 & 3 & 4 \\
        (1,2,3) & (1,3,2) & e & e \\
    \end{bmatrix}, \\
    & h(2,4,3) = \begin{bmatrix}
        1 & 2 & 3 & 4 \\
        (1,3,2) & (1,2,3) & e & (1,2,3) 
    \end{bmatrix}
\end{align*}
\normalsize
Similarly, for the $( 2, 3 , 4 ) $ coset,
\small
\begin{align*}
    & j_{( 2 , 3 ,  4 )} = \begin{bmatrix}
        1 & 2 & 3 & 4 \\
        3 & 1 & 2 & 4 \\
    \end{bmatrix}, \quad j_{(1,2)(3,4)} = \begin{bmatrix}
        1 & 2 & 3 & 4 \\
        3 & 4 & 1 & 2 \\
    \end{bmatrix},  \quad j_{(3,4,1)} = \begin{bmatrix}
        1 & 2 & 3 & 4 \\
        3 & 2 & 4 & 1 \\
    \end{bmatrix}, \\
    &h(2,3,4) = \begin{bmatrix}
        1 & 2 & 3 & 4 \\
        e & (1,2,3) & (1,3,2) & (1,3,2) \\
    \end{bmatrix}, \\
    & h( (1,2)(3,4) )= \begin{bmatrix}
        1 & 2 & 3 & 4 \\
        (1,2,3) & e & (1,3,2) & e \\
    \end{bmatrix}, \\
    & h(3,4,1) = \begin{bmatrix}
        1 & 2 & 3 & 4 \\
        (1,3,2) & (1,2,3) & (1,2,3) & e 
    \end{bmatrix}
\end{align*}
\normalsize
And lastly for the $(1,4,3)$ coset,
\small
\begin{align*}
    & j_{( 1 , 4 ,  3 )} = \begin{bmatrix}
        1 & 2 & 3 & 4 \\
        4 & 2 & 1 & 3 \\
    \end{bmatrix}, \quad j_{(2,3)(1,4)} = \begin{bmatrix}
        1 & 2 & 3 & 4 \\
        4 & 3 & 2 & 1 \\
    \end{bmatrix},  \quad j_{(1,4,2)} = \begin{bmatrix}
        1 & 2 & 3 & 4 \\
        4 & 1 & 3 & 2 \\
    \end{bmatrix}, \\
    &h(1,4,3) = \begin{bmatrix}
        1 & 2 & 3 & 4 \\
        e & (1,3,2) & (1,2,3) & (1,3,2) \\
    \end{bmatrix}, \\
    & h( (2,3)(1,4) ) = \begin{bmatrix}
        1 & 2 & 3 & 4 \\
        (1,2,3) & e & e & (1,3,2) \\
    \end{bmatrix}, \\
    & h(1,4,2) = \begin{bmatrix}
        1 & 2 & 3 & 4 \\
        (1,3,2) & e & (1,2,3) & (1,2,3) 
    \end{bmatrix}
\end{align*}
\normalsize
Now that we have explicit formulae for $j_{g}$ and $h(g)$ we can construct the induction of a function from domain $\mathbb{Z}_{3}$ to $A_{4}$.

\subsection{Counting Degrees of Freedom}

$\mathbb{Z}_{3}$ has three one dimensional irreducible representations $( \rho_{1}, V_{1}) , (\rho_{+}, V_{+} )$ and $( \rho_{-}, V_{-} )$. The actions are given by
\begin{align*}
    & v \in V_{1}, \quad	\rho_{1}(g)v = v  \\
    & \enspace v \in V_{\pm}, \quad	\rho_{\pm}(g)v = \exp(  \pm \frac{2\pi i   }{3} )v 
\end{align*}
where $(\rho_{1},V_{1})$ is the trivial representation and $( \rho_{+} , V_{+} )$ and $( \rho_{-}, V_{-})$ are conjugate representations.

We can now find the induced representation of $(\rho_{k} , V_{k})$ on $A_{4}$. The index is given by $| A_{4} : \mathbb{Z}_{3} | = 4$. Let $g_{1},g_{2},g_{3},g_{4}$ be representatives of the four left cosets in $A_{4} / \mathbb{Z}_{3}$. So that
\begin{align}\label{Left_Coset_Decomposition}
    A_{4} / \mathbb{Z}_{3} = \{ g_{1} \mathbb{Z}_{3} , g_{2} \mathbb{Z}_{3} , g_{3} \mathbb{Z}_{3}, g_{4} \mathbb{Z}_{3} \}
\end{align}
Note that $\mathbb{Z}_{3}$ is not normal in $A_{4}$ so $	A_{4} / \mathbb{Z}_{3}$ is not a group. Despite this, the decomposition in \eqref{Left_Coset_Decomposition} holds, via the fact that the set of representatives of cosets partitions $G$. The induced representation of the irreducible $( \rho_{k} , V_{k} )$ representation of $\mathbb{Z}_{3}$ on $A_{4}$ acts on the vector space
\begin{align*}
    k\in \{1,+,-\}, \quad W_{k} =	\text{Ind}_{ \mathbb{Z}_{3} }^{A_{4}} ( V_{k} )  = \bigoplus_{i=1}^{4} g_{i} V^{(i)}_{k}
\end{align*}
were the notation $g_{i}V^{(i)}_{k}$ is a label denoting the $i$-th independent copy of the vector space $V_{k}$. Let $R_{k} = \text{Ind}_{ \mathbb{Z}_{3} }^{A_{4}}( \rho_{k} )$ denote the action of $A_{4}$ on $W_{k}$. We have that,
\begin{align*}
    \forall g \in A_{4}, \quad R_{k}(g) \cdot \sum_{i=1}^{4} g_{i} v_{i} = \sum_{i=1}^{4} g_{ j_{g}(i) } \rho_{k}( h_{i}(g) ) v_{i} \in W_{k}
\end{align*}
where $\forall g \in A_{4}$, $j_{g}(i)\in S_{4} : \{1,2,3,4\} \rightarrow \{1,2,3,4\}$ is a permutation of the coset representatives and $h_{i}(g) \in \mathbb{Z}_{3}$. To summarize, irreducible representations of $\mathbb{Z}_{3} = \langle g \rangle $ are given by $(\rho_{k} , V_{k})$ with
\begin{align*}
    & v \in V_{1}, \quad	\rho_{1}(g)v = v \\
    & v \in V_{\pm}, \quad	\rho_{\pm}(g)v = \exp( \frac{\pm 2\pi i   }{3} )v
\end{align*}
The induced representations of $\mathbb{Z}_{3}$ on $A_{4}$ are given by $(R_{k} , W_{k})$ with
\begin{align*}
    &k \in \{1,+,-\}, \quad W_{k} = \bigoplus_{i=1}^{4} g_{i} V^{(i)}_{k} \\
    & R_{k}(g) \cdot \sum_{i=1}^{4} g_{i} v_{i} =   \sum_{i=1}^{4} g_{j_{g}(i)} \rho_{k}( h_{i}(g) ) v_{i}  \\
    & \text{with } g \cdot g_{i} = g_{j_{g}(i)} \cdot h_{i}(g) 
\end{align*}
Let us explicitly construct the induced representation of each irreducible of $\mathbb{Z}_{3}$ explicitly.

\subsubsection{Trivial Representation $( \rho_{1}, V_{1})$}
Consider first the trivial representation $( \rho_{1}, V_{1})$ of $\mathbb{Z}_{3}$. The induced action $R_{1} = \text{Ind}^{A_{4}}_{\mathbb{Z}_{3}}( \rho_{1} )$ is then given by
\small
\begin{align*}
&	R_{1}[e] \cdot \begin{bmatrix}
        v_{1} \\
        v_{2} \\
        v_{3} \\
        v_{4} \\
    \end{bmatrix} = \begin{bmatrix}
        v_{1} \\
        v_{2} \\
        v_{3} \\
        v_{4} \\
    \end{bmatrix} \quad 		R_{1}[(1,2,3)] \cdot \begin{bmatrix}
        v_{1} \\
        v_{2} \\
        v_{3} \\
        v_{4} \\
    \end{bmatrix} = \begin{bmatrix}
        v_{1} \\
        v_{4} \\
        v_{2} \\
        v_{3} \\
    \end{bmatrix} 	\quad	 R_{1}[(1,3,2)] \cdot \begin{bmatrix}
        v_{1} \\
        v_{2} \\
        v_{3} \\
        v_{4} \\
    \end{bmatrix} = \begin{bmatrix}
        v_{1} \\
        v_{3} \\
        v_{4} \\
        v_{2} \\
    \end{bmatrix}  \\
    &	R_{1}[(1,2,4)] \cdot \begin{bmatrix}
        v_{1} \\
        v_{2} \\
        v_{3} \\
        v_{4} \\
    \end{bmatrix} = \begin{bmatrix}
        v_{2} \\
        v_{4} \\
        v_{3} \\
        v_{1} \\
    \end{bmatrix} \quad 		R_{1}[(1,3)(2,4)] \cdot \begin{bmatrix}
        v_{2} \\
        v_{1} \\
        v_{4} \\
        v_{3} \\
    \end{bmatrix} = \begin{bmatrix}
        v_{1} \\
        v_{2} \\
        v_{3} \\
        v_{4} \\
    \end{bmatrix} 	\quad  R_{1}[(2,4,3)] \cdot \begin{bmatrix}
        v_{1} \\
        v_{2} \\
        v_{3} \\
        v_{4} \\
    \end{bmatrix} = \begin{bmatrix}
        v_{2} \\
        v_{3} \\
        v_{1} \\
        v_{4} \\
    \end{bmatrix}  \\
    &	R_{1}[(2,3,4)] \cdot \begin{bmatrix}
        v_{1} \\
        v_{2} \\
        v_{3} \\
        v_{4} \\
    \end{bmatrix} = \begin{bmatrix}
        v_{3} \\
        v_{1} \\
        v_{2} \\
        v_{4} \\
    \end{bmatrix} \quad 		R_{1}[(1,2)(3,4)] \cdot \begin{bmatrix}
        v_{1} \\
        v_{2} \\
        v_{3} \\
        v_{4} \\
    \end{bmatrix} = \begin{bmatrix}
        v_{3} \\
        v_{4} \\
        v_{1} \\
        v_{2} \\
    \end{bmatrix} 	\quad	R_{1}[(3,4,1)] \cdot \begin{bmatrix}
        v_{1} \\
        v_{2} \\
        v_{3} \\
        v_{4} \\
    \end{bmatrix} = \begin{bmatrix}
        v_{3} \\
        v_{2} \\
        v_{4} \\
        v_{1} \\
    \end{bmatrix}  \\ 
    &	R_{1}[(1,4,3)] \cdot \begin{bmatrix}
        v_{1} \\
        v_{2} \\
        v_{3} \\
        v_{4} \\
    \end{bmatrix} = \begin{bmatrix}
        v_{4} \\
        v_{2} \\
        v_{1} \\
        v_{3} \\
    \end{bmatrix} \quad 		R_{1}[(2,3)(1,4)] \cdot \begin{bmatrix}
        v_{1} \\
        v_{2} \\
        v_{3} \\
        v_{4} \\
    \end{bmatrix} = \begin{bmatrix}
        v_{4} \\
        v_{3} \\
        v_{2} \\
        v_{1} \\
    \end{bmatrix} 	\quad	R_{1}[(2,4,3)] \cdot \begin{bmatrix}
        v_{1} \\
        v_{2} \\
        v_{3} \\
        v_{4} \\
    \end{bmatrix} = \begin{bmatrix}
        v_{4} \\
        v_{1} \\
        v_{3} \\
        v_{2} \\
    \end{bmatrix}  
\end{align*}
Working in the standard Euclidean basis, we may write this as
\small
\begin{align*}
    &	R_{1}[e] = \begin{bmatrix}
        1 & 0 & 0 & 0  \\
        0 & 1 & 0 & 0  \\
        0 & 0 & 1 & 0  \\
        0 & 0 & 0 & 1  \\
    \end{bmatrix}  \quad 		R_{1}[(1,2,3)]  = \begin{bmatrix}
        1 & 0 & 0 & 0  \\
        0 & 0 & 0 & 1  \\
        0 & 1 & 0 & 0  \\
        0 & 0 & 1 & 0  \\
    \end{bmatrix}	\quad	 R_{1}[(1,3,2)]  = \begin{bmatrix}
        1 & 0 & 0 & 0  \\
        0 & 0 & 1 & 0  \\
        0 & 0 & 0 & 1  \\
        0 & 1 & 0 & 0  \\
    \end{bmatrix}  \\
    &	R_{1}[(1,2,4)] = \begin{bmatrix}
        0 & 1 & 0 & 0  \\
        0 & 0 & 0 & 1  \\
        0 & 0 & 1 & 0  \\
        1 & 0 & 0 & 0  \\
    \end{bmatrix} \quad 		R_{1}[(1,3)(2,4)] = \begin{bmatrix}
        0 & 1 & 0 & 0  \\
        1 & 0 & 0 & 0  \\
        0 & 0 & 0 & 1  \\
        0 & 0 & 1 & 0  \\
    \end{bmatrix} 	\quad  R_{1}[(2,4,3)] = \begin{bmatrix}
        0 & 1 & 0 & 0  \\
        0 & 0 & 1 & 0  \\
        1 & 0 & 0 & 0  \\
        0 & 0 & 0 & 1  \\
    \end{bmatrix} \\
    &	R_{1}[(2,3,4)] =  \begin{bmatrix}
        0 & 0 & 1 & 0  \\
        1 & 0 & 0 & 0  \\
        0 & 1 & 0 & 0  \\
        0 & 0 & 0 & 1  \\
    \end{bmatrix} \quad 		R_{1}[(1,2)(3,4)] =  \begin{bmatrix}
        0 & 0 & 1 & 0  \\
        0 & 0 & 0 & 1  \\
        1 & 0 & 0 & 0  \\
        0 & 1 & 0 & 0  \\
    \end{bmatrix}	\quad	R_{1}[(3,4,1)]  \begin{bmatrix}
        0 & 0 & 1 & 0  \\
        0 & 1 & 0 & 0  \\
        0 & 0 & 0 & 1  \\
        1 & 0 & 0 & 0  \\
    \end{bmatrix} \\
    &	R_{1}[(1,4,3)] =  \begin{bmatrix}
        0 & 0 & 0 & 1  \\
        0 & 1 & 0 & 0  \\
        1 & 0 & 0 & 0  \\
        0 & 0 & 1 & 0  \\
    \end{bmatrix} \quad 		R_{1}[(2,3)(1,4)]  =  \begin{bmatrix}
        0 & 0 & 0 & 1  \\
        0 & 0 & 1 & 0  \\
        0 & 1 & 0 & 0  \\
        1 & 0 & 0 & 0  \\
    \end{bmatrix} \quad	R_{1}[(2,4,3)]  =  \begin{bmatrix}
        0 & 0 & 0 & 1  \\
        1 & 0 & 0 & 0  \\
        0 & 0 & 1 & 0  \\
        0 & 1 & 0 & 0  
    \end{bmatrix} 
\end{align*}
\normalsize
Note that the induced action of a trivial representation acts only via permutation for all groups.

\subsubsection{$( \rho_{+}, V_{+})$ and $( \rho_{-}, V_{-})$ Representations}

Now, consider the two complex representations $( \rho_{+} , V_{+})$ and $( \rho_{-} , V_{-})$. These representations are conjugate representations,
\begin{align*}
    \overline{ ( \rho_{+} , V_{+}) } = ( \rho_{-} , V_{-}) \quad \overline{( \rho_{-} , V_{-})} = ( \rho_{+} , V_{+})
\end{align*} The induced representation of the conjugate is the conjugate of the induced representation,
\begin{align*}
    \text{Ind}_{H}^{G}[  \overline{(\rho , V)} ] = \overline{ \text{Ind}_{H}^{G}[ (\rho , V) ] }
\end{align*} 
Thus, we have that
\small
\begin{align*}
    &	R_{\pm}[e] \cdot \begin{bmatrix}
        v_{1} \\
        v_{2} \\
        v_{3} \\
        v_{4} \\
    \end{bmatrix} = \begin{bmatrix}
        v_{1} \\
        v_{2} \\
        v_{3} \\
        v_{4} \\
    \end{bmatrix} \quad 		R_{\pm}[(1,2,3)] \cdot \begin{bmatrix}
        v_{1} \\
        v_{2} \\
        v_{3} \\
        v_{4} \\
    \end{bmatrix} = \omega_{\pm} \begin{bmatrix}
        v_{1} \\
        v_{4} \\
        v_{2} \\
        v_{3} \\
    \end{bmatrix} 	\quad	 R_{\pm}[(1,3,2)] \cdot \begin{bmatrix}
        v_{1} \\
        v_{2} \\
        v_{3} \\
        v_{4} \\
    \end{bmatrix} =  \omega_{\mp} \begin{bmatrix}
        v_{1} \\
        v_{3} \\
        v_{4} \\
        v_{2} \\
    \end{bmatrix}  \\
    &	R_{\pm}[(1,2,4)] \cdot \begin{bmatrix}
        v_{1} \\
        v_{2} \\
        v_{3} \\
        v_{4} \\
    \end{bmatrix} = \begin{bmatrix}
        v_{2} \\
        \omega_{\pm} 	v_{4} \\
        \omega_{\mp} 	v_{3} \\
        \omega_{\mp}  v_{1} \\
    \end{bmatrix} \quad 		R_{\pm}[(1,3)(2,4)] \cdot \begin{bmatrix}
        v_{2} \\
        v_{1} \\
        v_{4} \\
        v_{3} \\
    \end{bmatrix} = \begin{bmatrix}
        \omega_{\pm}  v_{1} \\
        \omega_{\mp}	v_{2} \\
        v_{3} \\
        v_{4} \\
    \end{bmatrix} 	\quad  R_{\pm}[(2,4,3)] \cdot \begin{bmatrix}
        v_{1} \\
        v_{2} \\
        v_{3} \\
        v_{4} \\
    \end{bmatrix} = \begin{bmatrix}
        \omega_{\mp}	v_{2} \\
        \omega_{\pm}	v_{3} \\
        v_{1} \\
        \omega_{\pm}	v_{4} \\
    \end{bmatrix}  \\
    &	R_{1}[(2,3,4)] \cdot \begin{bmatrix}
        v_{1} \\
        v_{2} \\
        v_{3} \\
        v_{4} \\
    \end{bmatrix} = \begin{bmatrix}
        v_{3} \\
        \omega_{\pm}	v_{1} \\
        \omega_{\mp}	v_{2} \\
        \omega_{\mp}	v_{4} \\
    \end{bmatrix} \quad 		R_{\pm}[(1,2)(3,4)] \cdot \begin{bmatrix}
        v_{1} \\
        v_{2} \\
        v_{3} \\
        v_{4} \\
    \end{bmatrix} = \begin{bmatrix}
        \omega_{\pm} v_{3} \\
        v_{4} \\
        \omega_{\mp} v_{1} \\
        v_{2} \\
    \end{bmatrix} 	\quad	R_{\pm}[(3,4,1)] \cdot \begin{bmatrix}
        v_{1} \\
        v_{2} \\
        v_{3} \\
        v_{4} \\
    \end{bmatrix} = \begin{bmatrix}
        \omega_{\mp} v_{3} \\
        \omega_{\pm}	v_{2} \\
        \omega_{\pm}	v_{4} \\
        v_{1} \\
    \end{bmatrix}  \\ 
    &	R_{\pm}[(1,4,3)] \cdot \begin{bmatrix}
        v_{1} \\
        v_{2} \\
        v_{3} \\
        v_{4} \\
    \end{bmatrix} = \begin{bmatrix}
        v_{4} \\
        \omega_{\mp} v_{2} \\
        \omega_{\pm} v_{1} \\
        \omega_{\mp}	v_{3} \\
    \end{bmatrix} \quad 		R_{\pm}[(2,3)(1,4)] \cdot \begin{bmatrix}
        v_{1} \\
        v_{2} \\
        v_{3} \\
        v_{4} \\
    \end{bmatrix} = \begin{bmatrix}
        \omega_{\pm}	v_{4} \\
        v_{3} \\
        v_{2} \\
        \omega_{\mp}	v_{1} \\
    \end{bmatrix} 	\quad	R_{\pm}[(2,4,3)] \cdot \begin{bmatrix}
        v_{1} \\
        v_{2} \\
        v_{3} \\
        v_{4} \\
    \end{bmatrix} = \begin{bmatrix}
        \omega_{\mp}	v_{4} \\
        v_{1} \\
        \omega_{\pm}	v_{3} \\
        \omega_{\pm}	v_{2} \\
    \end{bmatrix}  
\end{align*}
\normalsize
Working in the standard Euclidean basis, we may write this as
\small
\begin{align*}
    &	\hspace*{-1cm}	R_{\pm}[e] = \begin{bmatrix}
        1 & 0 & 0 & 0  \\
        0 & 1 & 0 & 0  \\
        0 & 0 & 1 & 0  \\
        0 & 0 & 0 & 1  \\
    \end{bmatrix}  \enspace	R_{\pm}[(1,2,3)]  = \omega_{\pm} \begin{bmatrix}
        1 & 0 & 0 & 0  \\
        0 & 0 & 0 & 1  \\
        0 & 1 & 0 & 0  \\
        0 & 0 & 1 & 0  \\
    \end{bmatrix}	 \enspace		 R_{\pm}[(1,3,2)]  = \omega_{\mp} \begin{bmatrix}
        1 & 0 & 0 & 0  \\
        0 & 0 & 1 & 0  \\
        0 & 0 & 0 & 1  \\
        0 & 1 & 0 & 0  \\
    \end{bmatrix}  \\
    &\hspace*{-1cm}	R_{\pm}[(1,2,4)] = \begin{bmatrix}
        0 & 1 & 0 & 0  \\
        0 & 0 & 0 & \omega_{\pm}  \\
        0 & 0 & \omega_{\mp} & 0  \\
        \omega_{\mp} & 0 & 0 & 0  \\
    \end{bmatrix}  \enspace	 		R_{\pm}[(1,3)(2,4)] = \begin{bmatrix}
        0 & \omega_{\pm} & 0 & 0  \\
        \omega_{\mp} & 0 & 0 & 0  \\
        0 & 0 & 0 & 1  \\
        0 & 0 & 1 & 0  \\
    \end{bmatrix} 	 \enspace	  R_{\pm}[(2,4,3)] = \begin{bmatrix}
        0 & \omega_{\mp} & 0 & 0  \\
        0 & 0 & \omega_{\pm} & 0  \\
        1 & 0 & 0 & 0  \\
        0 & 0 & 0 & \omega_{\pm}  \\
    \end{bmatrix} \\
    &\hspace*{-1cm}	R_{\pm}[(2,3,4)] =  \begin{bmatrix}
        0 & 0 & 1 & 0  \\
        \omega_{\pm} & 0 & 0 & 0  \\
        0 & \omega_{\mp} & 0 & 0  \\
        0 & 0 & 0 & \omega_{\mp}  \\
    \end{bmatrix}  \enspace	 		R_{\pm}[(1,2)(3,4)] =  \begin{bmatrix}
        0 & 0 & \omega_{\pm} & 0  \\
        0 & 0 & 0 & 1  \\
        \omega_{\mp} & 0 & 0 & 0  \\
        0 & 1 & 0 & 0  \\
    \end{bmatrix}	 \enspace	R_{\pm}[(3,4,1)]  = \begin{bmatrix}
        0 & 0 & \omega_{\mp} & 0  \\
        0 & \omega_{\pm} & 0 & 0  \\
        0 & 0 & 0 & \omega_{\pm}  \\
        1 & 0 & 0 & 0  \\
    \end{bmatrix} \\
    &\hspace*{-1cm}	R_{\pm}[(1,4,3)] =  \begin{bmatrix}
        0 & 0 & 0 & 1  \\
        0 & \omega_{\mp} & 0 & 0  \\
        \omega_{\pm} & 0 & 0 & 0  \\
        0 & 0 & \omega_{\mp} & 0  \\
    \end{bmatrix}  \enspace	 	R_{\pm}[(2,3)(1,4)]  =  \begin{bmatrix}
        0 & 0 & 0 & \omega_{\pm}  \\
        0 & 0 & 1 & 0  \\
        0 & 1 & 0 & 0  \\
        \omega_{\mp} & 0 & 0 & 0  \\
    \end{bmatrix}  \enspace		R_{\pm}[(2,4,3)]  =  \begin{bmatrix}
        0 & 0 & 0 & \omega_{\mp}  \\
        1 & 0 & 0 & 0  \\
        0 & 0 & \omega_{\pm} & 0  \\
        0 & \omega_{\pm} & 0 & 0  \\
    \end{bmatrix} \\
\end{align*}
\normalsize
\begin{table}[h!]
    \centering
    \begin{tabular}{ |c|c|c|c|c| } 
        \hline
        & 	$e$ & $(1,2,3)$ & $(1,3,2)$ & $(12)(34)$ \\ 
        \hline
        $\chi_{R_{1}}$ & 4 & 1 & 1 & 0 \\ 
        \hline
        $\chi_{R_{+}}$ & 4 & $\omega_{+}$ & $\omega_{-}$ & 0 \\ 
        \hline
        $\chi_{R_{-}}$ & 4 & $\omega_{-}$ & $\omega_{+}$ & 0 \\ 
        \hline
    \end{tabular}
    \vspace{0.2cm}
    \caption{Character Table for induced representations of the irreducibles $( \rho_{1}, V_{1})$, $( \rho_{+}, V_{+})$ and $( \rho_{-}, V_{-})$ of $\mathbb{Z}_{3}$ on $A_{4}$, $R_{+} = \text{Ind}_{\mathbb{Z}_{3}}^{A_{4}}( \rho_{+}  )$ and $R_{-} = \text{Ind}_{\mathbb{Z}_{3}}^{A_{4}}( \rho_{-}  )$. $\omega_{+} = \exp(\frac{2\pi i}{3})= \bar{\omega}_{-}$. }
\end{table}
The group $A_{4}$ has four conjugacy classes: $e$, $(1,2,3)$, $(1,2)(3,4)$ and $(1,3,2)$. The four irreducible representations of $A_{4}$ are: The trivial $( \sigma_{1} , W_{1})$ representation, two conjugate one-dimensional representations $( \sigma_{1,+} , W_{1,+}) , (\sigma_{1,-}, W_{1,-})$ and one three dimensional representation $(\sigma_{3},W_{3})$.
\begin{table}[h!]
    \begin{tabular}{ |c|c|c|c|c| } 
        \hline
        & 	$e$ & $(1,2,3)$ & $(1,3,2)$ & $(12)(34)$ \\ 
        \hline
        $\chi_{1}$ & 	1 & 1 & 1 & 1 \\ 
        \hline
        $\chi_{1,-}$ & 1  &  $\omega_{+}$ & $\omega_{-}$ & 1 \\ 
        \hline
        $\chi_{1,+}$ &  1 & $\omega_{-}$ & $\omega_{+}$ & 1 \\ 
        \hline
        $\chi_{3}$ & 	3 & 0 & 0 & -1 \\ 
        \hline
    \end{tabular}
    \vspace{0.4cm}
    \centering
    \caption{Character Table for $A_{4}$. $\omega_{+} = \exp(\frac{2\pi i}{3})= \bar{\omega}_{-}$. $( \sigma_{1,+} , W_{1,+})$ and $( \sigma_{2,-} , W_{2,-})$ are conjugate representations. }
\end{table}
We can thus compute the induction coefficients of the induced representation of $\mathbb{Z}_{3}$ on $A_{4}$. We have that
\begin{align*}
    &\text{Ind}_{\mathbb{Z}_{3}}^{A_{4}}[  (\rho_{1} , V_{1} ) ] = ( \sigma_{3} , W_{3}) \oplus ( \sigma_{1} , W_{1})  \\ 	&\text{Ind}_{\mathbb{Z}_{3}}^{A_{4}}[  (\rho_{+} , V_{+} )  ] = ( \sigma_{3} , W_{3})\oplus ( \sigma_{1,+} , W_{1,+})  \\
    &\text{Ind}_{\mathbb{Z}_{3}}^{A_{4}}[  (\rho_{-} , V_{-} ) ] = ( \sigma_{3} , W_{3})\oplus ( \sigma_{1,-} , W_{1,-}) 
\end{align*}
Using Frobinous Reciprocity, we can derive the restrictions of $A_{4}$ irreducibles. We have that
\begin{align*}
&\text{Res}_{\mathbb{Z}_{3}}^{A_{4}}[ ( \sigma_{3} , W_{3})  ] = (\rho_{1} , V_{1} ) \oplus (\rho_{+} , V_{+} ) \oplus (\rho_{-} , V_{-} ) \\ 
&\text{Res}_{\mathbb{Z}_{3}}^{A_{4}}[ ( \sigma_{1+} , W_{1+})   ] = (\rho_{+} , V_{+} ) \\
&\text{Res}_{\mathbb{Z}_{3}}^{A_{4}}[ ( \sigma_{1-} , W_{1-}) ] = (\rho_{-} , V_{-} ) \\
&\text{Res}_{\mathbb{Z}_{3}}^{A_{4}}[ ( \sigma_{1} , W_{1}) ] = (\rho_{1} , V_{1} )
\end{align*}
\begin{figure}[H]
\centering
\begin{tabular}{cc} 
    \includegraphics[width=0.45\textwidth]{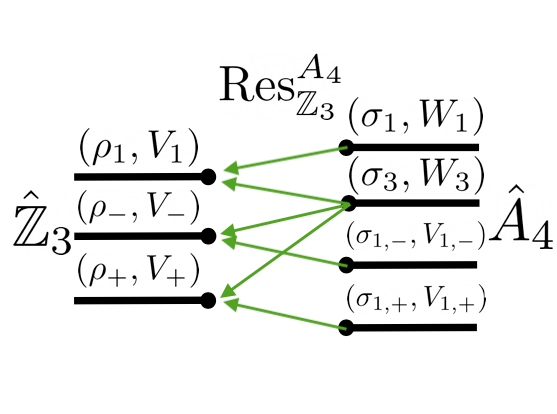} \quad \quad
    \includegraphics[width=0.45\textwidth]{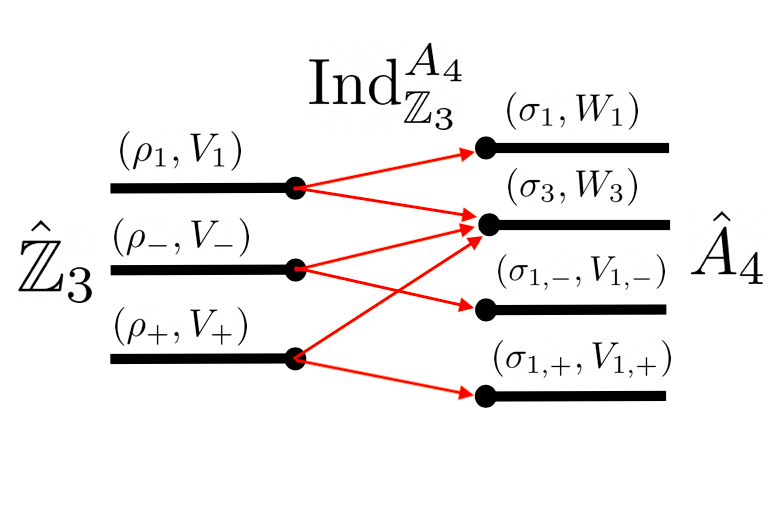}
\end{tabular}
\vspace{-0.5cm}
\caption{ Left: Decomposition of the restricted representation $\Res_{ \mathbb{Z}_{3} }^{ A_{4} }$ of $A_{4}$-irreducibles $( \sigma , W_{\sigma}) \in \widehat{A}_{4}$ into $\mathbb{Z}_{3}$-irreducibles  $(\rho , V_{\rho}) \in \widehat{\mathbb{Z}}_{3} $. Not every $\mathbb{Z}_{3}$-representation can be realized as the restriction of a $A_{4}$-representation. Right: Decomposition of the induced representation $\Ind_{\mathbb{Z}_{3}}^{ A_{4} }$ for $\mathbb{Z}_{3}$-irreducibles $(\rho , V_{\rho}) \in \widehat{\mathbb{Z}}_{3} $ into $A_{4}$-irreducibles $( \sigma , W_{\sigma}) \in \widehat{A}_{4}$. Not every $A_{4}$-representation can be realized as the induction of a $\mathbb{Z}_{3}$-representation. }\label{Figure:Induced_z_3_appendix}
\end{figure}
We are only interested in real representations. The most general real representation of $\mathbb{Z}_{3}$ is given by
\begin{align*}
    ( \rho , V) = m_{1} (\rho_{1} , V_{1} ) \oplus  m_{c} [ ( \rho_{+} ,V_{+} ) \oplus ( \rho_{-} ,V_{-} ) ]
\end{align*}
where $m_{1}$ and $m_{c}$ are integers. The dimension of the vector space $V$ is $\dim V = m_{1} + m_{c}$. The induced representation of $( \rho , V)$ is
\begin{align*}
    (R,W) =	\text{Ind}_{\mathbb{Z}_{3}}^{A_{4}}[ ( \rho , V)  ] = [m_{1} +  2m_{c}] ( \sigma_{3} , W_{3} ) \oplus  m_{c}[  ( \sigma_{1,+} , W_{1,+} ) \oplus ( \sigma_{1,-} , W_{1,-} ) ] \oplus m_{1} ( \sigma_{1}  , W_{1} )
\end{align*}
where the vector space $W$ of the induced representation has dimension $\dim W = 3(  m_{1} + 2m_{c} ) + 2m_{c} + m_{1}  = 4m_{1} + 8m_{c} = 4 (m_{1} + 2m_{c}) = 4 \dim V$ as expected. This result, although simple is extremely satisfying as it shows that any function on $A_{4}$ can be lifted from a function on $\mathbb{Z}_{3}$. To see this, note the following: By the Peter-Weyl theorem, the left regular representation $(  L  , \mathbb{R}^{\mathbb{Z}_{3}} )$ decomposes as
\begin{align*}
    ( L ,\mathbb{R}^{ \mathbb{Z}_{3} } ) = ( \rho_{1} , V_{1} ) \oplus [ ( \rho_{+} , V_{+} ) \oplus ( \rho_{-} , V_{-} ) ] 
\end{align*}
Thus, the induced representation of $ (L , \mathbb{R}^{\mathbb{Z}_{3}} ) $ is from $\mathbb{Z}_{3}$ to $A_{4}$ is thus
\begin{align*}
    (R,W) = \text{Ind}_{Z_{3}}^{A_{4}}[  \mathbb{R}^{ \mathbb{Z}_{3} }     ] = 3 (  \sigma_{3} , W_{3} ) \oplus [ (  \sigma_{1,+}, W_{1,+} ) \oplus ( \sigma_{1,-} , W_{1,-} ) ] \oplus  ( \sigma_{1} , W_{1} )
\end{align*}
Now, again by the Peter-Weyl theorem, the left regular representation $( L , \mathbb{R}^{A_{4}})$ of $A_{4}$ decomposes as
\begin{align*}
    (L , \mathbb{R}^{A_{4} } ) = 3 (  \sigma_{3} , W_{3} ) \oplus [ (  \sigma_{1,+}, W_{1,+} ) \oplus ( \sigma_{1,-} , W_{1,-} ) ] \oplus  ( \sigma_{1} , W_{1} )
\end{align*}
So the induced representation of the left regular representation of $\mathbb{Z}_{3}$ has the same decomposition into irreducibles as the left regular representation of $A_{4}$. Representations are completely determined by their decomposition into irreducibles and
\begin{align}\label{Equation:Induced Rep}
    (L ,	\mathbb{R}^{A_{4} } ) = \text{Ind}_{Z_{3}}^{A_{4}}[ (L, \mathbb{R}^{ \mathbb{Z}_{3} }   )   ]
\end{align}
Ergo, the space of functions from $A_{4}$ into $\mathbb{R}$ is identical to the induced representation from $\mathbb{Z}_{3}$ to $A_{4}$ of the space of functions of $\mathbb{Z}_{3}$ into $\mathbb{R}$. Using the linearity of the induced representation and taking the $c$-fold direct sum of both sides of \eqref{Equation:Induced Rep}, we have that
\begin{align*}
    (L ,	(\mathbb{R}^{c} )^{A_{4} } ) = \text{Ind}_{Z_{3}}^{A_{4}}[ (L,  ( \mathbb{R}^{c} )^{ \mathbb{Z}_{3} }  )    ]
\end{align*}
Thus, as expected, the induced representation bijectively maps group valued functions from $\mathbb{Z}_{3} \rightarrow \mathbb{R}^{c}$ into group valued functions from $A_{4} \rightarrow \mathbb{R}^{4c}$.

\end{document}